\pdfoutput=1
\documentclass{article}
\usepackage{iclr2027_conference,times}
\usepackage[T1]{fontenc}
\usepackage[utf8]{inputenc}
\usepackage{amsmath,amssymb,booktabs,tabularx,array,multirow,graphicx}
\usepackage{hyperref,url}
\usepackage{microtype}
\usepackage{algorithm,algpseudocode}
\usepackage[listings,breakable,skins]{tcolorbox}
\newtcblisting{promptlisting}[1]{listing only, breakable, colback=gray!5, colframe=black!60, boxrule=0.4pt, arc=1pt, left=3pt, right=3pt, top=2pt, bottom=2pt, title={#1}, fonttitle=\small\bfseries, listing options={basicstyle=\ttfamily\footnotesize, breaklines=true, breakatwhitespace=false, columns=fullflexible, keepspaces=true, upquote=true, breakindent=0pt, aboveskip=0pt, belowskip=0pt, literate={—}{{\textemdash}}1 {σ}{{$\sigma$}}1}}
\newcommand{\fixed}{\mathcal{F}}
\newcommand{\broken}{\mathcal{B}}

\newcolumntype{Y}{>{\raggedright\arraybackslash}X}
\usepackage{collcell}
\ExplSyntaxOn
\box_new:N \l__num_cell_box
\box_new:N \l__num_tmpl_box
\NewDocumentCommand \numcell { m m }
  {
    \tl_set:Ne \l_tmpa_tl { \tl_trim_spaces:n {#2} }
    \regex_match:nVTF { \d } \l_tmpa_tl
      {
        \hbox_set:Nn \l__num_cell_box { \tl_use:N \l_tmpa_tl }
        \hbox_set:Nn \l__num_tmpl_box {#1}
        \dim_compare:nNnTF { \box_wd:N \l__num_cell_box } > { \box_wd:N \l__num_tmpl_box }
          { \box_use:N \l__num_cell_box }
          { \hbox_to_wd:nn { \box_wd:N \l__num_tmpl_box } { \hfil \box_use:N \l__num_cell_box } }
      }
      { \tl_use:N \l_tmpa_tl }
  }
\ExplSyntaxOff
\newcolumntype{N}[1]{>{\collectcell{\numcell{#1}}}c<{\endcollectcell}}
\newcommand{\hd}[1]{\multicolumn{1}{c}{#1}}
\usepackage{xcolor}

\newcommand{\few}{\rlap{$^{\S}$}}
\newcommand{\tmark}[1]{\rlap{$^{\rm #1}$}}
\newcommand{\grouprow}[2]{\multicolumn{#1}{@{}l}{\textit{#2}}\\}
\newcommand{\tabnote}[1]{\par\vspace{2pt}{\raggedright\scriptsize\textit{Note.}~#1\par}}
\makeatletter
\newcommand{\captionof}[1]{\def\@captype{#1}\caption}
\makeatother
\title{Where Self-Evolving Memory Systems\\Lose Their Gains:\\Failure Modes of the Evolution Loop}
\author{\textbf{Bowen Fang$^{1,2,3}$, Xiaobin Wang$^{4}$, Tianshu Xiao$^{3}$, Kangning Zhang$^{5}$,}\\
\textbf{Xiaoqing Zhang$^{3,6}$\thanks{Corresponding author.}, Liang Wang$^{1,2}$, Xiaonan Shi$^{4}$, Junming Chen$^{4}$, Pengjun Xie$^{4}$}\\[2pt]
{\small\normalfont$^{1}$New Laboratory of Pattern Recognition (NLPR), Institute of Automation, Chinese Academy of Sciences}\\
{\small\normalfont$^{2}$School of Artificial Intelligence, University of Chinese Academy of Sciences}\\
{\small\normalfont$^{3}$Zhongguancun Academy}\\
{\small\normalfont$^{4}$Alibaba Token Hub, Alibaba Group \quad $^{5}$Shanghai Jiao Tong University}\\
{\small\normalfont$^{6}$Zhongguancun Institute of Artificial Intelligence}}
\iclrfinalcopy%
\begin{document}
\addtocontents{toc}{\protect\setcounter{tocdepth}{-1}}
\maketitle
\vspace{-11pt}
\begin{abstract}
Self-evolving memory systems use large language model (LLM) agents to evaluate performance, diagnose failures, and revise shared memory components. These loops often improve early and then plateau. Their logic rests on three assumptions: evaluation identifies failures, edits repair them, and accepted repairs persist. Using question-level evidence independent of the agent, we test where each assumption fails in several self-evolution methods for memory systems on three conversational memory benchmarks and identify three failure modes.
(1)~\textit{Evaluation provides unstable feedback for repair.} When an unchanged system is evaluated again, its set of failed questions changes, and at least half of the accepted single-step gains fall within this evaluation noise.
(2)~\textit{Edits break questions they do not target.} Edits aimed at the stage with the most failures raise scores much more often than other edits, yet do not reduce regressions, largely because the components they change are shared across queries.
(3)~\textit{Across rounds, repairs do not accumulate reliably.} Repairs decline toward the level of answer flips observed in an unchanged system, while regressions persist.
To address these failure modes, we extend MemPro, one of the methods we study, with verified feedback, repair rounds, and paired acceptance. Each mechanism reduces the loss it targets and raises development accuracy. Held-out accuracy also improves by about 3--5 points with much lower run-to-run variation. Our results suggest that self-evolving loops should evaluate each revision more than once and track repairs and regressions at the question level.
\end{abstract}
\vspace{-6pt}
\begin{figure}[h!]
\vspace{-6pt}
\centering
\includegraphics[width=\linewidth]{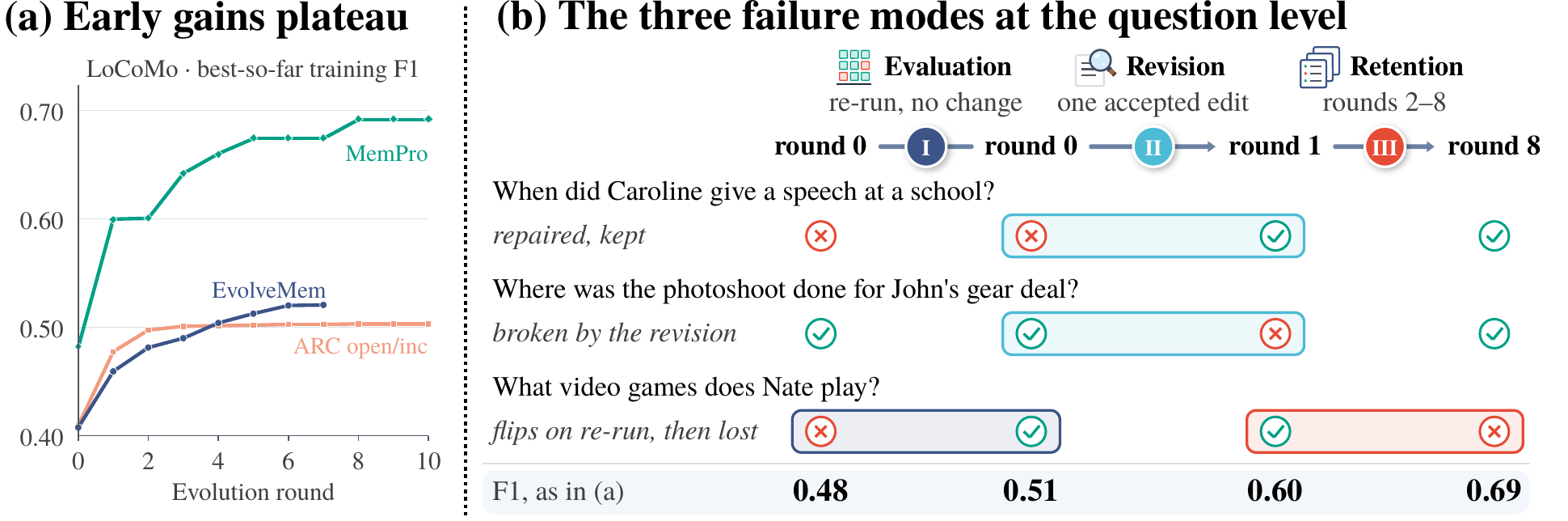}
\vspace{-16pt}
\caption{\textbf{F1 over evolution rounds and the three failure modes at the question level.} (a) Best-so-far training F1 on LoCoMo categories 1--4 (Appendix~\ref{app:conventions}). (b) The three failure modes on three development questions of the MemPro run in (a). The round-1 revision adds an answer prompt shared by all queries. It repairs the first question and breaks the second (II). The third question already turns correct when round 0 is re-evaluated without any change (I), so its recorded repair in round 1 may be only evaluation noise, and it fails again by round 8 (III).}
\label{fig:overview}
\vspace{-8pt}
\end{figure}
\raggedbottom
\section{Introduction}
\label{sec:intro}
{\looseness=-1 Large language model (LLM) agents rely on memory systems to extract, store, and retrieve information from past sessions that no longer fits in their context window~\citep{memgpt,amem,longmemeval,agentnative}. Since no single memory design works best across tasks~\citep{agentnative}, recent work turns to self-evolving memory systems, in which an evolving agent tunes the design in an evolution loop rather than leaving it fixed by hand~\citep{evolvemem,mempro}. In each round, the evolving agent evaluates the current version on a development set of questions, diagnoses the failed questions, and proposes a revision to components shared by all queries. The loop accepts the revision only if its development score improves, and later rounds build on accepted versions. Such loops typically improve quickly in their first few rounds and then plateau (Figure~\ref{fig:overview}a, \citealp{evolvemem,mempro}). Why the early gains are hard to sustain remains poorly understood.\par}

{\looseness=-1 To study this question, we examine three assumptions on which the evolution loop rests, on its evaluation, on its revisions, and on its retention of accepted repairs. Each assumption can fail, and the loop then loses a gain. We call these three failures the failure modes of the loop. \textbf{Evaluation (I)} must report which questions fail and how well a version performs, but re-evaluating an unchanged version can change both, and some failures depend on the answer model or the judge rather than on the memory. \textbf{Revision (II)} must repair the cause of a failure, but the evolving agent may miss the cause, and even a well-aimed change to a shared component can break other questions. \textbf{Retention (III)} must keep the accepted repairs, but a recorded repair may be only evaluation noise, and the question then fails again in a later round. Each failure mode also feeds the next one, since a revision can only target the failures that evaluation reports, and retention can only keep the repairs that revisions make.\par}

{\looseness=-1 These losses are hard to see from inside the loop, because it evaluates each version once and accepts a revision on its net change in development score. One evaluation cannot separate a real repair from evaluation noise, and a net change can hide the questions that a revision breaks. Studies of self-evolving agents find that evolver capability barely changes their gains~\citep{harnessupdating} and that evolved systems do not consistently beat test-time scaling at matched budgets~\citep{harnessrethink}. Work on acceptance rules and model selection shows that keeping every score increase on a noisy development set accumulates false commits~\citep{pace,cawley}. However, no prior work tests these three assumptions to locate where a self-evolving memory loop loses its gains. In this paper, we examine the full evolution loop of EvolveMem~\citep{evolvemem}, MemPro~\citep{mempro}, AutoResearchClaw (ARC)~\citep{arc}, and GEPA~\citep{gepa} on three conversational memory benchmarks~\citep{locomo,memgallery,longmemeval}. We record every question's outcome on every version and intervene by re-evaluating unchanged versions, replaying failed questions, and answering them with their annotated evidence in the answer model's context.\par}

\begin{figure}[t]
\centering
\includegraphics[width=\linewidth]{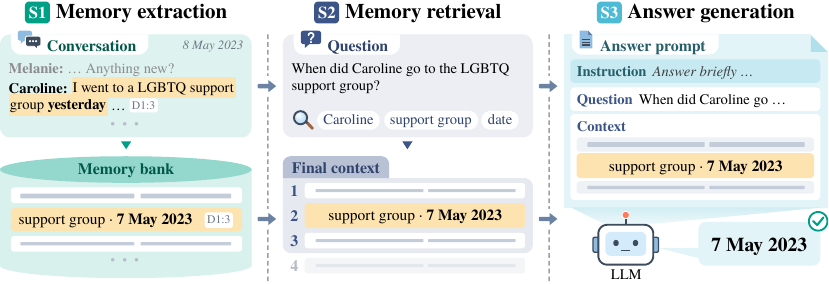}
\vspace{-15pt}
\caption{\textbf{The memory system that the loop revises.} The system follows a general pipeline, whose three failure stages group the memory modules of \citet{agentnative}, with planning, integration, and context assembly in S2 and the answer prompt and answer parsing in S3. The example traces one question through the three stages.}
\label{fig:stages}
\vspace{-8pt}
\end{figure}

{\looseness=-1 These records give one test per failure mode. For evaluation, we measure how much the failed questions and score change when an unchanged version is re-evaluated, when one pipeline stage is repeated at a time, or when the full loop is rerun. When annotated evidence resolves a failure, two further interventions locate it at memory extraction, memory retrieval, or answer generation (Figure~\ref{fig:stages}). For revision, we check whether the diagnosis and executed change target the bottleneck stage, where most attributable failures lie, and count each revision's repairs and regressions, the questions it turns from wrong to correct and from correct to wrong against its parent. For retention, we follow these counts along the accepted lineage and compare them with the flips of an unchanged system. Together, these counts decompose each step's observed change into repairs, regressions, and no-edit flips. Because this decomposition requires only per-item outcomes and repeated no-edit evaluations, it applies beyond memory systems to other iterative self-improving loops.\par}

{\looseness=-1 We find that the loop violates each of its three assumptions. In evaluation, re-evaluating an unchanged version alters its failed questions, and half or more of the accepted gains lie within this noise. Repeating one stage at a time on MemPro's root shows that these flips arise mainly during answer generation rather than from the judge or memory rebuilding, while re-extracting full conversations adds further flips in pipelines such as EvolveMem. Moreover, only a quarter to two fifths of the failures reported to MemPro's loop trace to a memory stage. In revision, 17 of the 28 revisions targeting the bottleneck stage raise the score in EvolveMem and the original ARC loop, against 1 of the 42 that do not. Yet targeting that stage breaks no fewer correct questions overall, largely because revisions change components shared by all queries. Even so, we find that stage information remains useful, as a proposer that follows the bottleneck stage through a fixed menu reaches a useful retrieval upgrade earlier than the original proposer or random menu orders. In retention, repairs shrink after the first rounds while regressions persist, so late revisions break at least as many questions as they repair in all but two settings. Subtracting no-edit flips shows that noise accounts for a third to two thirds of late flips in F1-selected loops while regressions offset the remaining repairs, and about half of the failures an unchanged version turns correct on re-evaluation fail again on a third evaluation.\par}

{\looseness=-1 These findings suggest that a loop should verify its feedback, weigh each revision by what it breaks as well as what it repairs, and keep only repairs that hold. We therefore add one mechanism per failure mode to MemPro. Verified feedback tells the agent which failures recur across repeated evaluations, and a repair round asks it to narrow a revision that breaks questions. Paired acceptance keeps a revision only if its repairs outnumber its regressions, counting only those that hold on re-evaluation. Each mechanism reduces the loss it targets and raises development accuracy. Held-out accuracy also rises by about 3--5 points with much lower run-to-run variation (Section~\ref{sec:intervention}).\par}

We make three contributions.
{\setlength{\leftmargini}{1.2em}\setlength{\topsep}{1pt}\setlength{\partopsep}{0pt}
\begin{itemize}
    \setlength{\itemsep}{1pt}
    \setlength{\parsep}{0pt}
    \item {\looseness=-1 \textbf{A full-loop study and step-level decomposition.} We locate where self-evolving memory loops lose their gains, in evaluation, revision, and retention, across four methods and three memory benchmarks. Using per-question outcomes and interventions, we decompose each step into repairs, regressions, and no-edit flips, an analysis applicable to other iterative self-improving loops.\par}
    \item {\looseness=-1 \textbf{Findings.} We find losses under each assumption. Half or more of the accepted gains lie within evaluation noise driven mainly by answer generation rather than the judge, targeting the bottleneck stage raises repairs without reducing regressions from shared components, and subtracting no-edit flips shows that repairs shrink across rounds while regressions persist and offset late gains.\par}
    \item {\looseness=-1 \textbf{Improvements.} We add verified feedback, repair rounds, and paired acceptance to MemPro. These mechanisms reduce the losses they target, raise development accuracy, and lower run-to-run variation on held-out questions.\par}
\end{itemize}}
\flushbottom

\section{Measuring the Evolution Loop}
\label{sec:protocol}
{\looseness=-1 We study the three failure modes in four self-evolution methods for memory systems, EvolveMem, ARC, MemPro, and GEPA. EvolveMem~\citep{evolvemem} tunes the configuration of a retrieval-based memory pipeline, whereas MemPro~\citep{mempro} lets its agent rewrite its own memory program. AutoResearchClaw (ARC)~\citep{arc} is the autonomous research loop that discovered the Omni-SimpleMem memory system~\citep{omnimem}. We run it on EvolveMem's initial pipeline in eight configurations, which vary its action space, the version its diagnoser reads, memory rebuilding, and guards. We also run it on Omni-SimpleMem itself (Appendix~\ref{app:omni}). GEPA~\citep{gepa} optimizes the prompts of MemPro's pipeline. Because it writes no diagnosis, we measure its feedback and its retention but cannot compare a diagnosis with its edits. We use LoCoMo~\citep{locomo}, caption-only Mem-Gallery~\citep{memgallery}, and LongMemEval~\citep{longmemeval} as benchmarks. On the first two, EvolveMem and ARC evolve on all questions and select by their native F1-based score, whereas MemPro and GEPA evolve on a random 10\% of the questions, select by judge accuracy, and hold out the rest. On LongMemEval, all methods evolve on 100 questions and hold out the other 400. Details of each configuration are provided in Appendix~\ref{app:loops}, with a summary in Table~\ref{tab:systems}, and the data splits are given in Appendix~\ref{app:data}. The main runs use Qwen models for the memory pipeline and the judge, and we repeat the measurements with GPT-4o and GPT-4o-mini as diagnoser, answer model, and ARC agent in Appendix~\ref{app:substitutions}. The Mem-Gallery runs give the answer model the dataset caption of the query image, and Appendix~\ref{app:galleryvlm} repeats them with the image itself.\par}

\paragraph{The evolution loop.}
Although these systems differ in what they edit, their loops share one round structure. The loop starts from an initial version called the root, and each round evaluates a version, diagnoses its failures, and executes a proposed change,
\begin{equation}
\begin{gathered}
\mathcal{E}(R)=\big\{y_q^{(k)}(R),\ \text{trace}\big\}_{q\in Q},\\
(\text{claim},\,\text{proposal})=D\big(\mathcal{E}(R_{\rm seen})\big),\\
R'=A\big(R_{\rm applied},\,\text{proposal}\big),\qquad \text{action}=\mathrm{diff}(R_{\rm applied},R'),
\end{gathered}
\label{eq:loop}
\end{equation}
{\looseness=-1 where $y_q^{(k)}(R)\in\{0,1\}$ is the outcome of question $q$ on version $R$ in evaluation $k$, scored as nonzero F1 or by the judge's verdict. The diagnoser $D$ returns a claim, the failure stage that its diagnosis names, together with a proposal. The executor $A$ applies the proposal to a parent version $R_{\rm applied}$ and returns the revision $R'$. The action, the difference between $R_{\rm applied}$ and $R'$, is the edit that the loop actually executes. Guards, no-ops, and execution deviations can make it differ from the proposal. An acceptance decision then selects the version that the next round builds on, which we call the incumbent. Note that the version the diagnoser reads, $R_{\rm seen}$, need not be the parent. ARC's labels record two choices. Under closed, ARC edits only the configuration, and under open it also edits the extraction and answer prompts. Its diagnoser reads the root under r0 and the incumbent under inc. Although $\mathcal{E}$ can record every outcome $y_q^{(k)}$, every loop evaluates each version once and accepts a revision on its net change in score. The diagnosis and the acceptance decision therefore rarely see whether an outcome $y$ holds across evaluations, or which questions an edit has repaired and which it has broken.\par}

\paragraph{Failure stages.}
To judge whether a claim, a proposal, or an action targets the right stage, we need a reference that does not come from the diagnosis itself. We therefore define stages by intervention rather than by the system's own account, and the same interventions sort the failures that the loop receives. Each failure is first replayed on the same version and then answered from its annotated gold evidence. A failure that turns correct on replay is \emph{replay-correct}, and one that stays wrong even with the gold evidence is \emph{gold-unresolved}. For the remaining failures, we ask at which stage the evidence was lost (Appendix~\ref{app:probedef}). If no memory was extracted from the evidence, or the extracted memories do not answer the question, the failure belongs to memory extraction (S1). If the memories already in the retrieved context answer it, it belongs to answer generation (S3), and if only injecting all of them does, to memory retrieval (S2). We call these failures \emph{attributable} and define the \emph{bottleneck stage} of a version as the stage with the most attributable failures,
\begin{equation}
s^*(R)=\operatorname*{arg\,max}_{s\in\{\mathrm{S1},\mathrm{S2},\mathrm{S3}\}}\big|\{q:\ y_q(R)=0,\ \mathrm{stage}(q)=s\}\big|,
\label{eq:reference}
\end{equation}
Separating S1 from S2 requires \emph{provenance}, the turns each memory was extracted from. As shown in Appendix~\ref{app:provenance}, recording it causes no measurable loss in accuracy.

\paragraph{Repairs and regressions.}
Targeting the right stage, however, does not show what an edit achieves. To measure it, we pair each revision $R'$ with the parent it actually modifies, $R=R_{\rm applied}$, and count its repairs and regressions,
\begin{equation}
\fixed=\{q:\,y_q(R)=0,\ y_q(R')=1\},\qquad
\broken=\{q:\,y_q(R)=1,\ y_q(R')=0\}.
\label{eq:ledger}
\end{equation}
{\looseness=-1 On the same $N$ questions, the change in accuracy is then exactly $(|\fixed|-|\broken|)/N$. The net change on which the loop accepts a revision thus shows only the difference between the two counts. Re-evaluations of unchanged versions give the noise reference for both counts (Appendix~\ref{app:conventions}).\par}

\section{Failure Mode I: Evaluation Gives Unstable Feedback}
\label{sec:feedback}
{\looseness=-1 Every later decision of the loop builds on the evaluation $\mathcal{E}$. From each evaluation, the loop takes two signals, the failed questions that say what to repair and the score that says which version to keep~\citep{evolvemem,mempro,gepa}. Evaluation must therefore be stable and attributable. Evaluated again, a version should fail on the same questions with the same score, and each failure should expose a defect of the memory system whose cause can be located. We therefore ask two questions: whether repeated evaluations and loop reruns return consistent failure sets and gains, and whether the reported failures actually trace to the memory system.\par}

\begin{table}[t]
\centering\small
\caption{Variation of the feedback under repetition on LoCoMo. (a) The version each loop selected, evaluated three more times. (b) The whole loop, rerun four times from the same root. Scores are the loops' own, mixed F1 for EvolveMem and ARC in points (pts) and judge accuracy for MemPro and GEPA in pp. The same two measurements on LongMemEval are in Appendix Table~\ref{tab:lmenoise}.}
\label{tab:noise}\label{tab:replicates}
\setlength{\tabcolsep}{3pt}
\begin{tabular*}{\linewidth}{@{\extracolsep{\fill}}lcN{00.0}cN{00/00}cN{00}N{$+00.0$}cN{$+00.0$}c@{}}
\toprule
 & & \multicolumn{4}{c}{\makebox[0pt]{\textbf{(a) Same version, 3 more evaluations}}} & \multicolumn{5}{c}{\makebox[0pt]{\textbf{(b) Same root, loop run 4 times}}} \\
\cmidrule(lr){3-6}\cmidrule(l){7-11}
 & Selection & Flip & $2\sigma$ of & \multicolumn{2}{c}{\makebox[0pt]{Steps within $2\sigma$}} & & \multicolumn{2}{c}{\makebox[0pt]{Dev.\ gain}} & \multicolumn{2}{c}{\makebox[0pt]{Held-out gain}} \\
\cmidrule(lr){5-6}\cmidrule(lr){8-9}\cmidrule(l){10-11}
System & metric & rate (\%) & score & $k/n$ & (\%) & Rounds & Mean & SD & Mean & SD \\
\midrule
EvolveMem      & F1    & 2.7  & 1.4 & 7/14  & 50 & 7   & $+8.2$ & 1.0 & $+9.5$ & 0.9 \\
ARC closed/r0  & F1    & 2.7  & 2.7 & 21/31 & 68 & 10  & $+10.8$ & 0.8 & $+10.5$ & 0.7 \\
ARC closed/inc & F1    & 3.2  & 1.6 & 15/27 & 56 & 10  & $+11.3$ & 0.5 & $+10.9$ & 0.8 \\
ARC open/r0    & F1    & 2.7  & 1.5 & 9/17  & 53 & 10  & $+6.2$ & 0.6 & $+0.9$ & 1.9 \\
ARC open/inc   & F1    & 2.4  & 1.6 & 9/16  & 56 & 10  & $+6.6$ & 1.6 & $+7.2$ & 4.9 \\
MemPro         & Judge & 8.7  & 4.7 & 22/23 & 96 & 10  & $+16.2$ & 0.5 & $+3.8$ & 2.0 \\
GEPA           & Judge & 10.8 & 5.3 & 15/17 & 88 & 29 & $+5.7$ & 1.4 & $+0.2$ & 1.0 \\
\bottomrule
\end{tabular*}
\vspace{-14pt}
\end{table}

\paragraph{Reruns change the failure set and the size of the gain.}
{\looseness=-1 We first test whether a fixed version gives reproducible feedback. When each loop's selected version is evaluated three more times with its configuration, memory, and prompts fixed, Table~\ref{tab:noise}a shows that re-evaluation alone alters the set of failed questions. Measured by the flip rate, the share of questions whose outcome $y_q$ changes between two evaluations, about 3\% of outcomes flip in the F1-selected loops, and about a tenth in the judge-selected ones. Most of this variation comes from regenerating answers rather than from the judge. On MemPro's root, rejudging fixed answers flips 0.4\% of outcomes, whereas regenerating from fixed memory flips 7.5\% and rebuilding memory first flips 7.8\% (Appendix~\ref{app:gold}, Table~\ref{tab:noisecontrols}). Token F1 flips on the same answers nearly as often. The failure list that the diagnoser reads also moves more than the score. Re-evaluating MemPro's roots, for instance, turns a third of their failures correct, while the net score moves by under 4 pp. Compared against twice the standard deviation ($2\sigma$) of score differences between evaluations, 50--96\% of the accepted steps change the score by less than $2\sigma$ of this variation. Such gains need not be spurious, but a loop that evaluates each revision once cannot tell them from a re-evaluation of its parent~\citep{pace}.\par}

{\looseness=-1 To test whether this single-step noise averages out across rounds, we run each loop four times from the same root and report, on the development and on the held-out questions, the gain of its best version over the root. As shown in Table~\ref{tab:noise}b, reruns reproduce the direction of the gain but not its size. Every F1-selected loop gains in every run, yet the spread of the selected versions across runs is several times that of the root. The two open ARC variants also end some runs on a version identical to their root. Among the judge-selected loops, MemPro's development gain is stable across runs, whereas its held-out gain varies with an SD of 2.0 pp, and GEPA's held-out gain changes sign. Moreover, the four runs in Table~\ref{tab:noise}b share one extracted memory, whereas rebuilding EvolveMem's memory with unchanged code alone flips 17\% of outcomes, five times the flips of answering twice from the same memory (Appendix~\ref{app:provenance}). Even the spreads in Table~\ref{tab:noise}b still understate the variation of a full rerun. Because a single run of a loop is only one draw~\citep[cf.][]{bouthillier2021variance}, Section~\ref{sec:intervention} compares each intervention across multiple runs.\par}

\paragraph{Only a subset of observed failures can be attributed.}
{\looseness=-1 Self-evolution also assumes that reported failures reflect defects of the memory system. To test this, we take the failures that MemPro's loop actually receives, replay each one, and then answer it from its gold evidence (Section~\ref{sec:protocol}, Appendix Table~\ref{tab:actualgold}). Only failures that recover with gold evidence can be traced to memory extraction, memory retrieval, or answer generation. Across all three benchmarks, these attributable root failures are a minority, accounting for 34\% (40/117) on LoCoMo, 26\% (35/136) on Mem-Gallery, and 40\% (10/25) on LongMemEval (Appendix~\ref{app:lme}). On LoCoMo, another 31\% turn correct on replay alone and 34\% stay wrong even with gold evidence. On Mem-Gallery, 21\% turn correct on replay, 38\% stay wrong with gold evidence, and 15\% are adversarial questions without gold evidence (versus 1\% on LoCoMo). Most failures handed to the diagnoser therefore come from sampling flips or from questions that memory edits cannot fix. Whether a gold-unresolved failure counts as wrong depends instead on the answer model, the prompt, and the judge (Appendix~\ref{app:gold}). For example, rejudging fixed MemPro answers with a second judge moves the held-out gain of a selected version from $+2.74$ to $-0.15$ pp (Appendix Table~\ref{tab:judges}, cf.\ \citealp{thakur2025judges}). Yet the loop receives every failure in the same way, so it cannot tell which failures are worth repairing.\par}

{\looseness=-1 In short, the loop acts on noisy single-pass scores and a failure list that mixes replay-correct and gold-unresolved failures with attributable ones. The next two sections separate them. They read repairs and regressions against the flips of an unchanged system and compute $s^*$ from attributable failures only. Section~\ref{sec:intervention} then gives the same distinctions to the evolving agent.\par}

\section{Failure Mode II: Revisions Break Questions They Do Not Target}
\label{sec:local}
\begin{table}[t]
\centering\small
\caption{Repairs, regressions, and score changes of every executed EvolveMem/ARC edit, paired with the parent it modified and grouped by whether it targets that parent's bottleneck stage. Each edit is classed against its own $2\sigma$ re-evaluation band, with the classes defined in Appendix Table~\ref{tab:alignfull}.}
\label{tab:alignment}
\setlength{\tabcolsep}{4pt}
\begin{tabular*}{\linewidth}{@{\extracolsep{\fill}}lcN{00}N{00.0}N{00.0}N{00}ccN{00}N{00}N{00}N{00}@{}}
\toprule
 & & & \multicolumn{2}{c}{Rate (\%)} & \multicolumn{4}{c}{Class against own $2\sigma$ band} & \multicolumn{3}{c}{Score change} \\
\cmidrule(lr){4-5}\cmidrule(lr){6-9}\cmidrule(l){10-12}
System & Aligned & $n$ & Repair & Regr. & Gain & In band & Offset & None & Up & Noise & Down \\
\midrule
EvolveMem     & yes & 14 & 33.7 & 5.7  & 8  & 0 & 3 & 3  & 7  & 3  & 4  \\
              & no  & 33 & 8.4  & 4.7  & 0  & 2 & 9 & 22 & 0  & 22 & 11 \\
\addlinespace[2pt]
ARC closed/r0 & yes & 14 & 40.2 & 3.6  & 11 & 0 & 0 & 3  & 10 & 4  & 0  \\
              & no  & 9  & 13.0 & 5.0  & 0  & 0 & 3 & 6  & 1  & 5  & 3  \\
\addlinespace[2pt]
ARC open/inc  & yes & 10 & 40.6 & 7.1  & 7  & 3 & 0 & 0  & 7  & 2  & 1  \\
              & no  & 4  & 19.7 & 16.2 & 0  & 0 & 2 & 2  & 0  & 1  & 3  \\
\bottomrule
\end{tabular*}
\vspace{-4pt}
\end{table}

\begin{figure}[t]
\centering
\includegraphics[width=\linewidth]{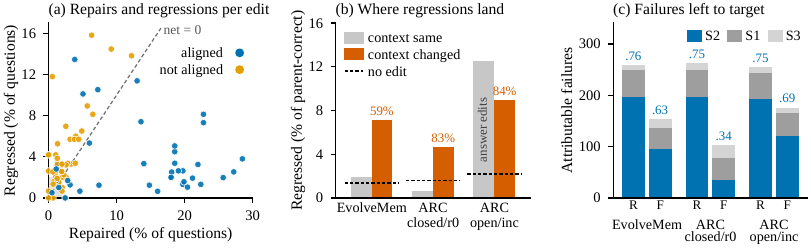}
\vspace{-15.4pt}
\caption{Where the repairs and regressions of the edits in Table~\ref{tab:alignment} fall. (a)~Questions repaired and regressed per edit (\% of evaluated) against the parent version, with the dashed line at zero net change and the grey band at unchanged-system flips. (b)~Regression rate among the parent's correct questions, split by whether the edit changes their retrieved context, with the share changed as labels and the dashed line at an unchanged system evaluated twice. ARC open/inc's unchanged-context regressions come from two answer-prompt edits. (c)~Attributable failures of the root (R) and final versions (F), summed over ten conversations, with the S2 share $\phi$ as labels.}
\label{fig:plane}
\vspace{-8pt}
\end{figure}

{\looseness=-1 Self-evolution presumes that finding why questions fail, and then editing accordingly, raises the score~\citep{reflexion,textgrad,gepa}. Taking the feedback of Section~\ref{sec:feedback} as given, we ask two questions about the revision step: whether the loop's diagnosis and executed edit actually target the primary failure stage, and whether targeting that stage prevents edits from breaking previously correct questions. We examine all executed revisions, accepted or rejected, in EvolveMem, ARC closed/r0, and ARC open/inc (other loops in Appendix~\ref{app:diagnosis}). These three loops record per-round diagnoses and span fixed-root, last-revision, and incumbent reading as well as configuration-only and prompt-plus-configuration edits. Using the bottleneck stage $s^*$ of Equation~\ref{eq:reference} as our \emph{stage reference}, evaluated on $R_{\rm seen}$ for the claim and proposal and on $R_{\rm applied}$ for the executed action, we call an edit \emph{aligned} when it targets the stage reference and \emph{misaligned} otherwise.\par}

\paragraph{Proposals often miss the bottleneck.}
{\looseness=-1 We first find that the diagnosis, the edit, and the stage reference rarely agree~\citep[cf.][]{promptoptimizers}, reflecting three recurring gaps in the loop (Appendix~\ref{app:diagnosis}). First, the diagnoser lacks the provenance defining the stages. Second, proposals drift from the diagnosis, and rejected recipes recur without an attempt history. In EvolveMem, 56 of the 67 proposals miss the bottleneck stage, and 18 of its 34 scored diagnoses give a retrieval explanation but propose an answer-stage change. Third, the diagnoser often inspects a version other than the parent being edited. ARC closed/r0, the original ARC loop, reads the root by design, yet it edits an evolved incumbent in 47 of its 57 rounds, and in 28 of these rounds this stale reading is where the diagnosis first goes wrong (Appendix~\ref{app:diagnosis_funnel}). Many revisions are therefore misdirected before execution even begins.\par}

\paragraph{Aligned edits repair more, but regressions remain.}
{\looseness=-1 We next ask whether hitting the bottleneck stage suffices for reliable gains. When each edit is paired with the parent it modified (Equation~\ref{eq:ledger}), Table~\ref{tab:alignment} shows that alignment raises repairs without reliably reducing regressions. Measured by the repair rate, the share of parent failures fixed, aligned edits repair 34--41\% of their parents' failures, compared with 8--20\% for misaligned ones, and all 26 edits whose repairs and net change both exceed their own binomial $2\sigma$ band (Gain in Table~\ref{tab:alignment}) are aligned (Figure~\ref{fig:plane}a). Read against the loops' own noise floor, 17 of the 28 aligned edits in EvolveMem and ARC closed/r0 (24 of 38 across all three loops) raise the score, compared with 1 of the 42 misaligned ones (1 of 46 overall). Alignment does not reliably lower the regression rate, the share of parent-correct questions broken~\citep[cf.\ negative flips in][]{yan2021pct,echterhoff2024muscle}. Aligned and misaligned edits break 4--6\% of previously correct questions in EvolveMem and ARC closed/r0, so the two groups in Figure~\ref{fig:plane}a separate mainly in repairs, while 14 of the 46 misaligned edits repair beyond noise yet have those repairs offset by regressions (Offset). Targeting the bottleneck thus raises the chance of a gain while leaving its price, the broken questions, largely unchanged.\par}

\paragraph{Shared components extend edits beyond their targets.}
{\looseness=-1 A major reason aligned edits break no fewer questions overall than misaligned ones is that every edit changes a component that all questions share. An edit aims at one class of failures, but what it changes, a retrieval knob, an extraction template, an answer prompt, or pipeline code, serves every question. After the edit, most of the parent's correct questions therefore receive a different retrieved context. As illustrated in Figure~\ref{fig:plane}b, these questions are lost at several times the rate of questions whose context did not change, which break about as often as under re-evaluation alone. Overall, 427 of the 499 regressions fall on changed contexts~\citep[cf.][]{shi2023distracted}. The answer prompt is shared in the same way. For instance, ARC open/inc's two answer-prompt edits leave every context intact, yet they lose 12.5\% of correct questions, and Appendix~\ref{app:cases} traces how a failed aligned edit and misaligned edits break correct answers via the same shared-component routes.\par}

Combining the two effects with evaluation noise, and assuming that repairs concentrate in the targeted pool, the observed change in accuracy of an edit is approximately
\begin{equation}
\Delta\mathrm{Acc}_{\mathrm{obs}}\approx
\underbrace{(1-p)\phi r}_{\text{repairs}}
-\underbrace{ph}_{\text{regressions}}
+\underbrace{(\varepsilon_+-\varepsilon_-)}_{\text{no-edit flips}},
\label{eq:decomposition}
\end{equation}
{\looseness=-1 where $p$ is the parent's accuracy, $\phi$ the targeted share of its failures, $r$ the repair rate within that share, $h$ the regression rate among correct questions, and $\varepsilon_+$, $\varepsilon_-$ the wrong-to-right and right-to-wrong flip rates of an unchanged version. Alignment raises $\phi r$, whereas $h$ comes mainly from editing shared components and is not reliably reduced by alignment. The repair term also depends on the pool of bottleneck failures, and this pool shrinks as the loop repairs them. At the common root, S2 holds three quarters of the attributable failures, so a retrieval upgrade meets a large pool. By each conversation's final version, however, the S2 failures fall by half or more in EvolveMem and ARC closed/r0 and by a third in ARC open/inc, whereas S1 and S3 failures barely change (Figure~\ref{fig:plane}c). The repair term thus shrinks with its pool while the regression term stays, eroding net gains and requiring the loop to verify and curb regressions alongside repairs at the question level (Section~\ref{sec:intervention}, cf.\ \citealp{xu2026aggregate}).\par}

\paragraph{Stage information reaches the retrieval upgrade earlier.}
{\looseness=-1 Since loops often miss the bottleneck stage, one might ask whether giving the proposer accurate stage labels would overcome the plateau on its own. To test this, we fix EvolveMem's round 0, guard, acceptance rule, and menu of 12 recipes, and compare the upstream LLM proposer with two others, random orders of the menu and a proposer that probes the incumbent each round and takes the next menu item for its bottleneck stage (Appendix Table~\ref{tab:proposer}). After seven rounds, the stage-guided proposer ends about 4 pts above both others, but its advantage comes from finding the retrieval upgrade first rather than from the later rounds. Likewise, all 17 aligned edits in EvolveMem and ARC closed/r0 that raise the score do so before the parent enables semantic retrieval. Within this menu, stage information thus helps by reaching the retrieval upgrade sooner rather than by extending the gains beyond it. Once that upgrade is in place, further gains depend on repairing more failures without breaking correct questions, whose cumulative balance across rounds we examine next.\par}

\section{Failure Mode III: Repairs Do Not Accumulate Across Rounds}
\label{sec:retention}
{\looseness=-1 Self-evolution presumes that each accepted version keeps the repairs of earlier versions and adds its own~\citep{evolvemem,mempro,dgm}. Once early upgrades shrink the main failure pool, however, unstable evaluations (Section~\ref{sec:feedback}) and shared-component regressions (Section~\ref{sec:local}) compound across rounds. Across six loops and all three benchmarks, we ask whether late revisions still repair more questions than they break, and whether the repairs recorded along an accepted lineage, the chain of versions a run keeps, persist beyond evaluation noise. Pairing every evaluated revision with its actual parent whether accepted or not, Table~\ref{tab:retention} reports two volumes per revision, the share of all evaluated questions that it repairs and the share that it breaks, on one common scale across the early (E) and late (L) halves of each run (counts and F1-scored rows in Appendix Table~\ref{tab:fulltrends}). As a baseline, we compare both volumes with the flips of an unchanged system between two evaluations, which we call no-edit flips.\par}

\begin{table}[t]
\centering\small
\caption{Repair and regression volumes of early and late revisions, against the no-edit flips of an unchanged system. The same six methods on three datasets, each scored by the metric it selects on. Entries are the questions repaired or regressed per revision against its actual parent, in pp of evaluated questions.}
\label{tab:retention}
\setlength{\tabcolsep}{2.4pt}\footnotesize
\begin{tabular*}{\linewidth}{@{\extracolsep{\fill}}lN{00.0}ccccc@{\hspace{8pt}}cccccc@{\hspace{8pt}}N{00.0}ccccc@{}}
\toprule
 & \multicolumn{6}{c}{LoCoMo} & \multicolumn{6}{c}{Mem-Gallery} & \multicolumn{6}{c}{LongMemEval} \\
\cmidrule(lr){2-7}\cmidrule(lr){8-13}\cmidrule(l){14-19}
 & \multicolumn{2}{c}{Repair} & \multicolumn{2}{c}{Regr.} & \multicolumn{2}{c}{No-edit} & \multicolumn{2}{c}{Repair} & \multicolumn{2}{c}{Regr.} & \multicolumn{2}{c}{No-edit} & \multicolumn{2}{c}{Repair} & \multicolumn{2}{c}{Regr.} & \multicolumn{2}{c}{No-edit} \\
\cmidrule(lr){2-3}\cmidrule(lr){4-5}\cmidrule(lr){6-7}\cmidrule(lr){8-9}\cmidrule(lr){10-11}\cmidrule(lr){12-13}\cmidrule(lr){14-15}\cmidrule(lr){16-17}\cmidrule(l){18-19}
System & E & L & E & L & Rep. & Reg. & E & L & E & L & Rep. & Reg. & E & L & E & L & Rep. & Reg. \\
\midrule
EvolveMem & 5.0 & 2.9 & 2.4 & 3.3 & 1.0 & 1.0 & 3.5 & 2.2 & 1.3 & 2.7 & 1.0 & 0.7 & 7.2 & 3.9 & 7.0 & 4.9 & 1.8 & 4.0 \\
ARC closed/r0 & 8.7 & 1.2 & 1.8 & 2.3 & 1.1 & 1.3 & 7.2 & 1.0 & 1.3 & 2.5 & 0.3 & 0.4 & 6.6 & 1.3 & 2.8 & 1.3 & 1.4 & 2.0 \\
ARC closed/inc & 10.9 & 2.8 & 2.7 & 5.5 & 0.6 & 0.8 & 6.8 & 1.2 & 1.6 & 2.1 & 0.5 & 0.0 & 8.5 & 4.7 & 5.1 & 4.7 & 1.7 & 3.0 \\
ARC open/inc & 7.6 & 2.2 & 3.7 & 3.0 & 1.2 & 1.9 & 4.4 & 1.8 & 1.4 & 2.4 & 0.6 & 0.5 & 11.2 & 5.0 & 7.4 & 5.2 & 1.5 & 2.8 \\
MemPro & 6.7 & 4.7 & 6.1 & 5.6 & 6.0 & 6.1 & 4.7 & 4.0 & 4.5 & 3.9 & 2.3 & 4.1 & 7.3 & 3.9 & 4.6 & 4.2 & 3.3 & 4.0 \\
GEPA & 6.6 & 6.3 & 6.0 & 7.0 & 5.5 & 6.7 & 5.3 & 4.0 & 4.3 & 2.9 & 2.3 & 4.1 & 6.3 & 4.2 & 7.0 & 7.5 & 3.3 & 4.0 \\
\bottomrule
\end{tabular*}
\vspace{-14pt}
\end{table}

\paragraph{Repair volumes shrink while regressions persist.}
{\looseness=-1 Table~\ref{tab:retention} shows that repair volume falls from early to late revisions in all eighteen settings, six methods on three datasets. Regression volume, in contrast, falls less or not at all in all but one setting, so late revisions break at least as many questions as they repair in sixteen of the eighteen settings. Equation~\ref{eq:decomposition} predicts this pattern once the early retrieval upgrades have drained much of the bottleneck-stage pool, shrinking the repair term $(1-p)\phi r$ while the regression term $ph$ from shared components persists. To separate evaluation noise, shrinking repairs, and regressions at the plateau, we subtract the no-edit flip rates $\varepsilon_+$ and $\varepsilon_-$ from each revision's flips (Appendix~\ref{app:retention}, Table~\ref{tab:steppartition}). In EvolveMem on LoCoMo, noise accounts for a third of late flips, repairs above noise fall from 4.0 to 1.9 pp, and regressions above noise rise from 1.3 to 2.3 pp. Late revisions thus continue to fix some failures, but concurrent regressions offset those gains.\par}

\paragraph{Many recorded repairs are only noise.}
{\looseness=-1 Compounding this imbalance, late repair volumes drop toward the no-edit flips (Table~\ref{tab:retention}), so late revisions often turn no more failures correct than re-evaluating their unchanged parent would. On LoCoMo, for instance, late ARC closed/r0 revisions repair about as many questions as its unchanged configuration flips between two evaluations, MemPro falls below that level, and GEPA stays near it throughout. In this regime, a single turn from wrong to correct is weak evidence of a repair. Of the questions that an unchanged system turns correct on re-evaluation, about half turn wrong again on a third evaluation (Appendix~\ref{app:retention}). Likewise, when parent and child versions in MemPro are each evaluated three times, only 4 of 98 late recorded repairs go from wrong in all three parent evaluations to correct in all three child evaluations, while the rest are not stably reproduced (Appendix~\ref{app:retention}, Table~\ref{tab:steppartition}). Followed along the accepted lineage, newly repaired questions consequently fail at the next step far more often than questions that were already correct (Appendix Table~\ref{tab:nextloss}), creating turnover much larger than the net gain. Along MemPro's LoCoMo lineages, a single step flips about as many held-out questions as a root re-evaluation and roughly a third change state at least once (Appendix Table~\ref{tab:churn}), yet each lineage ends 3--6 pp above its root. Although MemPro's net gain is positive on every evaluated split (Appendix Table~\ref{tab:cumulative}), much of what the loop records as repair is the same flipping that re-evaluating the unchanged parent would produce, and the loop counts it as progress.\par}

{\looseness=-1 In short, repairs shrink toward the level of no-edit flips as the rounds proceed, while regressions persist. The loops see neither effect. They evaluate each version once, count every turn from wrong to correct as a repair, and see the regressions of an edit only through its net change. Section~\ref{sec:intervention} gives MemPro's loop these two counts and lets it accept a revision only when the repairs outweigh the regressions.\par}

\section{Addressing the Three Failure Modes}
\label{sec:intervention}
{\looseness=-1 Having located losses in evaluation, revision, and retention, we ask whether intervening at each step of the loop (Equation~\ref{eq:loop}) reduces those losses and stabilizes evolution. We add one mechanism per failure mode to MemPro, whose open code edits and judge selection show high noise and turnover in Tables~\ref{tab:noise} and~\ref{tab:retention}. All three read the same per-question labels. To obtain them, the harness evaluates every new version twice and replays its failures at temperature zero. A question is \emph{fragile} when the two evaluations disagree, \emph{replay-correct} when both fail but the replay succeeds, and stable otherwise. For evaluation (I), \emph{verified feedback} gives the agent a ledger of these labels, so that it can tell stable failures from fragile ones. For revision (II), a \emph{repair round} shows a separate session only the questions that a revision repaired and regressed and asks it to narrow the change. For retention (III), \emph{paired acceptance} (Algorithm~\ref{alg:accept}) lets the harness keep a revision only if its stable repairs outnumber its stable regressions against the incumbent~\citep[cf.][]{blum2015ladder,pace}.\par}

{\looseness=-1 Across ten-round runs from the same root on the 154-question development split (Appendix~\ref{app:interventions}), we compare unmodified MemPro (E0, four runs) with four intervention arms. To test verified feedback alone, E1 provides the ledger (three runs), while E2 adds the failure stages of Section~\ref{sec:protocol} in an exploratory run. Adding repair rounds and paired acceptance yields E1 + repair (three runs) and E2 + repair (one run) to curb regressions and unstable flips. For each run, we report the development accuracy of the round-10 incumbent, the mean of its two evaluations, and its held-out gain, the paired change over the root on the 1,386 held-out questions, both in judge accuracy.\par}

\begin{table}[t]
\centering\small
\caption{Revisions and incumbents of the five arms on MemPro after ten rounds (seed 2028). E1 gives the agent the ledger, E2 adds the failure stages, and +repair adds repair rounds and paired acceptance. Targeting columns give the share of the agent's targets that are stable failures and the share of a child's newly correct questions that were fragile or replay-correct in its parent. Regression columns count stable regressions per revision before and after its repair round. Dev.\ is the mean of two evaluations on the 154 development questions (root 75.97) and $\Delta$HO the paired gain over the root on 1,386 held-out questions in pp, both averaged over the runs of an arm (per-run values in Appendix Table~\ref{tab:armruns}). Targeting, regressions, and the Dev.\ of E0 come from the runs whose versions were labeled.}
\label{tab:arms}
\setlength{\tabcolsep}{3pt}
\begin{tabular*}{\linewidth}{@{\extracolsep{\fill}}l*{6}{c}@{}}
\toprule
 & \multicolumn{2}{c}{Targeting (\%)} & \multicolumn{2}{c}{Stable regr.\ per revision} & \multicolumn{2}{c}{Incumbent, round 10} \\
\cmidrule(lr){2-3}\cmidrule(lr){4-5}\cmidrule(l){6-7}
Arm & Stable targets & False repairs & Before & After & Dev. & $\Delta$HO \\
\midrule
E0 (MemPro) & 69  & 52  & --- & --- & 86.69 & $+3.79$ \\
E1 (+ ledger) & 90  & 64  & --- & --- & 85.50 & $+2.60$ \\
E2 (+ failure stages) & 96  & 71  & --- & --- & 89.29 & $+2.61$ \\
\addlinespace[2pt]
E1 + repair & --- & --- & 2.5 & 1.4 & 92.31 & $+3.50$ \\
E2 + repair & --- & --- & 3.0 & 2.7 & 89.29 & $+3.05$ \\
\bottomrule
\end{tabular*}
\vspace{-14pt}
\end{table}

{\looseness=-1 As shown in Table~\ref{tab:arms}, each mechanism reduces the in-loop loss it targets. With the ledger, the agent turns to stable failures, which make up 90--96\% of its targets in E1 and E2 against 69\% in E0. Feedback verification alone, however, leaves acceptance exposed to evaluation flips. Across E0--E2, 52--71\% of the questions that a revision newly answers correctly in a single evaluation were fragile or replay-correct in its parent (False repairs in Table~\ref{tab:arms}). Paired acceptance curbs false repairs by excluding fragile questions and requiring a child to pass both evaluations for a repair, and the reverse for a regression (Algorithm~\ref{alg:accept}). Meanwhile, repair rounds reduce the stable regressions of each revision in every labeled run, from 2.5 to 1.4 in E1 + repair (Appendix~\ref{app:interventions}). The losses located in Sections~\ref{sec:feedback}--\ref{sec:retention} can thus be reduced within the loop.\par}

{\looseness=-1 With repair rounds and paired acceptance, end-to-end runs also become much more reproducible. E1 + repair reaches the highest development accuracy of the five arms, 92.31 against 86.69 for E0, with an SD of 0.49 across its three runs, and its held-out gain has an SD of 0.29 pp across runs against 2.00 for the four E0 runs. Across the five arms, development accuracy rises by 10 to 16 pp over the root. In E0--E2, selection uses only the first evaluation and the second stays well above the root, whereas in the repair arms both enter acceptance (Appendix~\ref{app:interventions}), yet in Table~\ref{tab:arms}, every arm gains about 3--4 pp on held-out questions, and the differences of E1 and E1 + repair from E0 are not significant, with 95\% intervals below $+3$ pp. To check whether E1 + repair's stability comes simply from evaluating twice, we run five matched controls per arm under one local task model and judge (Appendix~\ref{app:interventions}, Table~\ref{tab:armruns}). Evaluating unmodified MemPro twice and selecting by the mean ($+2.59\pm2.12$ pp) improves neither held-out gain nor variance over single-evaluation E0 ($+3.24\pm1.40$ pp), whereas E1 + repair gains $+4.76\pm0.84$ pp, though two-sided 95\% intervals still include zero.\par}

{\looseness=-1 It remains open why the larger development gains carry over to held-out questions only in part, beyond any held-out advantage smaller than the roughly 3 pp that 154-question runs can detect. Two explanations fit our data. First, once noisy flips are filtered out, most remaining development failures cannot be attributed to the memory system, as only 5 of the root's 20 stable failures turn correct when answered from their gold evidence (Appendix Table~\ref{tab:resolution}). Repairing them may take changes that suit these particular questions rather than the memory system. Second, late accepted steps change only one or two stable questions, within the variation of the development set itself (Appendix~\ref{app:bigsplit}, Table~\ref{tab:resolution}). Testing this resolution limit directly by doubling the development set to 308 questions with repeated evaluations lets unmodified MemPro raise held-out accuracy by nearly 6 pp over the root, yet only 2 of its 15 revisions exceed this variation while the other 13 stay within noise. A larger development set thus lets major steps come through, whereas smaller late steps remain hard to select. We leave it to future work to combine our mechanisms with larger or withheld selection sets and to check whether held-out gains come from the same kinds of failures as development gains.\par}

\section{Related Work}
\label{sec:related}
{\looseness=-1 \textbf{Memory systems for LLM agents} keep agent state outside the context window~\citep{memgpt,mem0,amem,zep}. Recent analyses decompose benchmark scores by module to locate the bottleneck of a fixed system~\citep{agentnative,retrievalutil,memeye}, whereas we locate it on every version of a system that changes. \textbf{Self-improving agents} turn feedback from their own runs into lasting changes~\citep{selfevolvesurvey}, and memory systems now evolve their skills, architecture, retrieval configuration, or whole program~\citep{memskill,memevolve,evolvemem,mstar,mempro}. Studies of this loop question its gains~\citep{harnessupdating,harnessrethink} and show that accepting every score increase on a noisy development set commits false edits~\citep{pace}, but they compare aggregate scores. A release pipeline for agents gates each new version on the cases it repairs and breaks~\citep{agentdevel}, and we track the same two sets across every version a loop accepts, together with whether the repairs survive. Since language models localize errors poorly from traces~\citep{tyen,whowhen}, recent work on \textbf{failure attribution} relies on interventions such as counterfactual replay, fault injection, and edit-and-rerun~\citep{agentracer,whowhenpro,dover}, and memory-specific tools attribute failures to memory operations or stages~\citep{memtrace,d2acci,mimemory}. These methods score attribution on fixed traces, whereas we use intervention labels as the reference for the diagnoses, proposals, and edits of a running loop. Appendix~\ref{app:related} discusses each line in more detail.\par}

\section{Conclusion}
{\looseness=-1 Self-evolving memory systems typically gain early and then level off. We traced where along their loop the gains are lost, with per-question outcomes and interventions that do not depend on the agent's own account. In evaluation, evaluating an unchanged version again alters its set of failed questions, and many failures either vanish on replay or persist with the annotated evidence. In revision, targeting the bottleneck stage raises repairs without reducing regressions from shared components. In retention, repairs shrink across rounds while regressions persist, and much of what the loop records as repair would also appear on re-evaluating the unchanged version. Verified feedback, repair rounds, and paired acceptance reduce these losses in MemPro and raise held-out accuracy with lower run-to-run variation. These findings suggest that a self-evolving system should evaluate each revision more than once and weigh its repairs against its regressions at the question level. Whether held-out gains come from the same kinds of failures as development gains is the natural next test.\par}

\label{main_end}
\clearpage
\section*{Reproducibility statement}
Section~\ref{sec:protocol} and Appendix~\ref{app:setup} specify the loops we study, the models in each role, the benchmarks and splits, and the measurement conventions. Appendix~\ref{app:probes} gives the details of the failure-stage probes and the acceptance rule of the intervention arms (Algorithm~\ref{alg:accept}), Appendix~\ref{app:prompts} the prompts of our measurements and of the arms, and Appendix~\ref{app:tables} the per-method tables behind every pooled number of the main text. We will release the harness code, the analysis scripts, and the question-level records of every evaluated version.
\section*{AI use statement}
Generative AI tools were used to assist with language revision and manuscript polishing. In particular,
AI assistance was used to improve the presentation of the paper. All AI-assisted manuscript revisions
were reviewed and verified by the authors, who take responsibility for the final content.
\bibliography{references}
\bibliographystyle{iclr2027_conference}
\clearpage
\appendix
\renewcommand{\contentsname}{Appendix Contents}
\addtocontents{toc}{\protect\setcounter{tocdepth}{2}}
{\small\tableofcontents}
\section{Extended Related Work}
\label{app:related}
\paragraph{Memory systems for LLM agents.}
Memory systems keep agent state outside the context window through tiered context~\citep{memgpt}, extraction and update pipelines~\citep{memorybank,mem0,memoryos}, linked notes~\citep{amem}, or knowledge graphs~\citep{zep}. Benchmarks such as LoCoMo~\citep{locomo}, LongMemEval~\citep{longmemeval}, and Mem-Gallery~\citep{memgallery} score final answers, and recent analyses decompose these scores by module~\citep{agentnative,retrievalutil,memeye}. These analyses locate the bottleneck of a fixed system, whereas we label failures by intervention on every version of a changing system and track how an edit to a shared component affects the questions it did not target (Sections~\ref{sec:protocol} and~\ref{sec:local}).

\paragraph{Self-improving agents.}
Self-improving agents turn feedback from their own runs into lasting changes~\citep{selfevolvesurvey}, from reflections and skills~\citep{reflexion,voyager} to prompts~\citep{gepa}, workflows~\citep{aflow}, harness code~\citep{metaharness}, and their own source~\citep{dgm}, and memory systems now evolve their skills~\citep{memskill}, architecture~\citep{memevolve}, retrieval configuration~\citep{evolvemem}, or whole program~\citep{mstar,mempro}, with one multimodal memory system discovered end to end by such a loop~\citep{omnimem}. Studies of the loop itself find that evolver capability barely changes harness gains~\citep{harnessupdating}, that evolved harnesses do not consistently beat test-time scaling at matched budgets~\citep{harnessrethink}, and that accepting every score increase on a noisy development set commits false edits~\citep{pace}, a known selection effect~\citep{cawley,siren}. These studies compare aggregate scores. AgentDevel gates each new version on the cases it repairs and breaks~\citep{agentdevel}, and we further check whether these repairs hold on re-evaluation and survive later rounds (Sections~\ref{sec:retention} and~\ref{sec:intervention}).

\paragraph{Failure attribution.}
Language models localize errors poorly from traces~\citep{tyen}, and the best method on Who\&When finds the decisive step of a multi-agent failure in 14.2\% of cases~\citep{whowhen}. Recent work therefore grounds attribution in interventions such as counterfactual replay, fault injection, and edit-and-rerun~\citep{agentracer,whowhenpro,dover}, and memory-specific tools trace operation graphs~\citep{memtrace} or gate interventions on stage-level traces~\citep{d2acci,mimemory}. In prompt optimization, reflecting on pseudo errors or wrong reference answers usually yields comparable prompts~\citep{promptoptimizers,textualgradients}. These works score attribution on fixed traces, whereas we use replay and gold-evidence labels as the reference inside a running loop and score the diagnosis, proposal, and executed edit against the version each read or modified (Section~\ref{sec:local}).
\section{Notation and Experimental Setup}
\label{app:setup}
This section collects the notation of Section~\ref{sec:protocol}, the four methods and their configurations, the model in each role, the datasets and splits, and the measurement conventions that the tables of the paper follow.

\subsection{Notation}
\label{app:notation}
Table~\ref{tab:notation} lists the symbols and names used in the main text and in Algorithm~\ref{alg:accept}.

\begin{table}[htbp]
\centering\small
\caption{Notation.}
\label{tab:notation}
\begin{tabularx}{\linewidth}{@{}lY@{}}
\toprule
Symbol & Meaning \\
\midrule
$R$, $R'$ & A version of the memory system, and the revision that an edit produces from it \\
$R_{\rm seen}$, $R_{\rm applied}$ & The version whose evaluation the diagnoser reads, and the parent that the edit actually modifies \\
root, lineage & The initial version, and the chain of versions that a run accepts from the root to its final version \\
incumbent, parent, revision, proposal & The version the loop currently keeps, the version an edit modifies, the version an edit produces, and the change the diagnosis proposes \\
$Q$, $q$ & The development questions of a run, and one question \\
$y^{(k)}_q(R)\in\{0,1\}$ & Outcome of $q$ on $R$ in evaluation $k$, namely nonzero F1 or the judge's verdict \\
$\mathcal{E}(R)$, $D$, $A$ & The evaluation output of $R$, the diagnoser, and the executor (Eq.~\ref{eq:loop}) \\
claim, proposal, action & The stage the diagnosis names, the change it proposes, and the executed diff \\
replay-correct, gold-unresolved & A failure that turns correct on replay, and a failure that stays wrong when answered from its gold evidence \\
$\mathrm{stage}(q)\in\{\mathrm{S1},\mathrm{S2},\mathrm{S3}\}$ & The probe-attributed stage of a failure, namely memory extraction, memory retrieval, or answer generation (Section~\ref{sec:protocol}) \\
$s^*(R)$ & The bottleneck stage of $R$, which holds the most attributable failures (Eq.~\ref{eq:reference}) \\
$\fixed$, $\broken$ & The questions an edit repairs and breaks, respectively (Eq.~\ref{eq:ledger}) \\
$2\sigma$ band & Twice the standard deviation of the score over re-evaluations of an unchanged version \\
$p$, $\phi$, $r$, $h$ & Parent accuracy, share of failures at the bottleneck stage, repair rate there, and regression rate among correct questions (Eq.~\ref{eq:decomposition}) \\
fragile, stable-correct, stable-wrong & Classes of a question under two evaluations and a temperature-0 replay of one version (Algorithm~\ref{alg:accept}) \\
$f$, $b$, $z$ & Stable repairs, stable regressions, and $(f-b)/\sqrt{f+b}$ in paired acceptance (Algorithm~\ref{alg:accept}) \\
closed, open & ARC action spaces, editing the configuration only or also the extraction and answer prompts \\
r0, inc & The version that ARC's diagnoser reads, the root or the incumbent \\
E, L & The first and the second half of a run's evaluated revisions in production order \\
E0, E1, E2 & Intervention arms, namely unmodified MemPro, the loop with the ledger, and the loop with the ledger and failure stages \\
pp, pts & Percentage points, and points of the mixed F1 score of EvolveMem and ARC \\
\bottomrule
\end{tabularx}
\end{table}

\subsection{Loops and configurations}
\label{app:loops}
Table~\ref{tab:systems} summarizes the four methods. The paragraphs below define the ARC configurations and list where our runs depart from the released systems.

\begin{table}[htbp]
\centering\footnotesize
\caption{The loops we study. What each loop edits, which version its diagnoser reads, how it selects the next incumbent, and what it evolves on. All loops run on LoCoMo and on caption-only Mem-Gallery.}
\label{tab:systems}
\setlength{\tabcolsep}{4pt}
\begin{tabularx}{\linewidth}{@{}lYYlYc@{}}
\toprule
System & Edited object & Diagnoser reads & Selection & Evolves on & Rounds \\
\midrule
EvolveMem & Retrieval configuration of a fixed pipeline & Last evaluated revision & Mixed F1 score & All questions of a conversation & $\leq$7 \\
ARC closed/r0 & Configuration only & Root & Mixed F1 score & All questions of a conversation & $\leq$10 \\
ARC closed/inc & Configuration only & Incumbent & Mixed F1 score & All questions of a conversation & $\leq$10 \\
ARC open/r0 & Configuration, extraction and answer prompts & Root & Mixed F1 score & All questions of a conversation & $\leq$10 \\
ARC open/inc & Configuration, extraction and answer prompts & Incumbent & Mixed F1 score & All questions of a conversation & $\leq$10 \\
MemPro & Its own memory program, code and prompts & Parent it edits & Judge accuracy & Random 10\% of the questions & 10--15 \\
GEPA & Five prompts of MemPro's initial pipeline & Incumbent's traces & Judge accuracy & The same 10\% & 15 passes \\
\bottomrule
\end{tabularx}
\tabnote{Open ARC configurations rebuild the memory bank whenever the extraction template changes and run with the execution guard on. EvolveMem's released loop is elitist, so its diagnoser reads the revision evaluated last, accepted or not, while the proposal is applied to the incumbent. MemPro and GEPA hold out the remaining 90\% of the questions.}
\end{table}

\paragraph{Configurations.}
We study EvolveMem, ARC, MemPro, and GEPA. EvolveMem tunes the retrieval configuration of a fixed pipeline, MemPro rewrites its own memory program, and GEPA optimizes five prompts of MemPro's initial pipeline. ARC runs in eight configurations that differ in four choices. Its action space is closed when the agent may edit only the configuration, and open when it may also edit the extraction and answer prompts. Its diagnoser reads the root under r0, as in the released ARC loop, and the incumbent under inc. The open configurations come in three versions, one that appends re-extracted memories to the existing bank and two that rebuild the bank whenever the extraction template changes, with the execution guard on or off. Together with closed/r0 and closed/inc, these make eight configurations. Unless noted otherwise, open/r0 and open/inc denote the guarded rebuild versions.

ARC reports Omni-SimpleMem~\citep{omnimem} as the memory system that its loop discovered, but the release contains only the final system, without its naive starting point or its iteration records. We therefore run ARC's released loop on EvolveMem's weak initial pipeline, whose every version we control and record. We also run the loop on the released Omni-SimpleMem (Appendix~\ref{app:omni}).

\paragraph{Departures from the released systems.}
EvolveMem and MemPro run their released code, and settings not listed here follow the original papers. EvolveMem starts from the initial configuration of its paper, BM25 retrieval only with $k=5$ and a context budget of 8, and runs round 0 plus at most seven evolution rounds, as in its reported runs~\citep{evolvemem}. It keeps the acceptance rule of the released code, an elitist loop that differs from Algorithm 1 of the paper, and we rerun that Algorithm 1 from the same round 0 in Table~\ref{tab:fulltrends}. Our harness replaces GPT-4o with Qwen3-max, fixes the model of each role, disables the LoCoMo-specific answer flags that the diagnoser could otherwise switch on, and records provenance during extraction (Appendix~\ref{app:provenance}).

MemPro runs the released evaluation, with task temperature 0.3, at most three research iterations, and the released judge prompt, on the paper's task model Qwen3-30B-A3B-Instruct-2507. As in the paper, it evolves on a random 10\% of LoCoMo and selects by judge accuracy~\citep{mempro}. The release contains no split, so we draw the splits with our own seeds. MemPro departs from the paper in three settings. Its evolving agent is kimi-k3, and its judge is Qwen3-max instead of GPT-4o-mini. Its task model decodes at the released temperature of 0.3 rather than at the paper's temperature 0.7 and top-p 0.9. At the paper's setting, the root scores 80.00\% on all 1,540 questions, against 78.25\% and 78.64\% in two evaluations at 0.3. The released root already contains evolved components and scores 78.50\% on held-out questions, above the 72.35\% of the paper's initial framework. Its held-out gain after ten rounds is nonetheless close to that of the paper, at $+3.8$ against $+4.7$ pp. GEPA follows MemPro's baseline setting. It updates the five prompts of MemPro's initial pipeline on the same development questions, with kimi-k3 as reflection model, Qwen3-30B-A3B as task model, and fifteen passes over the questions.

\subsection{Models and roles}
\label{app:models}
Table~\ref{tab:models} lists the model in each role of the loops and of our measurements. The main EvolveMem and ARC runs use Qwen3-max in every role, and in Appendix~\ref{app:substitutions} we substitute GPT-4o or GPT-4o-mini for the diagnoser, the answer model, and the ARC agent.

\begin{table}[htbp]
\centering\small
\caption{Models by role. Roles of the loops are listed first and roles of our measurements second.}
\label{tab:models}
\begin{tabularx}{\linewidth}{@{}lYY@{}}
\toprule
Role & Loops & Model \\
\midrule
Extraction and answer model & EvolveMem, ARC, Omni-SimpleMem & qwen3-max (GPT-4o-mini answers in Appendix~\ref{app:substitutions}) \\
Diagnoser or agent & EvolveMem, ARC, Omni-SimpleMem & qwen3-max (GPT-4o and GPT-4o-mini in Appendix~\ref{app:substitutions}) \\
Evolving agent & MemPro; GEPA reflection & kimi-k3 \\
Task model & MemPro, GEPA & Qwen3-30B-A3B-Instruct-2507 \\
Selection judge & MemPro, GEPA & qwen3-max \\
\midrule
Probe solver & EvolveMem, ARC stage probes & qwen3-max (GPT-4o-mini in Appendix~\ref{app:labels}) \\
Gold-evidence probe answer models & All & qwen3-max, qwen3-30b-a3b, gpt-4o-2024-11-20, gpt-4o-mini \\
Stage annotator of claims and diffs & EvolveMem claims; MemPro and GEPA diffs & qwen3.8-max, two samples \\
Second judge for fixed-answer rejudging & MemPro held-out answers & GPT-4o-mini \\
\bottomrule
\end{tabularx}
\end{table}

\paragraph{Scoring.}
EvolveMem and ARC select versions by a mixed F1 score. On LoCoMo it combines the F1 of categories 1--4 with the refusal score of category 5 and weights every conversation equally, and on Mem-Gallery it is the official F1. MemPro and GEPA select by judge accuracy. Our question-level analyses need a binary outcome. For EvolveMem and ARC we count an answer as correct when its category 1--4 F1 is nonzero, which is a lenient criterion rather than a check of semantic correctness. For MemPro and GEPA we use the judge's verdict. Fixed-answer rejudging applies a second judge to the same answers. On Mem-Gallery we report the scenario mean and the question-weighted aggregate separately.

\subsection{Datasets and splits}
\label{app:data}
Table~\ref{tab:datasets} gives the size of the three benchmarks and of their development and held-out splits, and Table~\ref{tab:scale} the scope of the measurements of each method. The paragraphs below describe each benchmark and the query-image comparison on Mem-Gallery.

\begin{table}[htbp]
\centering\small
\caption{Datasets and splits. Development questions are drawn at random within each dataset, and held-out questions are the rest. LoCoMo development and held-out questions are disjoint but share conversation histories, and Mem-Gallery splits share scenarios in the same way. LongMemEval development questions are drawn within each question type, and every question has its own history.}
\label{tab:datasets}
\setlength{\tabcolsep}{4pt}
\begin{tabularx}{\linewidth}{@{}lYccN{0,000}@{}}
\toprule
Dataset & Units & Questions & Development & Held-out \\
\midrule
LoCoMo & 10 conversations & 1,540 (categories 1--4) & 154 (seeds 2026--2029) & 1,386 \\
LoCoMo, larger split & 10 conversations & 1,540 & 308 (stratified) & 1,232 \\
Mem-Gallery, caption-only & 20 scenarios & 1,711 & 171 (seeds 2026--2029) & 1,540 \\
LongMemEval\_S & 500 histories of about 48 sessions & 500 & 100 (seed 2028) & 400 \\
\bottomrule
\end{tabularx}
\tabnote{EvolveMem, ARC, and Omni-SimpleMem evolve on all questions of one conversation or scenario and are evaluated in sample, with conversation-level transfer in Table~\ref{tab:noise}b. MemPro and GEPA evolve on the development split and are scored on the held-out split. On LongMemEval, every method does so.}
\end{table}

\begin{table}[htbp]
\centering\small
\caption{Scope of the measurements by method. Each count is in the unit that its row names.}
\label{tab:scale}
\begin{tabularx}{\linewidth}{@{}lY@{}}
\toprule
Setting & Recorded scope \\
\midrule
EvolveMem & 10 conversations, up to 8 rounds; 77 evaluations and 3,981 failed question--version records \\
ARC & 10 conversations, up to 10 rounds, eight configurations; 361 rounds \\
MemPro & Trees of three splits (seeds 2027--2029); 45 diagnoses and 1,234 failed question--version records \\
Same-input evaluation pairs & 4,152 EvolveMem; 6,410 ARC; 462 MemPro \\
GEPA & Four LoCoMo splits and Mem-Gallery seed 2026 \\
\bottomrule
\end{tabularx}
\end{table}

\paragraph{LoCoMo.}
LoCoMo~\citep{locomo} consists of ten long two-speaker conversations, each spanning 19 to 32 timestamped sessions. Across these conversations, the benchmark contains 1,540 non-adversarial questions in categories 1--4, namely single-hop, multi-hop, temporal, and open-domain questions, together with 446 adversarial abstention questions in category 5. EvolveMem and ARC build one memory per conversation and evolve on all its questions under their mixed F1 score. Our question-level analyses and the splits of MemPro and GEPA instead use categories 1--4, taking each question as the unit of evaluation (Table~\ref{tab:datasets}).

\paragraph{Mem-Gallery.}
Mem-Gallery~\citep{memgallery} extends this multi-session conversational setting to multimodal histories across twenty scenarios totaling 240 sessions. Its 1,711 questions span nine categories covering factual, temporal, visual, knowledge-update, and adversarial refusal questions, and all nine enter evaluation and the splits of Table~\ref{tab:datasets}. In the caption-only setting of the main text, images in the dialogue history and in the queries are represented by the captions supplied with the dataset, so that text-based memory pipelines apply directly.

\paragraph{LongMemEval.}
\label{app:lme}
LoCoMo and Mem-Gallery share one format, in which many questions are asked about a few long conversations. LongMemEval\_S~\citep{longmemeval} differs in this respect: each of its 500 questions has its own history of about 48 sessions, so it tests whether the findings of Sections~\ref{sec:feedback}--\ref{sec:retention} depend on the conversational format. We draw 100 development questions at random within each question type and hold out the other 400, so that development and held-out questions share no history. Every method evolves on the same development questions, and MemPro and GEPA are scored on the held-out split as on the other two benchmarks. EvolveMem and the four ARC configurations build one memory per question and evolve one configuration shared by all of them. We use Qwen3-30B-A3B for extraction, answering, and the stage probes, and Qwen3.7-max as the diagnoser or agent. MemPro and GEPA keep the models of Table~\ref{tab:models}, with Qwen3.7-max as the judge. We run each loop four times from the same root, and GEPA once.

\paragraph{Mem-Gallery query images.}
\label{app:galleryvlm}
The caption-only runs of Sections~\ref{sec:feedback}--\ref{sec:retention} replace each query image with the dataset caption, and 487 of the 1,711 questions carry one, so we test whether a finding depends on the substitution. To this end we run EvolveMem twice on Mem-Gallery from the same initial memory, once with the query image in the answer model's input and once with the caption in its place. The two arms differ in this one choice: both answer with Qwen3-VL-Plus, both diagnose and re-extract with Qwen3.7-max, both are scored by qwen-max, and both keep the protocol, the twenty scenarios, rounds 0--7, and the selection rule of the caption-only runs. The remaining 1,224 questions receive identical input in the two arms, so any difference between the arms there comes from the sampling of the two runs.

\subsection{Measurement conventions}
\label{app:conventions}
The conventions below fix which revisions enter each analysis, what a repetition holds fixed, and why we take the re-evaluation of an unchanged version, rather than temperature-0 decoding, as the noise reference.

\paragraph{Revision populations.}
Every edit is paired with the parent it actually modifies. All executed revisions enter the analysis of revision in Section~\ref{sec:local}, whether the loop accepted them or not. The repair and regression volumes of Section~\ref{sec:retention} likewise count every evaluated revision against its actual parent, whereas only accepted revisions enter its next-step and lineage analyses (Tables~\ref{tab:nextloss} and~\ref{tab:churn}). For the early and late halves of Tables~\ref{tab:retention}, \ref{tab:fulltrends}, and~\ref{tab:judgetrends}, we split the evaluated revisions of each run in production order and count an odd extra revision as late. A run is one conversation or scenario for EvolveMem and ARC and one split for MemPro and GEPA. Configuration and memory hashes separate steps that leave the system unchanged from actual updates. Under this rule, two ARC closed/r0 steps that change only category overrides count as actual updates.

\paragraph{Repetitions.}
We repeat measurements at two levels. A fixed-version re-evaluation freezes the configuration, memory, prompts, and questions and regenerates only the outcomes. Table~\ref{tab:noise}a applies it to the version that each loop selected, with three further evaluations. Re-evaluating MemPro's three roots once at the released temperature covers 462 question pairs. It turns 39 of their 117 failures correct and 22 of their 345 correct answers wrong, a net change of $+3.7$ pp at a flip rate of 13.2\%. An evolution repetition instead reruns the whole loop. It freezes the starting system, the initial memory, the questions, and the loop settings, and draws new proposals and executions. Each loop has four such repetitions. For EvolveMem and the four ARC configurations of Table~\ref{tab:noise}b, the standard deviation of a conversation's endpoint score across repetitions is 1.3--3.3 pts, compared with 0.4--0.5 pts at the initial version. The sign of each conversation's gain, however, is the same in every repetition. The held-out gains of MemPro and GEPA are measured against the re-evaluated root at 78.50\%, with ties in development score going to the earlier version. Each selected version is evaluated once on the held-out questions, so the spread of these gains also contains evaluation noise.

\paragraph{LongMemEval repetitions.}
Table~\ref{tab:lmenoise} repeats the two measurements of Table~\ref{tab:noise} on LongMemEval\_S. GEPA runs once and selects its seed prompt, so it has no re-evaluation entries. EvolveMem and ARC are scored by the mixed F1 score, in which the abstention questions take the refusal score, and MemPro and GEPA by judge accuracy. The held-out gain is the gain of the selected version over the root on the 400 held-out questions, and SD is the standard deviation over the four runs. As on LoCoMo, re-evaluation flips 2.5--5.7\% of outcomes in the F1-selected loops and 8.7\% in MemPro, and 44--83\% of the accepted steps stay within $2\sigma$. Reruns reproduce the direction of the gain but not its size: every loop gains on the held-out questions on average, yet MemPro's held-out gain ranges from $+4.0$ to $+12.8$ pp across its runs, and ARC open/inc ends two of its four runs on its root.

\begin{table}[htbp]
\centering\small
\caption{Variation of the feedback under repetition on LongMemEval, with the columns of Table~\ref{tab:noise}. (a) The version each loop selected, evaluated three more times. (b) The whole loop, rerun four times from the same root. GEPA runs once, so it has no re-evaluation entries.}
\label{tab:lmenoise}
\setlength{\tabcolsep}{3pt}
\begin{tabular*}{\linewidth}{@{\extracolsep{\fill}}lcN{00.0}cN{00/00}cN{00}N{$+00.0$}cN{$+00.0$}c@{}}
\toprule
 & & \multicolumn{4}{c}{\makebox[0pt]{\textbf{(a) Same version, 3 more evaluations}}} & \multicolumn{5}{c}{\makebox[0pt]{\textbf{(b) Same root, loop run 4 times}}} \\
\cmidrule(lr){3-6}\cmidrule(l){7-11}
 & Selection & Flip & $2\sigma$ of & \multicolumn{2}{c}{\makebox[0pt]{Steps within $2\sigma$}} & & \multicolumn{2}{c}{\makebox[0pt]{Dev.\ gain}} & \multicolumn{2}{c}{\makebox[0pt]{Held-out gain}} \\
\cmidrule(lr){5-6}\cmidrule(lr){8-9}\cmidrule(l){10-11}
System & metric & rate (\%) & score & $k/n$ & (\%) & Rounds & Mean & SD & Mean & SD \\
\midrule
EvolveMem      & F1    & 3.7  & 2.9 & 6/10  & 60 & 7   & $+7.7$  & 3.5 & $+10.1$ & 3.8 \\
ARC closed/r0  & F1    & 2.5  & 2.2 & 15/21 & 71 & 9   & $+10.1$ & 1.0 & $+11.5$ & 4.8 \\
ARC closed/inc & F1    & 5.7  & 3.4 & 8/13  & 62 & 8   & $+10.4$ & 2.3 & $+12.6$ & 2.5 \\
ARC open/r0    & F1    & 3.4  & 2.6 & 4/9   & 44 & 7   & $+9.8$  & 4.7 & $+14.2$ & 3.9 \\
ARC open/inc   & F1    & 4.3  & 2.9 & 6/8   & 75 & 8   & $+7.2$  & 8.6 & $+7.9$  & 9.5 \\
MemPro         & Judge & 8.7  & 5.9 & 20/24 & 83 & 10  & $+18.3$ & 2.2 & $+7.8$  & 4.0 \\
GEPA           & Judge & ---  & --- & ---   & ---& 30  & $0.0$   & --- & $+0.3$  & --- \\
\bottomrule
\end{tabular*}
\end{table}

\paragraph{Mem-Gallery query images, measured.}
Table~\ref{tab:galleryvlm} reports the two arms of Appendix~\ref{app:galleryvlm}. Panel (a) compares them on the shared memory of round 0, where the only difference in input is the 487 questions with a query image. The image arm scores $+6.25$ points of F1 higher on these questions, with a 95\% bootstrap interval over the twenty scenarios of $[+2.82, +8.38]$, and differs by $-0.07$ points on the 1,224 questions that receive identical input. Panel (b) gives the gain of each arm's selected version. The two gains differ by $0.53$ points, well inside the $2\sigma$ of a re-evaluation in panel (c), and all twenty scenarios gain in both arms. Panel (c) repeats the measurements of Table~\ref{tab:noise}a and the retention counts of Table~\ref{tab:retention} on the selected versions. The flip rates, the share of accepted steps within $2\sigma$, and the early-to-late decline of repairs with a flat or rising regression volume are the same as in the caption-only runs, and late revisions break at least as many questions as they repair in both arms. The substitution of captions for query images thus changes the score on the questions that carry an image but none of the findings of Sections~\ref{sec:feedback}--\ref{sec:retention}.

\begin{table}[htbp]
\centering\small
\caption{EvolveMem on Mem-Gallery with the query image sent to the answer model (image) and with the dataset caption in its place (caption). (a) Difference between the arms at round 0, from the same memory, with 95\% scenario-cluster bootstrap intervals. (b) Round 0 and selected version, token F1 in points. (c) The selected version evaluated three more times, with the retention counts of Table~\ref{tab:retention} in pp of evaluated questions.}
\label{tab:galleryvlm}
\begin{tabular*}{\linewidth}{@{\extracolsep{\fill}}lccc@{}}
\toprule
\multicolumn{4}{@{}l}{\textbf{(a) Same memory, image minus caption}} \\
 & $n$ & F1 (pts) & Judge binary (pp) \\
\midrule
With a query image & 487 & $+6.25$ [$+2.82$, $+8.38$] & $+5.13$ [$+2.95$, $+7.09$] \\
Without a query image & 1,224 & $-0.07$ [$-0.28$, $+0.12$] & $-0.20$ [$-0.48$, $+0.07$] \\
\midrule
\multicolumn{4}{@{}l}{\textbf{(b) Evolution gain}} \\
 & Round 0 & Selected & Gain \\
\midrule
Image & 55.32 & 68.60 & $+13.28$ \\
Caption & 53.50 & 67.31 & $+13.81$ \\
\midrule
\multicolumn{4}{@{}l}{\textbf{(c) Selected version, three more evaluations}} \\
 & Flip (\%) & $2\sigma$ (pts) & Steps within $2\sigma$ \\
\midrule
Image & 1.68 & 2.28 & 30/61 (49\%) \\
Caption & 1.56 & 2.26 & 33/65 (51\%) \\
\midrule
\multicolumn{4}{@{}l}{Repair and regression volume, early $\to$ late (pp)} \\
 & Repair & Regression & \\
\midrule
Image & 4.5 $\to$ 1.9 & 2.0 $\to$ 2.9 & \\
Caption & 5.0 $\to$ 2.2 & 2.9 $\to$ 2.9 & \\
\bottomrule
\end{tabular*}
\end{table}

\paragraph{Decoding temperature.}
Re-evaluation flips outcomes even at temperature 0. MemPro's released evaluation decodes its task model at temperature 0.3 and its judge at temperature 0. Setting all six task-model calls to temperature 0 does not make a version reproducible on the model service we use. As shown in Table~\ref{tab:temperature}, two evaluations at temperature 0 flip as many outcomes as the most divergent pair at 0.3. The variation arises in the service itself. Five calls with an identical 9,000-character prompt at temperature 0 return five distinct outputs from Qwen3-30B-A3B, with or without a fixed seed, and the outputs diverge within the first 50 characters. Qwen3-max returned five distinct outputs on one day and two on another. We therefore keep the released temperature and use the re-evaluation of an unchanged version as the noise reference throughout.

\begin{table}[htbp]
\centering\small
\caption{Two evaluations of the same MemPro root at temperature 0 and at the released temperature 0.3. Questions are those of the 308-question split, and the last column gives the share of questions whose answer text is identical in both evaluations.}
\label{tab:temperature}
\begin{tabular}{@{}lN{000}cc@{}}
\toprule
Evaluation pair & Questions & Outcomes flipped & Identical answers \\
\midrule
Temperature 0, twice & 78 & 12 (15.4\%) & 42\% \\
Temperature 0.3, pairs & 78 & 7.7--15.4\% & 44\% \\
Temperature 0.3, pairs & 308 & 7.8--11.0\% & 48\% \\
\bottomrule
\end{tabular}
\end{table}

\section{Measurement Protocol}
\label{app:probes}
This section details the failure-stage probes of Section~\ref{sec:protocol} and reports how reliable their labels are. It then defines how claims, proposals, and actions are mapped to stages, and gives the labels and the acceptance rule of the intervention arms of Section~\ref{sec:intervention}. Appendix~\ref{app:prompts} prints the prompts that the probes and the arms use.

\subsection{Failure-stage probes}
\label{app:probedef}
The probes attribute a failure to a stage by intervening on its answer context. Each intervention changes one stage at a time and keeps the system's own answer model, answer prompt, and context format, in the order of Section~\ref{sec:protocol}. A failure that turns correct on replay is replay-correct and forms its own category, and a failure that stays wrong when answered from its gold evidence is gold-unresolved and receives no stage. The remaining failures are attributable. A failure belongs to memory extraction (S1) when the needed information did not reach memory, and to memory retrieval (S2) when memory holds it but the answer context does not. It belongs to answer generation (S3) when the context holds the evidence and the answer is still wrong. In terms of the memory modules of \citet{agentnative}, planning, integration, and context assembly fall in S2, and the answer prompt and answer parsing fall in S3. A probe answer counts as correct at a token F1 of at least 0.5.

Two added interventions keep provenance from deciding a stage on its own. First, a failure goes to S3 only when answering from the gold-linked memories already in its answer context turns it correct, and to the injection probe otherwise. Second, an extractor may cite a turn next to the annotated one. When the memories that cite a gold turn directly are missing or fail, the probe therefore uses those that cite a turn within two positions of a gold turn in the same session. The injection replaces the tail of the retrieved context with these memories, up to half of the retrieval depth, and a failure that injection does not resolve stays unresolved. The two added interventions change 4.9\% of the EvolveMem and ARC labels, and Table~\ref{tab:labelvariants} reports the stage readings without them.

EvolveMem and ARC use this protocol for the stage comparisons of Section~\ref{sec:local}. MemPro, in contrast, answers through a research loop of planning, retrieval, integration, and sufficiency checks before the final answer. A probe that reruns this loop can label an unsuccessful injection as S3 even when the evidence of the final answer has not been isolated. The probe for MemPro and GEPA therefore assigns planning, retrieval, integration, sufficiency checks, and continued retrieval to S2, and S3 starts at the final evidence-summary answer interface. It applies the checks in the same order, with the gold evidence first, then all persistent memories that share provenance with the gold turns, and then the original integrated memory. Its primary criterion is three consistent answers, and two of three serves as a sensitivity analysis. This probe has 2,809 completed question-level records.

\subsection{Label reliability}
\label{app:labels}
Table~\ref{tab:labelagree} reports three checks of the probe labels, namely repeated probing of the same failures, a second probe solver, and manual annotation. Table~\ref{tab:labelvariants} reports the stage readings of the main text under the judge verdict.

\begin{table}[htbp]
\centering\small
\caption{Agreement of the probe labels under repeated probing, a second probe solver, and manual annotation. Bottleneck-stage rows compare the stage that holds most attributable failures per round.}
\label{tab:labelagree}
\begin{tabularx}{\linewidth}{@{}Ycc@{}}
\toprule
Check & Records & Agreement \\
\midrule
Repeated probe labels, EvolveMem & 293 & 0.90 \\
Repeated probe labels, ARC & 175 & 0.90 \\
Repeated probe labels, MemPro under the looser boundary & 20 & 10/20 individual, 12/20 four-way \\
Three-run probe agreement, MemPro / GEPA & 1,234 / 1,575 & 0.86 / 0.91 \\
GPT-4o-mini probe solver against Qwen3-max, LoCoMo & 739 sampled & 0.704 \\
\quad records attributable under both solvers & 145 & 0.883 \\
\quad bottleneck stage per round & 16 rounds & 14/16 \\
GPT-4o-mini probe solver against Qwen3-max, Mem-Gallery & 319 & 0.871 \\
\quad records attributable under both solvers & 68 & 1.000 \\
\quad bottleneck stage per round & 5 rounds & 5/5 \\
Manual annotation against the probe, EvolveMem and ARC & 92 & 0.86 \\
\bottomrule
\end{tabularx}
\end{table}

\paragraph{Repeated labels and a second solver.}
Probing the same failures again reproduces 90\% of the labels on EvolveMem and ARC. On MemPro, a looser boundary that lets the probe rerun integration and the research loop reproduces only half of the individual labels, which is why the MemPro and GEPA probe starts S3 at the final answer interface. Under that boundary, all three probe runs agree on 86\% of MemPro and 91\% of GEPA records. With GPT-4o-mini as probe solver, the two solvers often attribute different records but name the same bottleneck stage in 14 of 16 LoCoMo rounds and in all 5 Mem-Gallery rounds. Record-level agreement is therefore lower than round-level agreement, and the bottleneck stage, which the main text uses, is the more stable of the two.

\paragraph{Manual annotation.}
Repeated probes show that the labels are stable, not that they name the stage a reader would name. We therefore annotate 92 EvolveMem and ARC failures by hand, drawn at random within each probe label. The annotator sees the question, the gold answer and its turns, the conversation, the stored memory without provenance, and the prompts and outputs of the answer model, but no probe result. For each failure, the annotator decides whether the memory holds the needed fact and whether the answer context holds it. The probe stage matches the manual stage on 86\% of the failures to which both assign a stage, with each probe label weighted by its share of all failures.

\paragraph{Verdict rule.}
The F1 threshold decides both which loop answers count as failures and which probe answers count as correct. We repeat the stage readings with the released LoCoMo judge~\citep{mempro} in both places, with and without the two added interventions (Table~\ref{tab:labelvariants}). Since the judge passes answers that match the gold answer in meaning, fewer failures stop at the gold-evidence probe and the S1 share falls. We find that every variant preserves the four readings of Section~\ref{sec:local}. Aligned edits repair more of their parents' failures than misaligned ones, nearly all gains beyond noise are aligned, S2 is the bottleneck of the root in all ten conversations, and most EvolveMem proposals miss the bottleneck stage. The main text uses the F1 verdict with both interventions, since F1 is the metric that the loops optimize.

\begin{table}[htbp]
\centering\small
\caption{Stage readings of EvolveMem, ARC closed/r0, and ARC open/inc under the F1 and the judge verdict, with and without the two added interventions. Shares give S1, S2, and S3 among attributable failures, pooled over distinct question--version instances, and root $\phi$ gives the S2 share at EvolveMem's round 0. Repair gives the mean repair rate of aligned and of misaligned edits as a range over the three systems, gains count the aligned gains beyond noise among all such gains, and missed counts the EvolveMem proposals, out of 67, that miss the bottleneck stage. The judge rows use the judge for failures, probes, repairs, and regressions.}
\label{tab:labelvariants}
\setlength{\tabcolsep}{3pt}
\begin{tabular*}{\linewidth}{@{\extracolsep{\fill}}llccccc@{}}
\toprule
Verdict & Interventions & Shares & Root $\phi$ & Repair (\%) & Gains & Missed \\
\midrule
F1 & original & .32/.55/.14 & 0.68 & 35--41 / 9--20 & 26/26 & 53 \\
F1 & added (main text) & .26/.62/.12 & 0.76 & 34--41 / 8--20 & 26/26 & 56 \\
Judge & original & .25/.60/.15 & 0.74 & 33--43 / 8--18 & 27/29 & 51 \\
Judge & added & .18/.68/.14 & 0.82 & 28--42 / 0--23 & 26/29 & 46 \\
\bottomrule
\end{tabular*}
\end{table}

\subsection{Stage maps for claims, proposals, and edits}
\label{app:stagemaps}
Section~\ref{sec:local} compares the claim, the proposal, and the executed action of each round with the bottleneck stage. This subsection defines how each of the three is read as a stage and checks these readings against an independent annotator.

\paragraph{Stage maps.}
ARC's diagnosis output names a stage directly. EvolveMem's claim is read from the free-text root causes of its diagnosis by an LLM annotator (qwen3.8-max, two samples), and a claim that names all three stages is not scored. Proposed and executed changes of EvolveMem and ARC pass through an operator table. Extraction templates and re-extraction map to S1, retrieval and context settings to S2, and answer prompts, answer-format flags, and verification to S3. A field that acts on several stages, such as a per-category override, maps to a set of stages, and a change to memory alone counts as S1. MemPro and GEPA diffs are annotated by the same model against the boundary that starts S3 at the final answer interface.

\paragraph{Map checks.}
We check the operator table against an independent reading. Given only the documentation line of each field, qwen3.8-max agrees with the table on every field it labels and abstains on four. Replacing the table with this reading leaves the stage of all 47 EvolveMem and 24 ARC executed actions unchanged, so the action rows of Tables~\ref{tab:alignment} and~\ref{tab:alignfull} do not depend on the table. The annotator's reading of EvolveMem proposals agrees with the table in 62 of 67 rounds, where the five differences read S2 as S2+S3, and the proposal agreement in Table~\ref{tab:stageaccuracy} does not change. Its reading of the ARC diagnosis text matches the stage that the loop itself states in all 57 rounds. For MemPro and GEPA, the annotator sees only the parent--child diff, without scores, gold evidence, acceptance, or the diagnosis text. Its two samples agree on 112 of 119 edits and match a rule baseline on the same 112, and the remaining seven edits stay unresolved rather than receive a stage.

\subsection{Verified feedback, repair rounds, and paired acceptance}
\label{app:acceptance}
Algorithm~\ref{alg:accept} shows how the harness labels the development questions of a version and how the repair arms of Section~\ref{sec:intervention} accept a revision. The harness evaluates every version twice and replays the failures of the first evaluation at temperature 0. A question is fragile when the two evaluations disagree, stable-correct when both succeed, replay-correct when both fail but the replay succeeds, and stable-wrong otherwise. Stable repairs and stable regressions count only transitions in which neither version has the question fragile. E1 and E2 show these labels to the agent in its ledger, whereas in E0 they are computed for measurement only. The threshold $z\geq1$ of the gate is a design choice rather than a significance test.

\begin{algorithm}[htbp]
\caption{Verified feedback and paired acceptance of a revision}
\label{alg:accept}
\begin{algorithmic}[1]
\Require incumbent $R$, revision or repair child $R'$, development questions $Q$
\ForAll{$V\in\{R,R'\}$}
  \State $A_V \gets$ evaluation of $V$ on $Q$; $B_V \gets$ a second, complete evaluation of $V$ on $Q$
  \State $T_V \gets$ temperature-0 replay of the questions wrong in $A_V$
  \ForAll{$q\in Q$}
    \State class$_V(q) \gets$ \textsc{fragile} if $A_V(q)\neq B_V(q)$; \textsc{stable-correct} if both correct; \textsc{replay-correct} if both wrong and $T_V(q)$ correct; \textsc{stable-wrong} otherwise
  \EndFor
\EndFor
\State $\begin{aligned}[t]
  f \gets \bigl|\{q\in Q:\;& \text{class}_R(q)\in\{\textsc{stable-wrong},\,\textsc{replay-correct}\}, \\
  & \text{class}_{R'}(q)=\textsc{stable-correct}\}\bigr|
\end{aligned}$
\State $\begin{aligned}[t]
  b \gets \bigl|\{q\in Q:\;& \text{class}_R(q)=\textsc{stable-correct}, \\
  & \text{class}_{R'}(q)\in\{\textsc{stable-wrong},\,\textsc{replay-correct}\}\}\bigr|
\end{aligned}$\Comment{excluding fragile $q$}
\State $z \gets (f-b)/\sqrt{f+b}$
\State $R'$ passes the gate \textbf{iff} $f-b\geq 1$ and $z\geq 1$; among versions that pass, keep the largest $f-b$
\end{algorithmic}
\end{algorithm}

\paragraph{Repair rounds.}
In E1 + repair and E2 + repair, a revision with at least one stable repair and one stable regression against the incumbent receives one repair session. The session sees the development questions that the revision repaired and regressed and the files it changed, and it is asked to keep the repairs while narrowing or reverting the change behind the regressions. Special cases keyed to a question or to its gold answer are prohibited, and the harness checks the diff of every repair child for them. The repair prompt adds no stage labels, and E2 + repair keeps the stage note of E2. The revision and its repair child are then compared with the incumbent by Algorithm~\ref{alg:accept}.

\section{Prompts}
\label{app:prompts}
This section prints, verbatim, the prompts that our measurements and the intervention arms add to the released systems. Placeholders in braces are filled at run time. The prompts of the released loops themselves are unchanged and are not repeated, and MemPro's answer judge is the released prompt of \citet{mempro}.

\subsection{Provenance contract}
\label{app:prompt_provenance}
The EvolveMem harness prefixes every dialogue turn of an extraction window with a turn identifier and appends the following contract to the extraction request. A reply without valid identifiers is rejected and regenerated. The ARC harness on the released Omni-SimpleMem appends a contract of the same form with its own wording.

\begin{promptlisting}{Provenance contract appended to every EvolveMem extraction request}
[PROVENANCE OUTPUT CONTRACT — REQUIRED]
 Every memory object MUST include "source_dgids": ["exact DGID", ...]. Select all dialogue turns that support that memory from the [DGID=...] labels in the current dialogue window. Return a nonempty list of unique exact IDs. Do not cite previous-window memory context, invent IDs, or copy DGID labels into lossless_restatement, keywords, persons, entities or topic. Session identity is assigned by code; do not generate a session ID. Example object shape: {"lossless_restatement":"...", "keywords":[], "timestamp":null, "location":null, "persons":[], "entities":[], "topic":"...", "source_dgids":["lc::TENANT::D1"]}. A response missing source_dgids is invalid and must be regenerated.
\end{promptlisting}

\begin{promptlisting}{Provenance contract of the ARC harness on Omni-SimpleMem}
[PROVENANCE OUTPUT CONTRACT - REQUIRED]
Every memory entry MUST include "source_dgids": ["exact DGID", ...]. Select all dialogue lines that support that entry from the [DGID=...] labels in the current window dialogues. Return a nonempty list of unique exact IDs. Do not cite the previous-window entries, invent IDs, or copy DGID labels into lossless_restatement, keywords, persons, entities or topic. Example entry shape: {"lossless_restatement":"...", "keywords":[], "timestamp":null, "location":null, "persons":[], "entities":[], "topic":"...", "source_dgids":["lc::conv-26::D1:3"]}. A response missing source_dgids is invalid and must be regenerated.
\end{promptlisting}

\subsection{Gold-evidence probe}
\label{app:prompt_gold}
The dataset-level gold-evidence probe of Appendix~\ref{app:gold} sends the following system and user messages once per question and answer model at temperature 0, with the annotated gold turns as the only evidence. The per-failure probe of Section~\ref{sec:protocol} instead answers from the gold turns with the system's own answer prompt.

\begin{promptlisting}{Gold-evidence probe, system message}
You answer questions about a person's past conversations. Use only the evidence provided. Answer as briefly as possible: a short phrase, a date, a name, a number, or yes/no. Only if the evidence contains nothing relevant to the question, reply exactly: Not mentioned.
\end{promptlisting}

\begin{promptlisting}{Gold-evidence probe, user message}
Evidence:
{evidence}

Question: {question}
Answer:
\end{promptlisting}

\subsection{Stage annotators}
\label{app:prompt_judge}
The stage annotator of Appendix~\ref{app:stagemaps} reads each EvolveMem diagnosis with the system message and the user message below, two samples at temperature 0, and a stage set is kept only when both samples parse and agree. The ARC diagnosis form and the configuration-field form replace the subject of the user message and follow the same stage definitions. MemPro and GEPA revisions are annotated from their parent-to-child diff alone with the edit prompt, and MemPro's diagnosis texts with the claim prompt.

\begin{promptlisting}{Stage annotator, system message}
You are a careful annotator. Answer with a single JSON object and nothing else.
\end{promptlisting}

\begin{promptlisting}{Stage annotator, user message for EvolveMem root causes}
The memory system answers questions about a long multi-session conversation (LoCoMo) in three stages. A failure is attributed to the stage where the answer was lost.

S1 memory_formation (extraction / writing): an LLM reads the conversation session by session and writes a bank of memories, each with content, keywords, persons, entities, timestamp and topic. This stage is fixed during the optimisation. A failure belongs to S1 when the fact needed for the answer never made it into the bank, or was written in a form that does not answer the question (paraphrased away, wrong date, merged with other facts).

S2 retrieval (selection / ranking): for each question the system searches the bank with a BM25 keyword view, a dense (BGE) semantic view and structured indices over persons / entities / topics, fuses the views (keyword_only, weighted_sum, rrf, with per-view top-k and weights), optionally decomposes the question into sub-queries or plans intents, optionally reranks for diversity (MMR), and keeps at most max_context memories. A failure belongs to S2 when the right memory exists in the bank but is not in the retrieved list, is ranked below the context cut-off, or is crowded out by irrelevant memories.

S3 answering (generation): the answer LLM receives the question and the retrieved memories formatted as context (up to 25 items), under an answer prompt / style (concise, explanatory, verifying), optionally followed by a verification pass; the output is scored with token F1 against the gold answer. A failure belongs to S3 when the needed memory is in the context but the model still answers wrongly, abstains ('Not mentioned'), gives a vague or incomplete answer, or answers in a form that does not match the gold answer.

Below is the diagnosis an optimisation LLM returned, as JSON, for one evaluation round of the memory system: three free-text cause descriptions (their bucket labels are withheld because the LLM was required to fill all three buckets every round) and its priority_actions lines. Its parameter suggestions and recipe names are withheld because they describe actions, not causes.

cause_1: {cause_1}
cause_2: {cause_2}
cause_3: {cause_3}
priority_actions: {priority_actions}

Task: record the stage or stages that the author explicitly blames for the failures. This is a transcription task, not a diagnosis: do not decide which stage is actually at fault, and do not complete the author's reasoning for them. Read all four lines; each line can name a cause.

What counts as naming a stage as the cause: a phrase that says the stage's own output was insufficient, irrelevant, shallow, misaligned, missing or wrong. Examples that count: 'insufficient or irrelevant retrieval', 'retrieval insufficiency', 'retrieval appears insufficient', 'relevant memories are present but not surfaced', 'the facts exist in the bank but are not retrieved', 'extraction dropped the date', 'memories are missing from the bank', 'the answer model ignored the evidence in the context'. Examples that do not count: a stage mentioned only as the setting of a symptom ('facts are missed despite 5 retrieved memories', 'low F1 after retrieval'), or a stage mentioned only in a list of options.

Rules, applied in order:
1. Symptom descriptions never count, whatever stage they hint at: zero-score counts, abstentions, 'not mentioned' answers, vague, verbose, incomplete or wrong answers, 'missing key facts'. They describe the outcome and are compatible with every stage.
2. A count or a category name is not evidence.
3. Statements about intended improvements, recipes, flags to enable, or what a change is meant to achieve are not evidence.
4. The evidence field must be a verbatim quote of the phrase that names the cause. If you cannot quote such a phrase, the answer is the empty list.
5. When rules 1-4 leave any doubt, the empty list is the correct answer. An empty list is never a mistake here; an inferred stage is.

Return JSON only: {"stages": [<zero or more of "S1", "S2", "S3">], "evidence": "<verbatim quote or empty>", "reason": "<one sentence naming the rule applied>"}
\end{promptlisting}

\begin{promptlisting}{Stage annotator, ARC diagnosis form}
Below are the first three lines of the '## Diagnosis' section an optimisation agent wrote before changing the configuration of the memory system (its 'change' line is withheld because it describes actions, not causes).

failure_pattern: {failure_pattern}
suspected_stage: {suspected_stage}
target: {target}

Task: record the stage or stages that the author explicitly blames for the failures. This is a transcription task, not a diagnosis: do not decide which stage is actually at fault, and do not complete the author's reasoning for them. Read all three lines; each line can name a cause.

What counts as naming a stage as the cause: a phrase that says the stage's own output was insufficient, irrelevant, shallow, misaligned, missing or wrong. Examples that count: 'insufficient or irrelevant retrieval', 'retrieval insufficiency', 'retrieval appears insufficient', 'relevant memories are present but not surfaced', 'the facts exist in the bank but are not retrieved', 'extraction dropped the date', 'memories are missing from the bank', 'the answer model ignored the evidence in the context'. Examples that do not count: a stage mentioned only as the setting of a symptom ('facts are missed despite 5 retrieved memories', 'low F1 after retrieval'), or a stage mentioned only in a list of options.

Rules, applied in order:
1. Symptom descriptions never count, whatever stage they hint at: low or zero F1, abstentions, 'Not mentioned' answers, vague, incomplete, wrong or overly long answers, 'facts were missed'. They describe the outcome and are compatible with every stage.
2. The suspected_stage line alone: if it repeats three or more of the options, that line carries no information (the other two lines are still read); if it names exactly one stage, that stage counts; if it names two stages joined by '|', '/', 'or' and no other line commits to one of them, the answer is the set of those two; if it says 'mixed, but primarily / dominant X', X alone counts.
3. Statements about intended improvements or what a change is meant to achieve are not evidence.
4. The evidence field must be a verbatim quote of the phrase that names the cause. If you cannot quote such a phrase, the answer is the empty list.
5. When rules 1-4 leave any doubt, the empty list is the correct answer. An empty list is never a mistake here; an inferred stage is.

Return JSON only: {"stages": [<zero or more of "S1", "S2", "S3">], "evidence": "<verbatim quote or empty>", "reason": "<one sentence naming the rule applied>"}
\end{promptlisting}

\begin{promptlisting}{Stage annotator, configuration-field form}
Below is one configuration field of the memory system that an optimisation agent changed, with the documentation line the agent was shown and the agent's own explanation of its changes.

field: {field}
documentation: {documentation}
proposed_change: {proposed_change}
agent_explanation: {agent_explanation}

Task: record the stage or stages this field acts on directly. This is a transcription of the documentation, not a judgement about whether the change helps; do not guess from the field name alone when the documentation says nothing.

What counts as acting on a stage: the documentation says the field changes what that stage produces. S1: how memories are written into the bank. S2: which memories are retrieved or how they are ranked (retrieval views, per-view top-k, fusion mode and weights, query decomposition or intent planning, coverage checks that trigger more retrieval, reranking, entity or graph expansion, time decay and its anchor date). S3: how the answer is generated from the retrieved context (answer prompt or style, verification passes, the answer model). A field that sets how many retrieved memories are handed to the answer prompt acts on both S2 and S3.

Rules, applied in order:
1. The agent's explanation may confirm the documentation but cannot add a stage the documentation does not support.
2. A field the protocol forces back to its default still maps to the stage it would act on; a field that only groups other fields (a container of per-category overrides, a recipe name) maps to no stage.
3. The evidence field must be a verbatim quote of the documentation phrase, or of the explanation only when the documentation line is missing and the explanation states the mechanism. If you cannot quote such a phrase, the answer is the empty list.
4. When rules 1-3 leave any doubt, the empty list is the correct answer. An empty list is never a mistake here; a guessed stage is.

Return JSON only: {"stages": [<zero or more of "S1", "S2", "S3">], "evidence": "<verbatim quote or empty>", "reason": "<one sentence naming the rule applied>"}
\end{promptlisting}

\begin{promptlisting}{MemPro and GEPA edit annotation}
S1 is persistent memory formation: creation or rewriting of stored
abstracts and raw memory pages. S2 is the complete research loop up to the
final integrated summary: planning, retrieval, integration, date/identity
processing, information checks, stopping, and follow-up search. S3 is the final
answer module after that summary: answer prompt, router, validator, and answer
postprocessing. Integration and InfoCheck are S2. Classify the operation that
the supplied edit actually changes, not the likely reason it was proposed.

Read the parent-to-child edit below. Determine which stage(s) the edit directly
changes. Do not use evaluation results, gold evidence, diagnosis notes, or
selected/rejected status; they are intentionally absent. A prompt or function
that spans more than one stage may be MIXED. If the diff does not establish a
stage, use UNKNOWN. Quote exact short lines from the supplied edit as evidence.
Return JSON: {"stages":["S1|S2|S3|MIXED|UNKNOWN"],"operation":"memory_formation|planning|retrieval|integration|info_check|answer_generation|router_validation|mixed|unknown","evidence":["exact quote"],"reason":"brief"}.


EDIT (the only evidence available):
{edit}
\end{promptlisting}

\begin{promptlisting}{MemPro diagnosis-claim annotation}
Read one historical diagnosis source. Return only explicitly asserted observed
faults that identify a stage operation. Do not turn a repair or intention into a
diagnosis, and do not infer a stage from an outcome, score, or file name. Keep
tentative causes out. For every claim quote an exact short substring from the
source. Return JSON with claims, each having stage (S1/S2/S3), assertion,
statement_type (asserted_fault or other), quote, and reason.
Schema: {"claims":[{"stage":"S1|S2|S3","assertion":"...","statement_type":"asserted_fault|other","quote":"...","reason":"..."}]}.


SOURCE:
{source}
\end{promptlisting}

\subsection{Feedback of the intervention arms}
\label{app:prompt_arms}
Every MemPro iteration prompt consists of the released task prompt followed by the harness notes. E1 appends the evidence-ledger notes, and E2 appends one further note. Between iterations the harness rewrites the ledger of the current version in the agent's workspace, whose E1 view and E2 additions are printed after the notes.

\begin{promptlisting}{Task prompt of the released loop, all arms}
Read AGENTS.md, registry/versions.json, and the local scripts first. Continue MemPro framework evolution for locomo: select a base version, create one coherent candidate direction, use targeted debug before any full benchmark, and avoid sample-specific shortcuts.
\end{promptlisting}

\begin{promptlisting}{Harness notes appended in every arm}
REPRODUCTION NOTES (outer evolution iteration {iteration} of {total}; these are harness facts, not instructions about how to analyse):
- The evolution split is data/locomo/locomo10_train.json (154 questions over 10 conversations). Always pass
  `--data data/locomo/locomo10_train.json` to scripts/run_eval.py (after the `--` separator). No other dataset
  file exists here; the held-out split is not available.
- This session must produce exactly one new child version (scripts/register_version.py, edit only files under
  versions/<new_id>/), benchmark it on the full evolution split, and record its metrics in registry/versions.json
  yourself from runs/<new_id>/train/batch_statistics_0_9.json: metrics.train_score = overall_llm_as_judge_avg * 100
  (2 decimals), metrics.train_by_type = {category_type: llm_as_judge_avg * 100}, metrics.train_num_questions = 154,
  status = "evaluated". Record only full 154-question runs, never a partial or debug run. Do not write any
  metrics before runs/<new_id>/train/batch_statistics_0_9.json exists; the driver verifies the recorded train_score
  against the per-question results in runs/<new_id>/train/conv-*/qa_results.json (all 10 conversations, 154
  questions) and will send you back until they match. Writing batch_statistics or registry metrics by hand is
  detected and only wastes the session.
- A full train-split run takes 30-60 minutes. Launch it detached so it survives tool timeouts, e.g.
  `setsid nohup python scripts/run_eval.py --version <id> --split train -- --data data/locomo/locomo10_train.json
  > runs/<id>/train_eval.log 2>&1 < /dev/null &`, then poll with `sleep 120; tail -3 runs/<id>/train_eval.log;
  ls runs/<id>/train/` until batch_statistics_0_9.json appears. Never kill an evaluation process and never start a
  second evaluation into the same runs/<id>/train directory. If you edit the version after a debug run, pass
  --force-runtime to run_eval.py so the runtime is rebuilt from your files.
- Paths are relative to the workspace MemPro/locomo (your cwd). The evaluation output is
  runs/<id>/train/batch_statistics_0_9.json; check it with `ls runs/<id>/train/` from the workspace root, not from
  scripts/. Before launching any evaluation run `pgrep -af eval/locomo_test.py`; if one is running for <id>, wait.
  A second launch into the same runs/<id>/train is refused by run_eval.py (exit code 75).
- Do not create, copy or edit anything under versions/v0000/, best_versions/, eval/, data/, scripts/, or
  MemPro/_shared/; do not copy files from best_versions/ into a version directory. Do not change model names,
  API settings, temperature or worker counts.
- File editing: `apply_patch` needs the exact patch format (`*** Begin Patch`, `*** Update File: <path>`, `@@`
  hunks with a context line, lines prefixed by space/+/-; no line numbers, no JSON wrapper). If a patch is
  rejected, rewrite the whole file with `cat > <path> <<'EOF' ... EOF` and re-read it to verify.
- End the session with one JSON line:
  {"new_version": "<id>", "parent": "<id>", "train_score": <float or null>, "summary": "<one line>"}
\end{promptlisting}

\begin{promptlisting}{Evidence-ledger notes appended in E1 and E2}
EVIDENCE LEDGER (harness-measured facts about the registered versions; read evidence/README.md and evidence/INDEX.md first,
then evidence/<version>_ledger.md of the version you select as base):
- Every evaluated version was re-evaluated once more under the same configuration (same memory), and every judge-WRONG
  question of it was replayed at temperature 0. From these, each of the 154 questions is classed for that version as
  stable-wrong / ghost / fragile / stable-right (definitions in evidence/README.md). The ledger also lists which questions a
  child fixed and which it broke relative to its parent counting only stable transitions, and a table of everything tried
  so far with both evaluation scores.
- Only stable-wrong questions are reproducible failures. Ghost and fragile questions flip without any change to the
  framework; fixing one of them proves nothing and it flips back on the next evaluation. Choose the failure pattern to
  work on from the stable-wrong set, and read the debug you run against it.
- A single train score moves by several points between two evaluations of the same version; compare a child with its
  parent by both evaluations, not by one number.
\end{promptlisting}

\begin{promptlisting}{Additional evidence-ledger note in E2}
- For each stable-wrong question the ledger also gives the failure stage found by evidence probes (ineligible / S1 memory
  / S2 retrieval / S3 integration-or-answer), the gold evidence page ids, whether they are in the memory bank, and which
  pages were retrieved; definitions in evidence/README.md. Use the stage counts to decide which part of the pipeline to
  change.
\end{promptlisting}

\begin{promptlisting}{Ledger of one version, E1 view}
# Evidence ledger for {version} (harness-measured, generated {timestamp})

Two independent full evaluations of {version} on the 154-question evolution split, same configuration, same memory: A (the recorded run) = {score_A}, B (re-run) = {score_B}. Mean = {mean}. Every judge-WRONG question of run A was replayed once more at temperature 0 with the same version.

## Question classes

| class | n | meaning |
| --- | ---: | --- |
| stable-wrong | {n} | wrong in A, wrong in B, wrong on the temperature-0 replay. The only reproducible failures. |
| ghost | {n} | wrong in A and B but CORRECT on the temperature-0 replay of this same version: the failure is sampling noise, not a defect. |
| fragile | {n} | A and B disagree. Boundary questions; they flip under any change and flip back. A fix of one of these is not evidence. |
| stable-right | {n} | correct in A and B. |

### stable-wrong questions (work on these)

| question | type | recorded answer | gold |
| --- | --- | --- | --- |
| {conv} q{k}: {question} | {type} | {recorded_answer} | {gold} |

### ghost and fragile questions (do NOT treat as failures; they flip without any change)

| question | type | class | A | B |
| --- | --- | --- | --- | --- |
| {conv} q{k}: {question} | {type} | {ghost_or_fragile} | {C_or_W} | {C_or_W} |

## Change from parent {parent} to {version} (stable transitions only)

- fixed (stable-wrong/ghost in {parent} -> stable-right in {version}): {n} {question_ids}
- broken (stable-right in {parent} -> stable-wrong/ghost in {version}): {n} {question_ids}
- flips that involve a fragile question on either side (not counted): {n}

## Everything tried so far in this workspace

| version | parent | train A | train B | mean | note |
| --- | --- | ---: | ---: | ---: | --- |
| {version_id} | {parent_id} | {train_A} | {train_B} | {mean} | {registry_note} |

A single train score moves by up to ~6 points between two evaluations of the same version (2σ of the recorded re-evaluations); a child is only better than its parent if both of its evaluations are above both of the parent's.
\end{promptlisting}

\begin{promptlisting}{Ledger of one version, stable-wrong section of the E2 view}
### stable-wrong questions (work on these)

| question | type | stage (probe result) | gold pages | in bank | retrieved | recorded answer | gold |
| --- | --- | --- | --- | --- | --- | --- | --- |
| {conv} q{k}: {question} | {type} | {stage_text} ({sublabel}) | {gold_pages} | {gold_pages_in_bank} | {retrieved_pages} | {recorded_answer} | {gold} |

Stage counts over stable-wrong: {label} {count}, {label} {count}.
\end{promptlisting}

\begin{promptlisting}{Ledger of one version, texts of the stage column in the E2 view}
ineligible: the answer model + judge fail even when the gold evidence turns are pasted in directly; not fixable through memory/retrieval/integration
S1 memory: the gold pages are in the bank but their abstracts do not carry the needed fact; injecting the raw gold pages fixes it
S2 retrieval: the gold pages are in the bank and their abstracts suffice, but retrieval did not return them; injecting them fixes it
S3 integration/answer: the gold pages were retrieved, the answer is still wrong
S3 integration/answer: even with the gold pages injected into the retrieved set the answer stays wrong
probe failed (no label)
\end{promptlisting}

\subsection{Repair session}
\label{app:prompt_repair}
In the repair arms, a revision with at least one stable repair and one stable regression against the incumbent receives one repair session. Its prompt is the head below, followed by the harness and evidence notes of the arm and the closing line. Regular iterations of the repair arms end with the acceptance note.

\begin{promptlisting}{Repair session prompt, head}
REPAIR TURN (harness-initiated; outer evolution iteration {iteration} of {total}, repair session {k}).

Your candidate {candidate} (built on {base}; note: "{note}") was evaluated twice by the harness under the same configuration and its
wrong answers were replayed at temperature 0. Counting only STABLE transitions relative to {base} (definitions in evidence/README.md):
it FIXED {fixed_n} question(s) that {base} got reproducibly wrong, and it BROKE {broken_n} question(s) that {base} answered
correctly in both of its evaluations. Two scores of {candidate}: {cand_A} / {cand_B}; of {base}: {base_A} / {base_B}.

BROKEN by {candidate} (correct in {base} in both evaluations, wrong in {candidate} in both evaluations and on the replay):
| question | type | gold | {base} answer (judged CORRECT) | {candidate} answer (judged WRONG) |
| --- | --- | --- | --- | --- |
{broken_rows}

FIXED by {candidate} (keep these):
| question | type | gold | {candidate} answer (judged CORRECT) |
| --- | --- | --- | --- |
{fixed_rows}

Files that differ between versions/{base}/ and versions/{candidate}/: {changed_files}.

YOUR TASK IN THIS SESSION: register exactly one child of {candidate} (`python scripts/register_version.py --base-version {candidate} --new-version <next id>
--note "repair of {candidate}: ..."`) that KEEPS the fixes and REMOVES the regressions. Do not start a new direction. Procedure:
1. Read the diff (`diff -u versions/{base}/<file> versions/{candidate}/<file>` for each changed file) and the traces of the broken questions
   under runs/{candidate}/train/<conv>/ (research_trace_q<k>.json, qa_results.json) next to the same questions under runs/{base}/train/<conv>/,
   to find which part of the change turned each broken question wrong (typically: a new instruction that over-applies, a changed
   retrieval plan that drops the evidence, a reformatted answer the judge no longer accepts).
2. Narrow that part: make it conditional on the situation it was meant for, restrict it to the question types it helped, or revert
   it and keep the rest. Keep everything else byte-identical to {candidate}.
3. Debug on the broken and the fixed questions before the full run: `python scripts/run_eval.py --version <id> --split train -- --data
   data/locomo/locomo10_train.json --conv-id <conv> --question-index <k>` runs one question (indices as in the tables above; memory for
   that conversation is built on first use). Check that the broken ones are correct again and the fixed ones stay correct.
4. Then run the FULL train evaluation, record the metrics and finish with the JSON line, exactly as in a normal iteration.

HARD RULES: no special-casing by sample id, question index, question wording, gold answer or answer lexicon (the harness scans the
diff of the new version and rejects such versions); no edits outside versions/<new id>/; do not touch the memory construction unless
the regressions come from it (then memory must be rebuilt: do not reuse {candidate}'s memory).
\end{promptlisting}

\begin{promptlisting}{Repair session prompt, closing line}
HARNESS ACCEPTANCE (repair turn): the base version of the child you register MUST be {candidate} (`--base-version {candidate}`); the driver sends you back otherwise.
\end{promptlisting}

\begin{promptlisting}{Acceptance note appended to regular iterations of the repair arms}
HARNESS ACCEPTANCE: the harness keeps one incumbent version and decides acceptance from paired stable transitions (evidence/INCUMBENT.md). The current incumbent is {incumbent}. The base version of the child you register in this session MUST be {incumbent} (`--base-version {incumbent}`); versions not on the incumbent chain were rejected and must not be used as base (rejected so far: {rejected_list}). The driver sends you back if the registered child has another parent.
\end{promptlisting}

\section{Additional Experiments}
\label{app:experiments}
This section reports seven further experiments. The first checks that recording provenance does not change the outcomes it helps to measure. The next three repeat the measurements of Sections~\ref{sec:feedback}--\ref{sec:retention} on the released Omni-SimpleMem, with other model families, and on a third benchmark. The last three follow where diagnoses fail, which failures MemPro's loop actually receives, and how large a step the development set can resolve.

\subsection{Effect of provenance recording}
\label{app:provenance}
Separating S1 from S2 requires provenance, the dialogue turns from which each memory was extracted. The EvolveMem harness, on which ARC also runs, records it during extraction. Every turn in the extraction window carries an identifier, and a 726-character contract appended to the extraction prompt asks the extractor to cite the identifiers that support each memory (Appendix~\ref{app:prompt_provenance}). The harness checks the cited identifiers against the window and keeps them as metadata, outside the indexed text and the answer prompt. Only 2 of EvolveMem's 5,240 round-0 memories copy an identifier into their text. The longer prompt could still change what is extracted. MemPro's loop, in contrast, runs without provenance, which we add only when replaying frozen versions, so recording cannot affect that loop.

To check that the measurement does not cause the changes it measures, we apply the test of Section~\ref{sec:feedback} to provenance recording and ask whether it changes outcomes more than an unchanged system does. With Qwen3-max for extraction and answering and EvolveMem's weak initial configuration, we answer the first 20 LoCoMo questions of nine conversations from memory extracted without and with provenance, under an identical QA configuration. Since the two memories come from separate extractions, Table~\ref{tab:provenance} sets their difference against two references on the same questions and configuration, namely the same memory answered twice and memory extracted twice by the same provenance-recording code.

\begin{table}[htbp]
\centering\small
\caption{Effect of provenance recording on answers. EvolveMem static QA on the same 180 LoCoMo questions, scored by token F1$\times$100. Each row pairs two evaluations.}
\label{tab:provenance}
\setlength{\tabcolsep}{4pt}
\begin{tabular*}{\linewidth}{@{\extracolsep{\fill}}lcccN{000}@{}}
\toprule
 & $\Delta$F1 & 95\% interval & F1 changed & Answer text \\
Pair & (pp) & (pp) & (up/down) & changed \\
\midrule
Same memory, answered twice & $+0.43$ & $[-1.5, +2.0]$ & 12 (7/5) & 30 \\
Memory extracted twice, same code & $-0.94$ & $[-5.5, +2.9]$ & 51 (30/21) & 88 \\
Without $\to$ with provenance & $+0.48$ & $[-3.9, +4.4]$ & 69 (30/39) & 103 \\
\bottomrule
\end{tabular*}
\tabnote{Both evaluations of the first two rows record provenance. The first row evaluates EvolveMem's round 0 again, and the second compares two builds by the same code. Intervals come from a paired bootstrap over the nine conversations with 20,000 draws. The questions comprise 71 multi-hop, 76 temporal, 29 open-domain, and 4 single-hop questions. One conversation is excluded because its provenance build is incomplete.}
\end{table}

As Table~\ref{tab:provenance} shows, recording provenance moves mean F1 by $+0.48$ pp, less than the $-0.94$ pp between two extractions with unchanged code, and all three intervals include zero. We do not read the $+0.48$ pp as a gain, since excluding one baseline answer that was truncated at the output limit gives $-0.07$ pp. Recording changes the F1 of 69 questions, against 51 for re-extraction and 12 for re-answering. The turnover thus comes mostly from rebuilding memory, which every extraction run incurs. Across all 1,388 category 1--4 questions of the nine conversations, re-extraction alone changes 35\% of question scores and flips 17\% of zero/nonzero outcomes, against 3.5\% when the same memory is answered twice. Score levels agree as well, with four provenance-recording builds at 35.5--37.7 on these questions and the build without provenance at 37.2. The provenance build holds 4,866 memories against 4,850, each with non-empty and unique identifiers from its own session. No identifier appears in a searchable memory field or in any of the 180 answer prompts.

The cost falls on extraction. On the nine conversations, extraction uses 34.5\% more input and 20.0\% more output tokens, mainly for the per-turn identifiers and the repeated contract, which raises the estimated cost by 22.4\%. Answer-generation input tokens fall by 3.7\%.

Each arm is one run over nine conversations, so the intervals rule out shifts larger than about 4 pp but do not establish equivalence at the scale of a single evolution step (Table~\ref{tab:noise}). The check also concerns answers rather than which facts the contract leads the extractor to keep. Every EvolveMem and ARC version, rebuilt ones included, is extracted with the same recording, so the parent--child comparisons of the main text compare versions recorded the same way.

\subsection{The ARC loop on the released Omni-SimpleMem system}
\label{app:omni}
ARC reports Omni-SimpleMem~\citep{omnimem} as the memory system that its loop discovered, but releases only its final form. Our ARC configurations therefore evolve EvolveMem's initial pipeline, which leaves open whether the loop behaves differently on the system it produced. To test this, we run the same ARC harness on our reproduction of the released Omni-SimpleMem.

\paragraph{Setup.}
On LoCoMo we run five configurations, namely closed/r0, closed/inc, open/r0, open/inc, and open/inc with the GPT-4o agent of Appendix~\ref{app:substitutions}. On Mem-Gallery, Omni-SimpleMem stores each dialogue round verbatim without an extraction template, so only closed/r0, closed/inc, and closed/inc with the GPT-4o agent apply. Extraction carries a provenance contract of the same form as EvolveMem's (Appendix~\ref{app:prompt_provenance}). Since the released runner overrides the retrieval depth internally, we reimplement the retrieval and answer path step by step from the released code, so that its configuration fields take effect. Each configuration runs from two starting versions. The \emph{published} start is the released configuration. The \emph{weak} start switches off the discoveries that the system itself reports, whose intermediate versions are not released. On LoCoMo these are the rule-based answer prompt with structured output, hybrid retrieval, deeper retrieval, and context expansion, and on Mem-Gallery the dialogue text in context, hybrid retrieval, and the image catalogue. The weak start is our own construction, comparable to EvolveMem's weak initial configuration, rather than the original baseline of the paper. Everything else follows the ARC runs of the main text, including the ten conversations, the twenty scenarios, the stopping rule after two rounds without improvement, and every measurement. Table~\ref{tab:omni} reports the in-sample gain of each configuration's best version over round 0, averaged over conversations or scenarios.

\begin{table}[htbp]
\centering\small
\caption{The ARC loop on the released Omni-SimpleMem from the published and from the weak start. On LoCoMo, scores are the mixed F1 score$\times$100 averaged over ten conversations, with the gain in category 1--4 F1 in parentheses, and Round 2 gives the gain after round 2. On Mem-Gallery, scores are the official F1$\times$100 over all 1,711 questions, with the scenario-mean gain and its split-half replay. Accepted counts the accepted steps, with those inside the $2\sigma$ band of the loop's recorded same-configuration re-evaluations in parentheses.}
\label{tab:omni}
\setlength{\tabcolsep}{2pt}
\begin{tabular*}{\linewidth}{@{\extracolsep{\fill}}llccN{$+00.0$}cc@{}}
\toprule
Start & Configuration & Round 0 & Gain & Round 2 & Rounds & Accepted (in $2\sigma$) \\
\midrule
\multicolumn{7}{@{}l}{\emph{LoCoMo}} \\
Published & closed/r0 & 59.1 & $+1.8$ ($+1.2$) & $+1.1$ & 2--9 & 22 (19) \\
 & closed/inc & 59.4 & $+1.4$ ($+0.9$) & $+1.1$ & 2--7 & 11 (7) \\
 & open/r0 & 59.2 & $+1.7$ ($+0.4$) & $+1.0$ & 2--10 & 15 (10) \\
 & open/inc & 59.3 & $+1.7$ ($+0.6$) & $+1.6$ & 2--6 & 11 (7) \\
 & open/inc, GPT-4o & 59.0 & $+2.0$ ($+2.3$) & $+1.2$ & 2--8 & 16 (---) \\
Weak & closed/r0 & 23.8 & $+36.9$ ($+21.9$) & $+35.1$ & 3--10 & 31 (16) \\
 & closed/inc & 23.7 & $+36.6$ ($+21.2$) & $+35.6$ & 3--8 & 21 (4) \\
 & open/r0 & 23.9 & $+33.3$ ($+16.3$) & $+32.6$ & 3--6 & 21 (6) \\
 & open/inc & 23.9 & $+32.7$ ($+16.4$) & $+29.5$ & 3--10 & 32 (14) \\
 & open/inc, GPT-4o & 23.7 & $+36.4$ ($+21.9$) & $+30.3$ & 3--10 & 29 (11) \\
\midrule
\multicolumn{2}{@{}l}{\emph{Mem-Gallery}} & Round 0 $\to$ best & Scen.\ mean & Split-half & Rounds & Accepted \\
\cmidrule(l){3-7}
Published & closed/r0 & 69.0 $\to$ 70.4 & $+1.4$ & $+0.7$ & 67 & 26 \\
 & closed/inc & 69.0 $\to$ 70.5 & $+1.7$ & $+1.0$ & 74 & 29 \\
 & closed/inc, GPT-4o & 68.8 $\to$ 69.8 & $+1.0$ & $+0.2$ & 76 & 31 \\
Weak & closed/r0 & 41.7 $\to$ 68.1 & $+24.6$ & $+23.9$ & 96 & 47 \\
 & closed/inc & 41.8 $\to$ 68.1 & $+24.5$ & $+23.6$ & 93 & 46 \\
 & closed/inc, GPT-4o & 41.7 $\to$ 66.9 & $+23.2$ & $+22.2$ & 95 & 48 \\
\bottomrule
\end{tabular*}
\tabnote{On LoCoMo, Rounds gives the range of rounds run per conversation before the stopping rule. On Mem-Gallery, rounds and accepted steps are summed over the twenty scenarios, and every run stops after 2--6 rounds. Split-half replays the accepted steps on the other half of each scenario's questions. The GPT-4o agent run on LoCoMo records no same-configuration re-evaluation, so its steps are not classified.}
\end{table}

\paragraph{From the published start.}
We find that the loop, run again on the system it produced, mostly accepts steps that a re-evaluation of the parent could also produce. Its best versions gain $+1.4$ to $+2.0$ pts on LoCoMo and $+1.0$ to $+1.7$ F1 points on Mem-Gallery, and the runs stop after two to ten rounds. These gains are of the size of the measurement. Re-evaluating the published round 0 on the same memory flips 3.6\% of its 1,540 LoCoMo outcomes, 1.9\% up and 1.7\% down. In the four LoCoMo configurations with the Qwen agent, 43 of the 59 accepted steps lie within the $2\sigma$ band of the loop's recorded re-evaluations, among them 19 of the 22 in closed/r0. On Mem-Gallery, split-half replay keeps only $+0.2$ to $+1.0$ of these gains. On Omni-SimpleMem, too, the accepted steps show the pattern of Table~\ref{tab:noise}.

\paragraph{From the weak start.}
Because the published system may already sit near its ceiling, the weak start switches off its reported discoveries to test how the same loop behaves when substantial headroom remains. Under this setting, we find that the loop concentrates nearly all of its gain in the first two rounds, adding $+32.7$ to $+36.9$ pts on LoCoMo with $+29.5$ to $+35.6$ pts already in place after round 2. In nine of the ten closed/r0 conversations, the first accepted edit replaces the free-form answer call with the released rule-based prompt, and in six of them it also enables hybrid retrieval. The agent thus regains the first discoveries of the released system in one step. Later rounds accept adjustments of retrieval and context settings, and 51 of the 134 accepted steps of the five weak-start configurations lie within $2\sigma$. On Mem-Gallery, the weak start gains $+23.2$ to $+24.6$ scenario-mean F1 points, of which split-half replay keeps $+22.2$ to $+23.9$. This first step is a general repair rather than a fit to the development conversation. Applied to the banks of the other nine conversations, the configuration that closed/r0 selects on one conversation after its first round raises their mixed score from 24.1 to 58.5. The gain is positive in all nine conversations, from $+28.6$ to $+36.8$ pts. Likewise, the configuration selected on one Mem-Gallery scenario raises the F1 of the other nineteen from 42.5 to 64.8, with a positive gain in each. This one large step mirrors the retrieval upgrade that carries the EvolveMem and ARC gains in Section~\ref{sec:local}.

\paragraph{Diagnoses.}
The claims follow the pattern of Table~\ref{tab:stageaccuracy}. In closed/r0, the claim matches the bottleneck stage of the version read in 0.57 of the scorable rounds from the published start and in 0.82 from the weak start. Neither exceeds a constant claim of retrieval, which would score 0.63 and 1.00. Agreement falls to 0.00--0.34 once the diagnoser reads the incumbent or the action space is open. In the open configurations from the weak start, the agent names the answer stage in 39--48 rounds and rewrites the answer prompt. Repeated probe labels agree at 0.90--0.94 on the weak-start LoCoMo runs and at 0.96--0.98 on Mem-Gallery.

\subsection{Model substitutions}
\label{app:substitutions}
In the main runs, Qwen models fill every role of the loop. The low stage agreement of Table~\ref{tab:stageaccuracy} could therefore reflect a weak diagnoser rather than a property of the loop, and a stronger diagnoser could raise agreement and, with it, the gain. To test whether the readings of Sections~\ref{sec:feedback}--\ref{sec:retention} depend on the model family, we substitute GPT models on both datasets. The EvolveMem diagnoser becomes GPT-4o or GPT-4o-mini, the EvolveMem answer model becomes GPT-4o-mini under a GPT-4o diagnoser, and the ARC agent becomes GPT-4o. Each run keeps the root memory, the loop, the acceptance rule, and the Qwen3-max probe labels of its Qwen counterpart, and the answer-model runs re-answer round 0 on the same memory. Table~\ref{tab:substitutions} first reports, for each run, how often the stage of a proposal (EvolveMem) or of a claim (ARC) matches the bottleneck stage of the version read. It also gives the in-sample gain of the best version over round 0. We then repeat the three measurements of the main text, namely the same-input flips of Section~\ref{sec:feedback}, the alignment ledger of Table~\ref{tab:alignment}, and the early and late repair and regression volumes of Table~\ref{tab:retention}.

\begin{table}[htbp]
\centering\small
\caption{Stage agreement and in-sample gain under model substitution. Agreement compares the stage of the proposal (EvolveMem) or of the claim (ARC) with the bottleneck stage of the version read, on rounds with at least five attributable failures. Gain is the in-sample gain of the best version over round 0 in pts of the loop's score, the mixed F1 score on LoCoMo and F1 on Mem-Gallery, averaged over conversations or scenarios. The LoCoMo ARC runs rebuild memory when the extraction template changes, unlike the open/inc run of Table~\ref{tab:stageaccuracy}. The Mem-Gallery GPT-4o ARC row covers the 17 of 20 scenarios whose runs pass verification, and the Mem-Gallery GPT-4o-mini answer-model run has no probe labels.}
\label{tab:substitutions}
\begin{tabular}{@{}lllcN{$+00.0$}@{}}
\toprule
System & Diagnoser or agent & Answer model & Agreement & Gain (pts) \\
\midrule
\multicolumn{5}{@{}l}{\emph{LoCoMo}} \\
EvolveMem & Qwen3-max & Qwen3-max & 0.16 (11/67) & $+7.9$ \\
 & GPT-4o & Qwen3-max & 0.30 (19/63) & $+11.9$ \\
 & GPT-4o-mini & Qwen3-max & 0.69 (44/64) & $+2.9$ \\
 & GPT-4o & GPT-4o-mini & 0.58 (40/69) & $+7.9$ \\
ARC open/inc & Qwen3-max & Qwen3-max & 0.50 (20/40) & $+5.5$ \\
 & GPT-4o & Qwen3-max & 0.68 (41/60) & $+11.0$ \\
\midrule
\multicolumn{5}{@{}l}{\emph{Mem-Gallery}} \\
EvolveMem & Qwen3-max & Qwen3-max & 0.82 (59/72) & $+11.0$ \\
 & GPT-4o & Qwen3-max & 0.86 (69/80) & $+8.0$ \\
 & GPT-4o & GPT-4o-mini & -- & $+5.5$ \\
ARC open/inc & Qwen3-max & Qwen3-max & 0.22 (8/37) & $+5.7$ \\
 & GPT-4o & Qwen3-max & 1.00 (23/23) & $+13.8$ \\
\bottomrule
\end{tabular}
\end{table}

\paragraph{Agreement and gain.}
We find that agreement with the stage reference does not order the gains, whereas the round in which a loop reaches the retrieval upgrade of Section~\ref{sec:local} does. On LoCoMo, the GPT-4o diagnoser agrees more often than Qwen3-max (0.30 against 0.16) and gains $+11.9$ rather than $+7.9$ pts. The GPT-4o-mini diagnoser, however, agrees most often (0.69) and gains only $+2.9$ pts. GPT-4o proposes semantic retrieval in round 0 of all ten conversations, as the stage-guided proposer of Section~\ref{sec:local} does, and ends close to that proposer's $+12.3$ pts (Table~\ref{tab:proposer}). GPT-4o-mini instead changes a retrieval fusion setting without enabling semantic retrieval, so only three of its conversations reach the upgrade, in rounds 4 to 7, compared with seven for Qwen3-max. With GPT-4o-mini as the answer model, the answer stage becomes the bottleneck in 32 of 69 rounds, against none of 67 with Qwen3-max. The proposals of the GPT-4o diagnoser match the bottleneck stage in 40 of the 69 rounds, yet the loop gains the same $+7.9$ pts as the Qwen loop.

The GPT-4o ARC agent claims retrieval in 62 of its 66 diagnoses and agrees with the bottleneck more often than the Qwen agent (0.68 against 0.50). Since its edits stay within the retrieval settings, all 67 claims in these diagnoses are executed, whereas the guard stops the edits behind 26 of the Qwen agent's 43 diagnoses. The GPT-4o agent thus doubles the in-sample gain ($+11.0$ against $+5.5$ pts). On Mem-Gallery, nearly every attributable round has its bottleneck at retrieval. As the EvolveMem diagnoser, GPT-4o agrees with it slightly more often than Qwen3-max but gains less ($+8.0$ against $+11.0$ pts). As the ARC agent, it claims retrieval in all 23 attributable rounds and gains $+13.8$ pts, against $+6.6$ for the Qwen agent on the same 17 scenarios. Across these runs, the substituted model changes which stage the loop keeps editing and how early it finds the one upgrade that pays, which is the result of Table~\ref{tab:proposer} seen from another side.

\begin{table}[htbp]
\centering\small
\caption{Alignment ledger under model substitution. The ledger of Table~\ref{tab:alignment} on the LoCoMo runs of Table~\ref{tab:substitutions}, with the Qwen runs repeated for reference. The noise band of the GPT-4o ARC agent, whose run records a single same-configuration re-evaluation, is that of the Qwen run with the same harness and evaluation models.}
\label{tab:substalign}
\setlength{\tabcolsep}{3pt}
\begin{tabular*}{\linewidth}{@{\extracolsep{\fill}}lcN{00}N{00.0}N{00.0}N{00}cN{00}cN{00}N{00}N{00}@{}}
\toprule
 & & & \multicolumn{2}{c}{Rate (\%)} & \multicolumn{4}{c}{Class against own $2\sigma$ band} & \multicolumn{3}{c}{Score change} \\
\cmidrule(lr){4-5}\cmidrule(lr){6-9}\cmidrule(l){10-12}
Run & Aligned & $n$ & Repair & Regr. & Gain & In band & Offset & None & Up & Noise & Down \\
\midrule
EvolveMem, Qwen3-max & yes & 14 & 33.7 & 5.7 & 8  & 0 & 3  & 3\,(1)   & 7  & 3  & 4  \\
                     & no  & 33 & 8.4  & 4.7 & 0  & 2 & 9  & 22\,(6)  & 0  & 22 & 11 \\
\addlinespace[2pt]
GPT-4o diagnoser     & yes & 27 & 27.3 & 4.5 & 10 & 1 & 3  & 13\,(4)  & 11 & 7  & 9  \\
                              & no  & 20 & 12.9 & 5.7 & 1  & 1 & 5  & 13\,(3)  & 1  & 11 & 8  \\
\addlinespace[2pt]
GPT-4o-mini diagn.   & yes & 44 & 11.3 & 5.0 & 9  & 9 & 3  & 23\,(10) & 8  & 24 & 12 \\
                              & no  & 16 & 8.9  & 5.6 & 0  & 4 & 3  & 9\,(7)   & 0  & 11 & 5  \\
\addlinespace[2pt]
GPT-4o-mini answers  & yes & 20 & 30.5 & 7.7 & 10 & 2 & 6  & 2\,(1)   & 11 & 4  & 5  \\
                              & no  & 32 & 12.7 & 8.4 & 1  & 1 & 11 & 19\,(7)  & 4  & 11 & 17 \\
\addlinespace[2pt]
ARC, Qwen3-max agent & yes & 10 & 40.6 & 7.1 & 7  & 3 & 0  & 0\,(0)   & 7  & 2  & 1  \\
                              & no  & 4  & 19.7 & 16.2 & 0 & 0 & 2  & 2\,(2)   & 0  & 1  & 3  \\
\addlinespace[2pt]
ARC, GPT-4o agent    & yes & 41 & 25.8 & 7.3 & 19 & 7 & 7  & 8\,(1)   & 17 & 19 & 5  \\
                              & no  & 19 & 12.6 & 4.4 & 2  & 2 & 8  & 7\,(3)   & 3  & 8  & 8  \\
\bottomrule
\end{tabular*}
\tabnote{Columns as in Table~\ref{tab:alignment}. Edits without a change or with fewer than five attributable parent failures are excluded.}
\end{table}

\paragraph{Feedback variability.}
The flips of Section~\ref{sec:feedback} carry over to every substitution. We pool each loop's own same-configuration re-evaluations with a re-evaluation of its round 0. On these pairs, the three GPT EvolveMem runs flip 3.1--3.6\% of category 1--4 outcomes, compared with 2.8\% for the Qwen loop. Of their accepted steps, 10 of 20, 15 of 25, and 8 of 15 change the score by less than $2\sigma$ of this variation. The GPT-4o ARC agent records no such pair. Under the $2\sigma$ of its Qwen counterpart, 20 of its 38 accepted steps fall within the band, against 9 of 16 for that counterpart. On a same-input re-evaluation of round 0, GPT-4o-mini answers flip 48 of 1,540 category 1--4 outcomes (3.1\%), against 53 (3.4\%) for Qwen3-max. The label checks of Table~\ref{tab:labelagree} also cover a GPT-4o-mini probe solver.

\paragraph{Alignment ledger.}
As shown in Table~\ref{tab:substalign}, the per-edit picture of Table~\ref{tab:alignment} holds under every substitution. Aligned edits repair 26--31\% of their parents' failures, compared with 13\% for misaligned ones, and the GPT-4o-mini diagnoser is lower on both sides (11\% against 9\%). Of the 52 gains beyond noise in the four substituted runs, 48 are aligned. Regressions again barely depend on alignment. In the three substituted EvolveMem runs, aligned and misaligned edits differ by at most 1.2 pp in their regression rate. The GPT-4o ARC agent, in contrast, regresses more on its aligned edits (7.3\% against 4.4\%), which are its larger retrieval changes. The gains also sit where Section~\ref{sec:local} places them, on parents that have not yet enabled semantic retrieval. This holds for 10 of the 11 gains with the GPT-4o diagnoser, 10 of 11 with the GPT-4o-mini answer model, and all 9 with the GPT-4o-mini diagnoser. The GPT-4o ARC agent is the one run that keeps gaining after the upgrade, with 12 gains before it and 9 after, 7 of them aligned. It goes on enlarging retrieval settings, whereas the Qwen agent turns to the extraction and answer prompts that the guard then stops.

\begin{table}[htbp]
\centering\small
\caption{Repair and regression volumes under model substitution. The measurement of Table~\ref{tab:retention} on the runs of Table~\ref{tab:substitutions}, all scored by F1$>$0, with the Qwen runs repeated for reference. E and L are early and late revisions, and Unchanged gives the flips of the same configuration and memory evaluated twice. $^{\S}$Fewer than five such pairs; --: the run never evaluates an unchanged configuration. The Mem-Gallery ARC rows use the 17 scenarios of the GPT-4o run.}
\label{tab:substretention}
\setlength{\tabcolsep}{4pt}
\begin{tabular*}{\linewidth}{@{\extracolsep{\fill}}l*{6}{c}@{}}
\toprule
 & \multicolumn{2}{c}{Repair} & \multicolumn{2}{c}{Regression} & \multicolumn{2}{c}{Unchanged} \\
\cmidrule(lr){2-3}\cmidrule(lr){4-5}\cmidrule(l){6-7}
Run & E & L & E & L & Rep. & Reg. \\
\midrule
\multicolumn{7}{@{}l}{\emph{LoCoMo}} \\
EvolveMem, Qwen3-max          & 5.0 & 2.9 & 2.4 & 3.3 & 1.0 & 1.0 \\
GPT-4o diagnoser              & 8.8 & 1.8 & 3.4 & 3.2 & 1.5 & 1.6 \\
GPT-4o-mini diagnoser         & 4.2 & 3.2 & 3.0 & 3.5 & 1.4 & 1.2 \\
GPT-4o-mini answer model      & 8.8 & 2.6 & 5.5 & 4.2 & 1.5 & 1.7 \\
ARC open/inc, Qwen3-max agent & 7.6 & 2.2 & 3.7 & 3.0 & 1.2 & 1.9 \\
GPT-4o agent                  & 9.6 & 2.9 & 3.4 & 5.3 & 1.2\few & 2.5\few \\
\midrule
\multicolumn{7}{@{}l}{\emph{Mem-Gallery}} \\
EvolveMem, Qwen3-max          & 3.5 & 2.2 & 1.3 & 2.7 & 1.0 & 0.7 \\
GPT-4o diagnoser              & 2.3 & 2.8 & 1.5 & 2.1 & 1.7 & 1.1 \\
GPT-4o-mini answer model      & 1.8 & 2.6 & 1.6 & 2.7 & 0.5 & 1.3 \\
ARC open/inc, Qwen3-max agent & 5.2 & 1.9 & 1.4 & 2.7 & 0.6 & 0.4 \\
GPT-4o agent                  & 7.7 & 2.4 & 2.5 & 3.1 & -- & -- \\
\bottomrule
\end{tabular*}
\end{table}

\paragraph{Retention.}
We repeat the measurement of Table~\ref{tab:retention} on these runs (Table~\ref{tab:substretention}). On LoCoMo, repair volume falls from early to late revisions in every substituted run. With the GPT-4o diagnoser, for instance, it falls from 8.8 to 1.8 pp, and with the GPT-4o ARC agent from 9.6 to 2.9 pp. Regression volume stays or rises, so late revisions regress at least as many questions as they repair in all four runs, and late repairs approach the 1.2--1.7 pp that the unchanged systems flip. On Mem-Gallery, the GPT-4o ARC agent shows the same profile, whereas the two GPT EvolveMem runs have no early peak to lose. Their repairs stay between 1.8 and 2.8 pp throughout, close to the flips of the unchanged system, which matches their smaller gains. Their early-to-late differences of 0.5 and 0.8 pp lie within the spread across revisions (permutation $p=0.34$ and $0.06$, against $0.04$ for the 1.3-pp fall of the Qwen run). Their regressions still rise, from 1.5 to 2.1 pp and from 1.6 to 2.7 pp, and only in the GPT-4o diagnoser run do late repairs stay above regressions. Across the seven substituted runs, 5--19\% of the repairs that a step newly makes fail again at the next update, against 7--12\% for the Qwen runs of Section~\ref{sec:retention}.

Overall, the substituted model changes which stage the loop keeps editing, how early it reaches the retrieval upgrade, and with it the in-sample gain. The three readings of the main text hold on every run. The feedback flips on the scale of single-step gains, repairs concentrate on aligned edits while regressions do not, and repairs decay toward the flips of an unchanged system. The conclusions of Sections~\ref{sec:feedback}--\ref{sec:retention} therefore do not rest on the Qwen family.

\subsection{Diagnosis funnel}
\label{app:diagnosis_funnel}
Section~\ref{sec:local} scores a claim, its proposal, and its executed action separately. This experiment orders the stage checks into a sequence and records, for every round, the first check that fails, so that the rounds surviving each check can be counted. We run it on the two closed ARC configurations, closed/r0 and closed/inc. In closed/r0, the released loop, the diagnoser reads the summary of the root in every round. In closed/inc it reads the summary of the incumbent that the loop currently keeps. Both use the same ten LoCoMo conversations, the same editable action space, and the claim as the stage of the diagnosis, so the version the diagnoser reads is the only difference between them.

\paragraph{Checks.}
Rounds whose version read has fewer than five attributable failures are unscorable and excluded before the first check. L1 compares the claim with the bottleneck stage of the version the diagnoser read, $R_{\rm seen}$. L2 asks whether that claim still holds for the parent being edited, $R_{\rm applied}$, and can fail only when the two versions differ. L3 requires the proposal to be executed and non-empty, so blocked and no-op edits fail here. L4 requires the executed action to cover the parent's bottleneck stage. L5 requires a change in the raw score above the $0.011$ noise floor.

\begin{figure}[htbp]
\centering
\includegraphics[width=\linewidth]{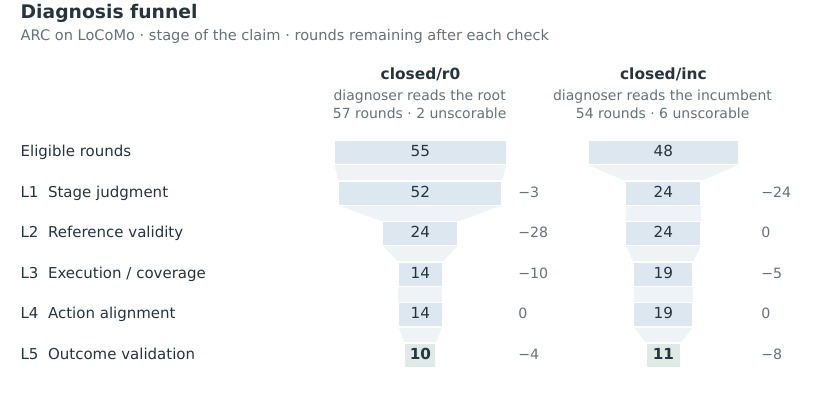}
\caption{Diagnosis funnels of ARC closed/r0 and closed/inc, with the claim as the stage of each diagnosis. Widths and internal counts show the rounds remaining after each check on a shared count scale, and the numbers at right show the rounds that first fail that check. Unscorable rounds are excluded before L1.}%
\label{fig:diagnosis_funnel}%
\end{figure}

\paragraph{Results.}
Figure~\ref{fig:diagnosis_funnel} shows the two funnels. In closed/r0, 55 of 57 rounds are scorable and 52 pass L1, because the agent claims retrieval in 54 rounds and retrieval is the bottleneck of the root in all ten conversations. The loop nevertheless edits an evolved incumbent in 47 rounds, and in 28 of them the parent's bottleneck has moved away from retrieval, so these rounds first fail at L2. Of the remaining 24 rounds, 10 are blocked or empty edits (L3), none miss the parent's bottleneck stage (L4), and 4 change the score within noise (L5), which leaves 10 rounds with a gain. In closed/inc, $R_{\rm seen}$ equals $R_{\rm applied}$ in all 54 rounds, so L2 cannot fail. Instead, 24 of the 48 scorable rounds first fail at L1. The agent claims retrieval in 44 rounds and answer generation in 10, and never extraction, whereas the bottleneck of the version it reads is extraction in 12 and answer generation in 10 of the 24 failing rounds. Five further rounds fail at L3, none at L4, and 8 at L5, which leaves 11 rounds with a gain. We find that reading the incumbent moves the failures from L2 to L1 while barely raising the number of rounds that reach a gain. The incumbent's summary removes the stale reference, but it does not carry the provenance that would let the agent recognize a bottleneck in extraction or answer generation.

\paragraph{Reading the counts.}
These are ordered counts of first failures rather than a decomposition of causal effects, since an earlier failure can mask a later one. Applied to EvolveMem with the proposal in place of the claim, the same sequence places 56 of 67 rounds at L1 and none at L2, although $R_{\rm seen}$ differs from $R_{\rm applied}$ in 39 rounds. Those 56 rounds fail on the stage of the proposal, and Table~\ref{tab:stageaccuracy} scores the claims of the same loop separately.

\subsection{Gold evidence on observed MemPro failures and judge sensitivity}
\label{app:gold}
This subsection applies the gold-evidence probe of Section~\ref{sec:protocol} to the failures that MemPro's loop actually receives. It also tests how far the held-out gains of the selected versions depend on the judge and on the selection metric.

\paragraph{Gold evidence.}
For the failures of each root and for every failed question--version record of the trees, we first replay the failure at temperature 0 with the version's own integration and answer prompts. The gold evidence is supplied only when the replay fails (Table~\ref{tab:actualgold}). On LoCoMo, the 117 root failures split almost evenly into replay-correct, gold-recovered, and gold-unresolved failures (36, 40, and 40). Whether a gold-unresolved failure stays wrong depends on the answer model. Answering once from the annotated gold turns under a neutral prompt (Appendix~\ref{app:prompt_gold}), at least one of four answer models (Qwen3-max, Qwen3-30B-A3B, GPT-4o, and GPT-4o-mini) answers 27 of the 40 correctly. The other 13 stay wrong under all four, and six of them are open-domain questions. On Mem-Gallery, a fifth of the 136 root failures are replay-correct and a quarter are gold-recovered, and 21 adversarial questions have no evidence. Of the 52 gold-unresolved root failures, 36 are answered correctly by at least one of the four models and 16 by none.

\begin{table}[htbp]
\centering\small
\caption{Probes on the failures that MemPro's loop actually receives. Replay uses temperature 0 rather than the original 0.3 and keeps each version's integration and answer prompts. Gold evidence is supplied only after the replay fails, and a gold-correct failure is assigned to a stage. The ``All versions'' rows count failed question--version records rather than questions. Mem-Gallery failures without evidence are adversarial questions.}
\label{tab:actualgold}
\begin{tabular}{@{}lN{0,000}N{000}N{000}N{000}N{000}@{}}
\toprule
Set & Failures & Replay correct & Gold correct & Gold wrong & No evidence \\
\midrule
\multicolumn{6}{@{}l}{\emph{LoCoMo}} \\
Root, seed 2027 & 35 & 3 & 18 & 14 & 0 \\
Root, seed 2028 & 42 & 21 & 7 & 14 & 0 \\
Root, seed 2029 & 40 & 12 & 15 & 12 & 1 \\
Root total & 117 & 36 & 40 & 40 & 1 \\
All 48 versions (records) & 1,234 & 368 & 475 & 381 & 10 \\
\midrule
\multicolumn{6}{@{}l}{\emph{Mem-Gallery}} \\
Root, seed 2026 & 27 & 6 & 9 & 8 & 4 \\
Root, seed 2027 & 43 & 11 & 10 & 15 & 7 \\
Root, seed 2028 & 41 & 4 & 15 & 18 & 4 \\
Root, seed 2029 & 25 & 7 & 1 & 11 & 6 \\
Root total & 136 & 28 & 35 & 52 & 21 \\
All 64 versions (records) & 1,453 & 318 & 612 & 338 & 185 \\
\bottomrule
\end{tabular}
\end{table}

\paragraph{Judge and selection metric.}
Fixed-answer rejudging changes the measurement without changing the system's answers. Over 11,405 unique zero-F1 question--prediction pairs from the loops, the Qwen and GPT judges each accept some of the nominally failed answers and agree with each other on 95.5\% of the pairs. As shown in Table~\ref{tab:judges}, rejudging the same held-out answers with GPT-4o-mini lowers the gain of every selected version, and the gain of seed 2029 changes sign. The selection metric matters in the same way. A MemPro run on the seed-2028 split selects by category 1--4 F1 for fifteen rounds, with no judge in its feedback or selection. On the 1,386 held-out questions, it ends $+5.70$ F1 points above the root but only $+0.29$ pp in judge accuracy. The judge-selected round-15 version of the same split gains $+3.17$ F1 points and $+5.56$ pp. Re-evaluated with its own F1 evaluator, the F1-selected run flips 6.1\% of outcomes with a $2\sigma$ of 3.2 pts. Of the 9 steps on the lineage of its selected version, 7 fall within $2\sigma$.

\begin{table}[htbp]
\centering\small
\caption{Judge sensitivity on fixed MemPro held-out answers. Each column compares the selected version of the fifteen-round run on each split (Table~\ref{tab:cumulative}) with the same re-evaluated root answers under that column's judge.}
\label{tab:judges}
\begin{tabular}{@{}lcc@{}}
\toprule
Seed & Qwen gain (pp) & GPT gain (pp) \\
\midrule
2027 & $+5.55$ & $+1.87$ \\
2028 & $+5.27$ & $+4.04$ \\
2029 & $+2.74$ & $-0.15$ \\
\bottomrule
\end{tabular}
\end{table}

\paragraph{Sources of re-evaluation flips.}
To separate judge nondeterminism from generation and extraction variance, we hold MemPro's LoCoMo root (\texttt{v0000}, seed 2028) fixed and repeat one stage at a time (Table~\ref{tab:noisecontrols}). Rejudging fixed answers at temperature zero flips 0.43\% of development outcomes and 0.31\% of held-out outcomes, whereas generating new answers from stored memory flips 7.47\% and rebuilding memory before answering flips 7.79\%. We also score 21,098 same-version MemPro answer pairs under both metrics. The judge changes its verdict on 9.83\% of pairs, comprising 9.67\% where the answer text also changes and 0.16\% with identical text, while token F1 flips on 8.12\% at F1$>0$ and 11.48\% at F1$\geq0.5$. Nearly all re-evaluation flips on MemPro's root thus originate in answer generation rather than in the judge or memory rebuilding.

\begin{table}[htbp]
\centering\small
\caption{Repeated evaluations of MemPro's LoCoMo root (\texttt{v0000}, seed 2028). Fixed answers are rejudged three more times at temperature zero. The other conditions generate answers three more times from stored memory or perform three fresh builds from the conversations and answer after each build. Entries give means and SDs across scores and across pairwise flip rates, respectively.}\label{tab:noisecontrols}
\begin{tabular}{@{}llcc@{}}
\toprule
Operation repeated & Questions & Accuracy (\%) & Pairwise flip rate (\%) \\
\midrule
Judge fixed answers & Dev.\ (154) & $82.14\pm0.38$ & $0.43\pm0.34$ \\
                    & Held-out (1,386) & $79.30\pm0.10$ & $0.31\pm0.06$ \\
Answer from fixed memory & Dev.\ (154) & $82.63\pm1.44$ & $7.47\pm0.90$ \\
Rebuild memory and answer & Dev.\ (154) & $81.82\pm2.83$ & $7.79\pm1.72$ \\
\bottomrule
\end{tabular}
\end{table}

\paragraph{Reading the counts.}
Gold-unresolved failures cannot be attributed to memory extraction or retrieval. Many of them can still be answered under another answer model or prompt, so whether they count as failures depends on the answering setup and the judge rather than on the memory. A held-out gain of a few points therefore has to be reported together with the judge and the selection metric that produced it.

\subsection{Resolution of the development set}
\label{app:bigsplit}
The development set decides every acceptance of the loop, so it has to resolve the steps the loop accepts. This subsection estimates how large a step the 154 development questions of Section~\ref{sec:intervention} can resolve, and tests whether a development set twice as large, evaluated three times per version, resolves the later steps.

\paragraph{154 development questions.}
We find that on the 154-question development split, the steps accepted in later rounds fall well below the split's own noise floor (Table~\ref{tab:resolution}). Two sources of variation set that floor near 6 pp. First, drawing 154 questions from all 1,540 LoCoMo questions introduces a sampling spread of $2\sigma = 6.3$ pp across 20,000 bootstrap draws from the root's fixed predictions. Consistent with this 95\% interval of [72.08, 84.42]\%, four random splits span 75.97--82.47\% around the 78.64\% full-pool score. Second, even with the split fixed to seed 2028, five re-evaluations of the unchanged root span 74.03--81.82\%, giving a re-evaluation band of $2\sigma = 5.76$ pp, or about 7.1 pp when the loop's own 72.73\% initial evaluation is included. Against this noise floor, the accepted steps of rounds 6--10 move only one or two questions, or 0.65--1.3 pp. Paired acceptance narrows the gap without closing it, since excluding fragile questions shrinks the unchanged-root $2\sigma$ band to 2.74 pp on 154 questions and 1.35 pp on 308 questions, yet the late steps still sit inside that narrower band. A simple scaling check shows why small development sets hit this ceiling. When a fraction $\rho$ of $N$ questions flips between two evaluations of one version, a single comparison carries a binomial $2\sigma$ band of about $2\sqrt{\rho/N}$, so resolving a step of size $\Delta$ requires $N\geq 4\rho/\Delta^2$. At MemPro's flip rates of 8.7--13.2\%, reliably separating genuine late steps of 0.65--1.3 pp with single evaluations would take roughly 2,000 to 12,500 development questions, more than the entire 1,540-question LoCoMo benchmark.

\begin{table}[htbp]
\centering\small
\caption{Development-set resolution for MemPro seed 2028, against the size of accepted steps.}
\label{tab:resolution}
\begin{tabular}{@{}lc@{}}
\toprule
Quantity & pp \\
\midrule
Resample the 154 questions, $2\sigma$ & 6.3 \\
Re-evaluate the root, $2\sigma$ & 5.8 \\
Gold-recoverable failures (5)$^{\rm a}$ & 3.2 \\
Accepted step, rounds 1--5 & 1.3--4.5 \\
Accepted step, rounds 6--10 & 0.65--1.3 \\
One question & 0.65 \\
\bottomrule
\end{tabular}
\tabnote{Accepted steps are measured by their net stable repairs in the repair arms. $^{\rm a}$Root stable failures that turn correct when answered from their annotated gold turns under a neutral prompt. Fuller evidence recovers more, and the value does not bound the effect of changing prompts, code, or answer formatting.}
\end{table}

\paragraph{308 development questions.}
On the 308-question split of Section~\ref{sec:intervention}, stratified by conversation, category, and the outcome of the gold-evidence probe with the other 1,232 questions held out, one run of unmodified MemPro (E0) produces 16 versions evaluated three times each on the development questions. Averaging three evaluations narrows the $2\sigma$ band of the root to 2.54 pp. Yet only 2 of the 15 revisions exceed it against their parents, at $+2.71$ and $+2.60$ pp, and both add date anchoring for temporal questions. Along the lineage of the last version, held-out accuracy rises from 77.92\% to 83.85\%. Only these two steps, however, raise the temporal questions significantly, by 15.6 and 9.7 pp, and the other questions change within noise at every step. Across all 15 revisions, 18 questions turn from wrong in all three evaluations of the parent to correct in all three of the child, compared with 6.0 expected for the unchanged root, and 15 of them are temporal. The five questions that the first of the two revisions repairs all stay wrong when the gold evidence is supplied, so its gain does not come from failures at the bottleneck stage. The other 13 revisions move the score by at most 1.6 pp and stay inside the band.

Taken together, the two splits suggest that the later gains are smaller than the resolution of the development set, which makes them difficult to select reliably. Doubling the development set and averaging three evaluations resolves only the largest steps, and each revision changes too few questions for its stage alignment to be tested. These estimates assume that flips are independent across questions, and they cover a single seed and a single run of E0.

\section{Full Tables by Method}
\label{app:tables}
This section gives, for each method separately, the quantities behind Tables~\ref{tab:alignment}, \ref{tab:retention}, and~\ref{tab:arms}, together with the held-out gains of every split.

\subsection{Stage agreement and edit alignment}
\label{app:diagnosis}
Table~\ref{tab:stageaccuracy} reports the stage agreement of claims and proposals on all eligible records, and Table~\ref{tab:alignfull} breaks the ledger of Table~\ref{tab:alignment} down by revision class.

\begin{table}[!htb]
\centering\small
\caption{Stage agreement on eligible records. Statement accuracy (A) and proposal alignment (B) have different eligible populations.}
\label{tab:stageaccuracy}
\begin{tabular}{@{}llcc@{}}
\toprule
System & Record & $n$ & Agreement \\
\midrule
EvolveMem & Proposal (B) & 67 & 0.16 \\
EvolveMem & Statement (A) & 34 & 0.76 \\
ARC closed/inc & Statement (A) & 48 & 0.50 \\
ARC open/inc & Statement (A) & 49 & 0.22 \\
\bottomrule
\end{tabular}
\tabnote{ARC open/inc here is the version that appends re-extracted memories to the existing bank; the rebuild version used elsewhere is scored in Table~\ref{tab:substitutions}.}
\end{table}

Statement--proposal disagreement does not by itself prove a misdirected proposal, since execution constraints, available operators, and the stated reasons must also be checked. EvolveMem additionally has $R_{\rm seen}\ne R_{\rm applied}$ in 39/67 rounds. The original ARC loop reads the initial version in all 57 rounds and edits an evolved incumbent in 47 of them, and in 28 the stale reference is where the diagnosis first fails (Appendix~\ref{app:diagnosis_funnel}). These mismatches are recorded directly, rather than inferred from a final score.

\begin{table}[!htb]
\centering\small
\caption{Full breakdown of Table~\ref{tab:alignment}. Each revision is classified against its own $2\sigma$ re-evaluation band: gain (repairs and net change beyond the band), in band (repairs beyond it, positive net change within it), offset (repairs beyond it, regressions $\geq$ repairs), none (repairs within it). Positive/neutral/negative counts use the $\pm0.011$ noise floor on the raw score and are separate from the binary statistics.}
\label{tab:alignfull}
\setlength{\tabcolsep}{4pt}
\begin{tabular}{@{}lcN{00}N{00.0}N{00.0}cc@{}}
\toprule
System & Aligned? & $n$ & Repair (\%) & Regression (\%) & gain / in band / offset / none & $+\,/\,0\,/\,-$ \\
\midrule
EvolveMem & Yes & 14 & 33.7 & 5.7 & 8 / 0 / 3 / 3 & 7/3/4 \\
 & No & 33 & 8.4 & 4.7 & 0 / 2 / 9 / 22 & 0/22/11 \\
ARC closed/r0 & Yes & 14 & 40.2 & 3.6 & 11 / 0 / 0 / 3 & 10/4/0 \\
 & No & 9 & 13.0 & 5.0 & 0 / 0 / 3 / 6 & 1/5/3 \\
ARC open/inc & Yes & 10 & 40.6 & 7.1 & 7 / 3 / 0 / 0 & 7/2/1 \\
 & No & 4 & 19.7 & 16.2 & 0 / 0 / 2 / 2 & 0/1/3 \\
\bottomrule
\end{tabular}
\end{table}

Table~\ref{tab:proposer} gives the round-by-round gains of the three proposers compared in Section~\ref{sec:local}.

\begin{table}[!htb]
\centering\small
\caption{Gain of the incumbent under three proposers that share EvolveMem's root, guard, acceptance rule, and 12-recipe menu. Entries are the in-sample gain (pts) of the incumbent's mixed score over round 0 after each round, averaged over ten conversations.}
\label{tab:proposer}
\setlength{\tabcolsep}{3pt}
\begin{tabular*}{\linewidth}{@{\extracolsep{\fill}}llN{$+00.0$}N{$+00.0$}N{$+00.0$}N{$+00.0$}N{$+00.0$}N{$+00.0$}N{$+00.0$}@{}}
\toprule
 & & \multicolumn{7}{c}{Gain after round (pts)} \\
\cmidrule(l){3-9}
Proposer & Choice & \hd{1} & \hd{2} & \hd{3} & \hd{4} & \hd{5} & \hd{6} & \hd{7} \\
\midrule
A0 & Upstream LLM & $+4.1$ & $+5.9$ & $+6.1$ & $+7.0$ & $+7.8$ & $+7.8$ & $+7.9$ \\
A2 & Three random orders & $+2.2$ & $+2.7$ & $+5.6$ & $+6.5$ & $+7.3$ & $+8.1$ & $+8.3$ \\
A3 & Bottleneck-stage item & $+10.5$ & $+10.8$ & $+11.5$ & $+11.7$ & $+12.0$ & $+12.2$ & $+12.3$ \\
\bottomrule
\end{tabular*}
\tabnote{A0 follows the upstream LLM diagnosis; A2 walks the menu in a random order, pooled over three orders (30 runs); A3 takes the next untried item for the incumbent's bottleneck stage. The three A2 orders end at $+11.6$, $+4.4$, and $+9.0$ pts. Early-stopped runs keep their last incumbent.}
\end{table}

In Table~\ref{tab:proposer}, A0 may propose two changes per round, whereas A2/A3 take one menu action. A3 recomputes the incumbent's attributable stages and reaches the useful retrieval upgrade in round 1. The comparison assesses the proposal policy as instantiated, including its action menu and ordering. It is in-sample, and the three random orderings do not support a general equivalence claim for A0 and random search.

\paragraph{Operator scope.}
Across the EvolveMem/ARC edits, 59--84\% of the parent's correct questions receive a different retrieved context, and 427 of the 499 regressions occur on them (Figure~\ref{fig:plane}b), and among EvolveMem's misaligned edits that lower the score, the share is 77 of 93. MemPro gates some edits more narrowly. Restricting an edit to the diagnosis's named category does not isolate its repairs. An offline composition control for revisions that do not rebuild memory takes child answers inside the named category and parent answers outside it. It retains 31\%/26\%/27\% of repairs for EvolveMem/ARC closed/ARC open, comparable to 34\%/25\%/27\% for equally sized random category gates, while net gains decline. This is an answer-composition analysis rather than an executed runtime gate, so an edit's affected query set has to be measured rather than inferred from its diagnosis.

\paragraph{MemPro and GEPA stage analysis.}
Under the sequential protocol, MemPro has 29 aligned, 11 unaligned, and five unscored actions among 45 revisions, and GEPA has 25 aligned and seven unaligned actions among 32 fully evaluated revisions. The bottleneck stage is S2. Forty-two additional GEPA revisions rejected by the small-batch filter do not enter its full-query table. Since these two protocols have different filtering and observability, they are not pooled into Table~\ref{tab:alignment}. A split with 308 development questions does not settle MemPro's alignment either, as each of its revisions changes too few questions (Appendix~\ref{app:interventions}).

\subsection{Repair, regression, and retention}
\label{app:retention}
Tables~\ref{tab:fulltrends}--\ref{tab:judgetrends} report, with counts and under both scores, the quantities of Section~\ref{sec:retention}. Every run splits its revisions into a first and a second half in production order, per conversation or scenario for EvolveMem and ARC and per seed for MemPro and GEPA (rounds 1--7 and 8--15 for MemPro). Table~\ref{tab:fulltrends} scores every row of Table~\ref{tab:retention} by F1$>$0 and adds two F1-selected arms, the loop of the EvolveMem paper and a MemPro run that selects by F1 for fifteen rounds on the seed-2028 split. Judge-selected EvolveMem and ARC runs are in Table~\ref{tab:judgetrends}.

The LongMemEval column of Table~\ref{tab:retention} repeats the measurement on the third benchmark. Repair volume falls from early to late revisions there for all six methods, whereas regression volume falls by less or rises, so late revisions break at least as many questions as they repair for every method. The late repairs of ARC closed/r0, MemPro, and GEPA are within one point of their no-edit flips, whereas those of the other three loops stay above them. The aligned-edit ledger of Table~\ref{tab:alignment} is not repeated on LongMemEval, because an EvolveMem or ARC version there has about eight attributable failures on average among its 100 development questions, most of them on the same few questions, and only one of the 30 scored ARC edits there is aligned.

The two query-image arms of Appendix~\ref{app:galleryvlm} repeat the same counts on Mem-Gallery (Table~\ref{tab:galleryvlm}c). Repair volume falls from $4.5$ to $1.9$ pp in the image arm and from $5.0$ to $2.2$ pp in the caption arm, while regression volume rises from $2.0$ to $2.9$ pp in the image arm and stays at $2.9$ pp in the caption arm. Late revisions therefore break at least as many questions as they repair in both arms, as in the rows above. These two runs are a check on the caption substitution of Section~\ref{sec:protocol} and do not enter the eighteen settings of Section~\ref{sec:retention}.

\begin{table}[!htb]
\centering\small
\caption{All settings scored by F1$>$0: repair and regression volumes in pp of evaluated queries. Rows of Table~\ref{tab:retention}, whose MemPro and GEPA runs select by the judge, plus two F1-selected arms. ``No edit'' is unchanged-system repair/regression volume with the number of comparison pairs; next loss is the share of new repairs and of controls that are wrong after the next update (\%, as in Table~\ref{tab:nextloss}).}
\label{tab:fulltrends}
\setlength{\tabcolsep}{4.5pt}
\begin{tabular}{@{}llcN{00.0/0.0}ccc@{}}
\toprule
Data & System & Early/late $n$ & Repair E/L & Regress E/L & No edit (pairs) & Next loss \\
\midrule
LoCoMo & EvolveMem & 29/38 & 5.0/2.9 & 2.4/3.3 & 1.0/1.0 (20) & 8.3/2.0 \\
 & ARC closed/r0 & 26/31 & 8.7/1.2 & 1.8/2.3 & 1.1/1.3 (31) & 9.8/1.2 \\
 & ARC closed/inc & 23/31 & 10.9/2.8 & 2.7/5.5 & 0.6/0.8 (3) & 7.7/2.2 \\
 & ARC open/inc & 20/23 & 7.6/2.2 & 3.7/3.0 & 1.2/1.9 (31) & 0.0$^{\S}$/11.2 \\
 & MemPro & 28/32 & 5.3/4.5 & 4.5/4.4 & 5.2/5.7 (10) & 33.3/3.5 \\
 & GEPA & 20/22 & 5.1/4.8 & 4.9/5.3 & 3.5/4.7 (9) & 46.3/3.9 \\
 & EvolveMem paper loop$^{\rm a}$ & 20/26 & 7.9/4.2 & 3.5/3.1 & 2.0/0.0 (1) & 23.6/2.9 \\
 & MemPro F1-selected & 7/8 & 5.0/4.2 & 4.3/4.1 & 5.2/5.7 (10) & 14.5/3.9 \\
Gallery & EvolveMem & 60/79 & 3.5/2.2 & 1.3/2.7 & 1.0/0.7 (10) & 11.7/1.3 \\
 & ARC closed/r0 & 48/54 & 7.2/1.0 & 1.3/2.5 & 0.3/0.4 (23) & 7.8/0.9 \\
 & ARC closed/inc & 53/64 & 6.8/1.2 & 1.6/2.1 & 0.5/0.0 (3) & 7.1/1.1 \\
 & ARC open/inc & 34/47 & 4.4/1.8 & 1.4/2.4 & 0.6/0.5 (55) & 3.4$^{\S}$/3.5 \\
 & MemPro & 28/32 & 4.4/3.3 & 4.1/3.0 & 2.3/3.5 (1) & 29.0/2.2 \\
 & GEPA & 5/5 & 4.2/3.2 & 4.6/2.7 & 2.3/3.5 (1) & 26.7$^{\S}$/0.9 \\
\bottomrule
\end{tabular}
\tabnote{$^{\S}$Fewer than 50 new repairs. $^{\rm a}$The non-elitist loop of the EvolveMem paper's Algorithm 1, instead of the released code's default elitist loop, from the same round 0.}
\end{table}

\begin{table}[!htb]
\centering\small
\caption{Share of new repairs (wrong at $t-1$, correct at $t$) and controls (correct at both) that are wrong at $t+1$ along lineages (\%), by whether step $t+1$ changes configuration or memory. Rows and scores as in Table~\ref{tab:retention}, with MemPro and GEPA scored by the judge.}
\label{tab:nextloss}
\begin{tabular*}{\linewidth}{@{\extracolsep{\fill}}llN{000}N{00.0}N{00.0}N{000}N{00.0}c@{}}
\toprule
 & & \multicolumn{3}{c}{After an update} & \multicolumn{3}{c}{After no edit} \\
\cmidrule(lr){3-5}\cmidrule(l){6-8}
Data & System & $n$ & New & Control & $n$ & New & Control \\
\midrule
LoCoMo & EvolveMem & 145 & 8.3 & 2.0 & 33 & 6.1\few & 0.5 \\
 & ARC closed/r0 & 92 & 9.8 & 1.2 & 167 & 5.4 & 0.9 \\
 & ARC closed/inc & 273 & 7.7 & 2.2 & 0 & --- & --- \\
 & ARC open/inc & 13 & 0.0\few & 11.2 & 97 & 5.2 & 0.8 \\
 & MemPro & 336 & 28.9 & 4.6 & 0 & --- & --- \\
 & GEPA & 160 & 36.9 & 5.5 & 0 & --- & --- \\
Gallery & EvolveMem & 163 & 11.7 & 1.3 & 2 & 0.0\few & 1.5 \\
 & ARC closed/r0 & 179 & 7.8 & 0.9 & 87 & 1.1 & 0.1 \\
 & ARC closed/inc & 211 & 7.1 & 1.1 & 2 & 0.0\few & 0.0 \\
 & ARC open/inc & 29 & 3.4\few & 3.5 & 20 & 15.0\few & 0.2 \\
 & MemPro & 344 & 33.4 & 2.5 & 0 & --- & --- \\
 & GEPA & 35 & 28.6\few & 1.1 & 0 & --- & --- \\
\bottomrule
\end{tabular*}
\tabnote{$n$: new repairs followed by that step type; $^{\S}$fewer than 50; ---: no such step.}
\end{table}

Table~\ref{tab:nextloss} follows individual repairs one accepted step ahead. The two groups are selected differently, since a new repair is defined by a recent flip, which selects less stable questions, whereas a control has two consecutive correct outcomes. The higher loss of new repairs therefore also appears without any edit (for example, 5.4\% versus 0.9\% for ARC closed/r0) and is not evidence that later edits target recent repairs, and only the adjacent step is measured. In a separate pure-re-evaluation sequence, MemPro loses 12/25 (48\%) of queries that just flipped correct, GEPA 14/31 (45\%), and ARC open/inc 8/14 (57\%), while controls lose 4.8\%, 4.4\%, and 1.1\%, respectively. Those samples show why one flip should be verified before it is treated as a lasting repair. Other pure-re-evaluation cohorts contain at most four newly correct queries and are not used for a broad comparison. Counts are adjacent question-steps, which can share versions and queries, so no independence-based significance claim is made.

\begin{table}[htbp]
\centering\small
\caption{Judge-selected EvolveMem and ARC runs, scored with their own judge: repair/regression volumes (pp) and next-update failure rates (\%). MemPro and GEPA, which select by the judge natively, are in Table~\ref{tab:retention}.}
\label{tab:judgetrends}
\begin{tabular}{@{}llcN{00.0/0.0}cc@{}}
\toprule
Data & System & Early/late $n$ & Repair E/L & Regress E/L & Next loss: new/control \\
\midrule
LoCoMo & EvolveMem & 30/39 & 7.3/2.5 & 2.2/2.8 & 11.9/1.2 \\
 & ARC closed/r0 & 24/28 & 10.4/2.0 & 2.3/4.6 & 10.9/1.2 \\
 & ARC closed/inc & 15/23 & 16.2/2.3 & 1.9/4.9 & 8.0/0.9 \\
 & ARC open/inc & 18/22 & 4.9/1.3 & 2.2/2.5 & 0/0 \\
Gallery & EvolveMem & 59/79 & 3.8/4.0 & 2.9/4.5 & 26.2/3.3 \\
\bottomrule
\end{tabular}
\end{table}

ARC open/inc has only two newly repaired queries followed by an update in the judge-selected cohort. Gallery EvolveMem is an exception to declining repair volume under this selection metric. Selection metric and outcome metric must therefore be specified together, because cross-scoring a judge-selected tree with F1 answers a different question.

\paragraph{Estimating repairs and regressions above no-edit flips.}
Table~\ref{tab:steppartition} separates real edit effects from evaluation noise to show why late rounds stall. For EvolveMem and the two ARC variants, we subtract the directional no-edit flip rates $\varepsilon_+$ and $\varepsilon_-$, estimated from 20 unchanged-version pairs in EvolveMem and 31 in each ARC variant, from the observed wrong-to-right and right-to-wrong fractions $F$ and $B$ in Equation~\ref{eq:decomposition}. We find that early and late revisions diverge in two ways. First, no-edit flips grow from roughly a quarter of all early flips to a third in EvolveMem and about two thirds in ARC, so single-pass changes become increasingly dominated by noise. Second, the real repairs that remain above noise collapse while real regressions persist or grow. In ARC closed/r0, adjusted repairs fall from 7.62 to 0.13 pp while regressions nearly double to 1.05 pp, making the late adjusted net gain strictly negative, with a 95\% conversation-bootstrap interval of $[-1.26,-0.54]$ pp. EvolveMem and ARC open/inc still fix 1.88 and 1.01 pp of failures above noise in late rounds, yet concurrent regressions of 2.30 and 1.03 pp cancel those gains, leaving late net intervals of $[-1.82,1.17]$ pp and $[-0.73,0.95]$ pp around zero.

MemPro shows an even sharper version of this pattern when individual flips are verified directly. On the 16-version run over 308 development questions (Section~\ref{sec:intervention}), where each version has three evaluations on identical stored memory, we retain a stable repair only when the parent fails all three evaluations and the child passes all three, and apply the reverse rule for a stable regression. We find that 90--95\% of MemPro's single-pass flips are unconfirmed by three evaluations. Across the last eight revisions, only 4 repairs and 7 regressions out of 232 single-pass flips are stably reproduced, while the remaining 221 involve unstable answers whose direction three evaluations cannot settle. That leaves 0.16 pp of stable late repairs per revision, indistinguishable from the 0.15 pp per direction produced by splitting six evaluations of the unchanged root into two groups of three. Taken together, the decomposition shows two reasons late rounds stop producing net progress. In MemPro, nearly all late single-pass flips are unconfirmed noise and stable repairs drop to the level of an unchanged root, whereas in the F1-selected loops, the real repairs that remain are offset or outweighed by concurrent regressions.

\begin{table}[htbp]
\centering\small
\caption{Single-pass flips per LoCoMo revision, in pp of evaluated questions. Parentheses give the no-edit or unconfirmed share of all flips. The first three loops use binary F1 and report no-edit flips and the excess in each direction. MemPro uses judge outcomes on 308 development questions and reports unconfirmed flips and stable transitions. Early and late halves follow Table~\ref{tab:retention} (7 and 8 revisions on the MemPro tree). Entries are rounded independently.}\label{tab:steppartition}
\begin{tabular}{@{}llN{00.00}ccc@{}}
\toprule
Loop & Half & All flips & No-edit flips & Adj.\ repairs & Adj.\ regressions \\
\midrule
EvolveMem     & Early & 7.37 & 2.07 (28\%) & 3.97 & 1.34 \\
              & Late  & 6.25 & 2.07 (33\%) & 1.88 & 2.30 \\
ARC closed/r0 & Early & 10.56 & 2.40 (23\%) & 7.62 & 0.54 \\
              & Late  & 3.59 & 2.40 (67\%) & 0.13 & 1.05 \\
ARC open/inc  & Early & 11.23 & 3.13 (28\%) & 6.37 & 1.73 \\
              & Late  & 5.16 & 3.13 (61\%) & 1.01 & 1.03 \\
\midrule
 & & & Unconfirmed & Stable repairs & Stable regressions \\
\midrule
MemPro & Early & 10.16 & 9.18 (90\%) & 0.65 & 0.32 \\
       & Late  &  9.42 & 8.97 (95\%) & 0.16 & 0.28 \\
\bottomrule
\end{tabular}
\end{table}

\subsection{Held-out gains and turnover}
\label{app:heldout}
Table~\ref{tab:cumulative} gives each split's held-out gain over its original root, and Table~\ref{tab:churn} gives the question-level turnover along the same lineages.

\begin{table}[htbp]
\centering\small
\caption{Held-out gains of MemPro and GEPA by split. Gain (pp) of the selected version over each split's original root, Qwen judge. MemPro: 15 rounds; GEPA: its own prompt search.}
\label{tab:cumulative}
\begin{tabular*}{\linewidth}{@{\extracolsep{\fill}}l*{4}{c}@{}}
\toprule
System & Split 1 & Split 2 & Split 3 & Split 4 \\
\midrule
\grouprow{5}{LoCoMo}
\quad MemPro & ---     & $+3.03$ & $+5.56$ & $+3.39$ \\
\quad GEPA   & $+4.18$ & $-0.72$ & $-0.65$ & $-0.79$ \\
\grouprow{5}{Mem-Gallery}
\quad MemPro & $+6.30$ & $+2.34$ & $+4.87$ & $+5.84$ \\
\quad GEPA   & $+0.84$ & ---     & ---     & --- \\
\bottomrule
\end{tabular*}
\tabnote{---: no result. MemPro's round-15 selection on LoCoMo split 1 was not evaluated on held-out questions; its best version after eight rounds gains $+6.06$. GEPA was run on Mem-Gallery split 1 only. Split 3 is the seed-2028 split used in Tables~\ref{tab:noise} and~\ref{tab:arms}; its entry is the round-15 version against the original root evaluation (78.21\%), whereas Table~\ref{tab:arms} counts the round-10 version of the same run as one of four E0 runs, against the re-evaluated root (78.50\%); Table~\ref{tab:noise}b gives run-to-run variation. Seeds per split: Appendix~\ref{app:setup}.}
\end{table}

\begin{table}[htbp]
\centering\small
\caption{Held-out turnover along the final MemPro lineages. LoCoMo, 1,386 held-out questions, Qwen judge. Transitions are summed over consecutive evaluated versions on the lineage of each split's final version.}
\label{tab:churn}
\begin{tabular*}{\linewidth}{@{\extracolsep{\fill}}l*{6}{c}@{}}
\toprule
Seed & Steps & W$\to$C & C$\to$W & Changed (\%) & Net & Root rerun \\
\midrule
2027 & 9 & 809 & 767 & 41.1 & $+42$ & 150 \\
2028 & 6 & 472 & 395 & 28.9 & $+77$ & 162 \\
2029 & 8 & 624 & 577 & 32.6 & $+47$ & 157 \\
\bottomrule
\end{tabular*}
\tabnote{W$\to$C / C$\to$W: wrong-to-correct and correct-to-wrong transitions summed over lineage steps; individual steps flip 108--236 questions. Changed: questions whose outcome changes at least once along the lineage. Net: end-to-end change in correct questions, equal to W$\to$C $-$ C$\to$W; it corresponds to the $+3.03$/$+5.56$/$+3.39$ pp of Table~\ref{tab:cumulative}. Root rerun: questions whose outcome differs when the original root is evaluated again.}
\end{table}

\subsection{Intervention arms}
\label{app:interventions}
Tables~\ref{tab:armruns}--\ref{tab:budget} give the details behind Table~\ref{tab:arms}. The initial ledger of the seed-2028 split, classified by Algorithm~\ref{alg:accept}, contains 106 stable-correct, 22 fragile, six replay-correct, and twenty stable-wrong questions, totaling 154, and its two root evaluations score 72.73\% and 79.22\%, whose mean is 75.97\%.

\paragraph{Runs and held-out gains.}
Table~\ref{tab:armruns} lists the round-10 incumbent of every run behind Table~\ref{tab:arms} with its development accuracy and held-out gain, alongside the five-run matched controls under a single judge. E2+repair keeps its round-7 incumbent through rounds 8--10. Every run gains on held-out questions, and the three E1 + repair runs of Table~\ref{tab:arms} end within 0.6 pp of each other, whereas the four E0 runs span 4.4 pp. In E0, E1, and E2, the loop selects the version by its first evaluation, so a second evaluation measures the selected version afresh. Across the five labeled runs, it scores 3--6 pp below the first, which is the selection bias of keeping the highest score~\citep[cf.][]{cawley}, and it still lies 4.5--11.7 pp above the root mean (80.52--87.66\% vs.\ 75.97\%), against held-out gains of 1.2--3.8 pp in the same runs. The development gains are therefore real, and only part of them carries over to held-out questions. In the repair arms, both evaluations enter acceptance, so neither measures the incumbent afresh.

\begin{table}[htbp]
\centering\small
\caption{Round-10 incumbent of every run behind Table~\ref{tab:arms}, plus the five-run matched controls under a single judge. Dev.\ gives the two-evaluation mean on the 154 development questions (single evaluation where marked $^{\rm a}$), and $\Delta$HO the paired gain over the root on the 1,386 held-out questions in pp.}
\label{tab:armruns}
\begin{tabular*}{0.65\linewidth}{@{\extracolsep{\fill}}l*{3}{c}@{}}
\toprule
Arm & Run & Dev. & $\Delta$HO \\
\midrule
E0 (MemPro) & 1 & 86.69 & $+1.15$ \\
 & 2 & 88.96\tmark{a} & $+5.55$ \\
 & 3 & 88.96\tmark{a} & $+3.32$ \\
 & 4 & 89.61\tmark{a} & $+5.12$ \\
E1 (+ ledger) & 1 & 83.44 & $+1.52$ \\
 & 2 & 86.36 & $+2.53$ \\
 & 3 & 86.69 & $+3.75$ \\
E2 (+ failure stages) & 1 & 89.29 & $+2.61$ \\
\addlinespace[2pt]
E1 + repair & 1 & 92.84 & $+3.44$ \\
 & 2 & 92.21 & $+3.25$ \\
 & 3 & 91.88 & $+3.82$ \\
E2 + repair & 1 & 89.29 & $+3.05$ \\
\midrule
\grouprow{4}{\makebox[0pt][l]{Matched single-judge controls$^{\rm b}$ (root dev.\ 81.50\%, held-out 79.37\%)}}
E0 (MemPro) & 1 & 92.21\tmark{a} & $+3.60$ \\
 & 2 & 88.96\tmark{a} & $+2.95$ \\
 & 3 & 87.66\tmark{a} & $+1.08$ \\
 & 4 & 90.91\tmark{a} & $+4.90$ \\
 & 5 & 89.61\tmark{a} & $+3.67$ \\
E0-2x (2-eval mean) & 1 & 88.31 & $+0.21$ \\
 & 2 & 90.26 & $+5.19$ \\
 & 3 & 89.61 & $+2.38$ \\
 & 4 & 87.66 & $+0.93$ \\
 & 5 & 90.91 & $+4.25$ \\
\addlinespace[2pt]
E1 + repair & 1 & 92.86 & $+4.76$ \\
 & 2 & 90.26 & $+3.39$ \\
 & 3 & 93.51 & $+5.62$ \\
 & 4 & 94.16 & $+5.19$ \\
 & 5 & 92.86 & $+4.83$ \\
\bottomrule
\end{tabular*}
\tabnote{$^{\rm a}$Single evaluation on development (unlabeled E0 runs). $^{\rm b}$Because the original judge endpoint was retired, the five-run matched controls run local Qwen3-30B-A3B with Qwen3.7-max as judge (root evaluations: 81.82\% and 81.17\% on development, mean 81.50\%; 79.37\% on held-out). E0-2x evaluates each version twice (memory reused) and selects by the two-evaluation mean without ledger, replay, or repair rounds.}
\end{table}

\paragraph{Feedback arms and targeting.}
E0 receives the loop's own evaluation output and per-query results, and our measurement records are withheld from it. E1 receives repeated evaluation, replay, parent--child differences, and attempted-action records. E2 additionally receives attributable stages and evidence provenance. Root files, development split, and initial evaluation match across arms. E0/E1/E2 each complete ten rounds with the original selection logic. Table~\ref{tab:targetruns} gives the targeting of every labeled run. Stable failures make up 86--97\% of the agent's targets in the three E1 runs and 96\% in E2, against 69\% in E0. In every run, more than half of the questions a child newly answers correctly were fragile or replay-correct in its parent, that is, in uncertain baseline states, and fragile questions alone account for 38--57\%.

\begin{table}[htbp]
\centering\small
\caption{Targeting of every labeled run without paired acceptance, summed over ten rounds. Targets are the questions the agent names as the cause or motivation of a revision, and Stable gives the share of them that are stable failures of the parent. Newly correct questions are wrong in the parent's evaluation and correct in the child's, and the last two columns give the share of them that were fragile or replay-correct in the parent and fragile alone.}
\label{tab:targetruns}
\setlength{\tabcolsep}{4pt}
\begin{tabular}{@{}lcccN{000}cc@{}}
\toprule
Arm & Run & Targets & Stable (\%) & Newly correct & Fragile or replay-correct (\%) & Fragile (\%) \\
\midrule
E0 (MemPro) & 1 & 13 & 69 & 91 & 52 & 38 \\
E1 (+ ledger) & 1 & 81 & 86 & 98 & 66 & 57 \\
 & 2 & 74 & 97 & 98 & 60 & 45 \\
 & 3 & 65 & 86 & 108 & 65 & 47 \\
E2 (+ failure stages) & 1 & 28 & 96 & 106 & 71 & 49 \\
\bottomrule
\end{tabular}
\end{table}

\begin{table}[htbp]
\centering\small
\caption{All eight within-round repair comparisons with held-out results. Each difference pairs the repair child with its uncorrected revision on common questions. Positive and negative threshold crossings are both retained.}
\label{tab:repairpairs}
\begin{tabular}{@{}lcccN{$+0.00$}c@{}}
\toprule
Arm & Round & Paired $n$ & Repair $-$ revision (pp) & $z$ & Reported threshold \\
\midrule
E1 + repair & 2 & 1,377 & $+2.98$ & 3.10 & Positive \\
 & 3 & 1,359 & $+6.25$ & 6.18 & Positive \\
 & 4 & 1,379 & $+2.10$ & 2.31 & Positive \\
 & 5 & 1,371 & $+1.24$ & 1.52 & Within band \\
 & 7 & 1,367 & $+0.73$ & 0.98 & Within band \\
 & 8 & 1,382 & $-1.95$ & $-2.54$ & Negative \\
E2 + repair & 3 & 1,384 & $+0.22$ & 0.26 & Within band \\
 & 5 & 1,378 & $-0.65$ & $-0.76$ & Within band \\
\bottomrule
\end{tabular}
\end{table}

Common-query counts for Table~\ref{tab:repairpairs} range from 1,359 to 1,384. Six pairs are positive, with three positive and one negative $|z|\geq2$ result. The two E2+repair pairs have $z=0.26$ and $-0.76$, both below the threshold. E1+repair generates seven repair children, of which five are accepted, and their mean stable repair/regression counts are 4.3/1.0 versus 4.1/2.7 for the original revisions. In the second E1+repair run, six repair children reach 4.0/1.8 against 3.3/2.2 for their revisions. E2+repair generates nine repair children, none of which is accepted, with corresponding means of 3.0/2.7 versus 3.1/3.0. Thus, the same repair design has different outcomes across arms and runs. Table~\ref{tab:repairchain} lists the incumbents that E1+repair selects and their held-out gains.

\begin{table}[htbp]
\centering\small
\caption{Selected incumbents of the first E1+repair run. Held-out gains use each version's common-query pairing with the root. Later improvements on development do not imply that held-out accuracy rises monotonically.}
\label{tab:repairchain}
\begin{tabular}{@{}lcN{+00.00}N{+00.00}N{+00.00}N{+00.00}N{+00.00}N{+00.00}N{+00.00}N{+00.00}@{}}
\toprule
Accepted step & Root & \hd{1} & \hd{2} & \hd{3} & \hd{4} & \hd{5} & \hd{6} & \hd{7} & \hd{8} \\
\midrule
Dev. (\%) & 75.97 & 82.14 & 84.09 & 87.98 & 89.29 & 91.16 & 91.86 & 93.18 & 92.84 \\
Held-out gain & 0 & +0.22 & +3.76 & +1.81 & +3.05 & +1.88 & +1.37 & +2.34 & +3.44 \\
\bottomrule
\end{tabular}
\end{table}

\paragraph{Budget.}
The arms are matched in rounds rather than in cost, and Table~\ref{tab:budget} reports what each one consumed. Verified feedback adds one full re-evaluation per version and a temperature-0 replay of its 20--40 wrong answers, which the harness runs between rounds, so E1 and E2 evaluate the development set twice per version while the agent's own sessions cost about as much as under E0 (22--27 hours and 120--132 million kimi-k3 tokens per arm). Repair rounds nearly double the rest, since each repair child is a further agent session with its own evaluation and re-evaluation, and the two repair arms produce 17 and 19 versions, spend 26 additional agent hours, and use 265--375 million tokens. E0 therefore matches the other arms in rounds only. To match E1's evaluation budget without its ledger, replay, or repair rounds, we also ran five matched controls each of unmodified MemPro (E0), a two-evaluation mean baseline (E0-2x), and E1 + repair under a single local task model and judge, with per-run values in Table~\ref{tab:armruns}. Adding a second evaluation alone improves neither held-out gain nor run-to-run stability. E0-2x reaches $89.35\pm1.35\%$ on development and gains $+2.59\pm2.12$ pp on held-out questions, compared with $89.87\pm1.76\%$ and $+3.24\pm1.40$ pp for single-evaluation E0. By contrast, E1 + repair reaches $92.73\pm1.48\%$ on development and gains $+4.76\pm0.84$ pp on held-out questions, ahead of both E0 and E0-2x with lower run-to-run spread, though two-sided 95\% confidence intervals against both baselines still include zero (Welch $p=0.08$). Wall-clock is not comparable across arms, because all five ran concurrently on one API quota.

\begin{table}[!htb]
\centering\small
\caption{Budget of the intervention arms. MemPro, seed 2028, ten rounds, first run of each arm. Dev.\ evaluations count full 154-question evaluations, the agent's own (A) plus the harness re-evaluation (B). Replayed: wrong answers replayed at temperature 0 and probed for their stage. Held-out: 1,386-question evaluations. Hours are active session time of the evolving agent, of the repair sessions, and of the harness hooks (B, replay, ledger), and wall-clock from the first to the last round. Agent tokens are kimi-k3 input plus output over all sessions, of which 97--98\% are cached input.}
\label{tab:budget}
\setlength{\tabcolsep}{3pt}
\begin{tabular*}{\linewidth}{@{\extracolsep{\fill}}lccccN{00}ccN{00.0}cc@{}}
\toprule
 & & \multicolumn{2}{c}{Dev.\ evals} & & & \multicolumn{4}{c}{Hours} & Agent \\
\cmidrule(lr){3-4}\cmidrule(lr){7-10}
Arm & Versions & A & B & Replayed & Held-out & Agent & Repair & Hooks & Wall & tokens (M) \\
\midrule
E0 (MemPro) & 10 & 10 & 10\tmark{a} & 238\tmark{a} & 5 & 25.5 & --- & 13.7\tmark{a} & 28.4 & 132 \\
E1 (+ ledger) & 10 & 10 & 10 & 273 & 4 & 22.2 & --- & 9.2 & 32.7 & 120 \\
E2 (+ failure stages) & 10 & 10 & 10 & 206 & 4 & 26.5 & --- & 11.6 & 40.6 & 130 \\
\addlinespace[2pt]
E1 + repair & 10 + 7 & 17 & 17 & 314 & 16 & 27.0 & 25.7 & 11.0 & 67.1 & 375 \\
E2 + repair & 10 + 9 & 19 & 19 & 390 & 7 & 26.4 & 26.3 & 15.2 & 69.8 & 265 \\
\bottomrule
\end{tabular*}
\tabnote{Versions: agent revisions plus repair children. $^{\rm a}$E0's re-evaluations, replays, and hooks were run for measurement only and never shown to the agent. Held-out evaluations of E1 + repair cover every accepted or revised version for Table~\ref{tab:repairpairs}. All arms shared one API quota with other experiments, so wall-clock includes rate-limit waits, and the same 154-question re-evaluation took between 10 and 250 minutes. Calls and tokens of the task model and judge were not recorded.}
\end{table}

\section{Case Studies: Four Failed Edits}
\label{app:cases}
This section traces four edits at the question level and shows by which route an edit breaks questions that its parent answered correctly.

Table~\ref{tab:cases} traces four edits that lower the score, one for each kind of operator, one of which targets its parent's bottleneck stage. For every question that the parent answers correctly and the child answers wrong, we check whether its gold memory still reaches the child's answer context.

\begin{table}[!htb]
\centering\small
\caption{Four edits that lower the score and the route of their regressions. Context changed: parent-correct questions whose retrieved context changes. Gold in context: parent-correct questions whose gold memory reaches the answer context, in the parent and in the child.}
\label{tab:cases}
\setlength{\tabcolsep}{3pt}
\begin{tabularx}{\linewidth}{@{}lYccN{000/000}cY@{}}
\toprule
 & & & Fixed/ & Context & Gold in & \\
 & Edit (stage) & Aligned & broken & changed & context & Main route of the regressions \\
\midrule
A & EvolveMem, intent planning (S2) & yes & 6/21 & 123/124 & 33$\to$17 & gold memory displaced (15 of 21) \\
B & ARC closed/inc, time decay and fusion weights, context 20$\to$15 (S2) & no & 5/60 & 158/158 & 83$\to$36 & gold memory displaced (45 of 60) \\
C & EvolveMem, targeted re-extraction (S1) & no & 4/4 & 41/116 & 4$\to$3 & new memories in context (2 of 4) \\
D & ARC open/inc, per-category answer prompt (S3) & no & 11/15 & 0/139 & 45$\to$45 & answer changes with gold in context (13 of 15) \\
\bottomrule
\end{tabularx}
\end{table}

Cases A and B, one aligned and one not, lose correct answers through the same route, as the edit reorders or shortens the retrieved context and pushes the gold memory out. In A, the parent answers ``Which country was Evan visiting in May 2023?'' with ``Canada'', and the child answers ``Not enough information'' once the memory of the road trip to Jasper leaves the context. In B, ``In what country did Jolene's mother buy her the pendant?'' changes from ``France'' to ``Colombia'' for the same reason. Cases C and D keep most gold memories in the context and change the answer instead. In C, a new near-duplicate of the gold memory enters the context and the answer to ``How did Joanna feel on October 25, 2022 about seeing her characters come alive on the big screen?'' drifts from ``Surreal and satisfying'' to a paraphrase of that memory, and the same question breaks in both targeted re-extractions from this parent. In D, the rewritten answer prompt leaves every context intact, yet ``Which outdoor gear company likely signed up John for an endorsement deal?'' changes from ``Under Armour'' to a refusal, although the supporting memory is still in the context. Whether an edit targets the bottleneck stage thus does not decide how it breaks other questions, since the routes follow the shared component it changes.

\section{Limitations}
\label{app:scope}
\paragraph{Stage-reference coverage.}
Our failure-stage probe abstains on failures that neither replay nor gold evidence resolves, so the stage readings of Section~\ref{sec:local} cover roughly a third of observed failures across three coarse stages. Extending attribution to finer pipeline sub-stages and gold-unresolved failures lies outside the scope of this work.

\paragraph{Development-to-held-out transfer.}
We diagnose the three failure modes primarily on the development questions that drive loop selection, and on LoCoMo and Mem-Gallery the held-out questions share conversation histories with the development set, whereas LongMemEval holds out whole histories (Appendix~\ref{app:lme}). Tracing which individual development repairs and regressions carry over to unseen histories is left to future work.

\paragraph{Intervention scale and statistical power.}
Our intervention arms evaluate MemPro on one 154-question split where the agent inspects the same questions used for acceptance, and with three to five runs per arm, held-out mean differences smaller than about 3 pp remain underpowered. Testing these mechanisms with larger development sets, withheld acceptance splits, and broader method and model coverage is left to future work.

\end{document}